\pdfoutput=1
\documentclass[11pt]{article}

\usepackage[final]{acl}

\usepackage{times}
\usepackage{latexsym}

\usepackage[T1]{fontenc}
\usepackage[utf8]{inputenc}

\usepackage{microtype}

\usepackage{inconsolata}

\usepackage{graphicx}
\usepackage{subfigure}
\usepackage{booktabs} % for professional tables
\usepackage{hyperref}

\usepackage{amsmath}
\usepackage{amssymb}
\usepackage{mathtools}
\usepackage{amsthm}
\usepackage{url}            % simple URL typesetting
\usepackage{nicefrac}       % compact symbols for 1/2, etc.
\usepackage{xcolor}         % colors
\usepackage{multirow}
\usepackage{wrapfig}
\usepackage[capitalize,noabbrev]{cleveref}
\usepackage{algorithm}
\usepackage{algpseudocode}
\theoremstyle{plain}

\theoremstyle{definition}

\theoremstyle{remark}

\title{Spectra 1.1: Scaling Laws and Efficient Inference for Ternary Language Models}

\author{Tejas Vaidhya$^{1,2,3}$\thanks{Equal contribution, listed in alphabetical order.} \;,\;
Ayush Kaushal$^{1, 2, 3}$\footnotemark[1] \;,\;
Vineet Jain$^{2,4}$,\;
Francis Couture-Harpin$^{5}$,\\
\textbf{Prashant Shishodia}$^{6}$,\; \;
\textbf{Majid Behbahani}$^{7}$,\;
\textbf{Yuriy Nevmyvaka}$^{7}$,\; 
\textbf{Irina Rish}$^{1,2,3}$ \\
\textsuperscript{1}Nolano AI \quad
\textsuperscript{2}Mila- Quebec AI institute \quad 
\textsuperscript{3}Université de Montréal \quad
\textsuperscript{4}McGill University \\
\textsuperscript{5}École de technologie supérieure, Université du Québec \quad
\textsuperscript{6}Google, India \quad
\textsuperscript{7}Morgan Stanley \\
}
\vspace{-16mm}

\begin{document}
\maketitle

\begin{abstract}
Large language models (LLMs) are increasingly used across research and industry applications, yet their inference efficiency remains a significant challenge. As the computational power of modern GPU architectures continuously improves, their memory bandwidth and capacity have not scaled proportionally, creating a critical bottleneck during inference. To address this, we investigate ternary language models (TriLMs) that employ quantization-aware training to significantly reduce memory requirements. We first analyze the scalability of TriLMs by conducting a scaling law analysis, revealing that TriLMs benefit more from increasing training data than from scaling model parameters. Based on this observation, we introduce Spectra-1.1, an open suite of TriLMs trained on up to 1.2 trillion tokens, demonstrating sustained performance gains at scale. Furthermore, to improve inference efficiency, we propose novel 2-bit and 1.6-bit packing schemes for ternary weights, which demonstrate accelerated inference across various CPU architectures. Also, building on the 2-bit packing, we develop a GPU kernel called TriRun that accelerates end-to-end model inference by up to 5 times compared to floating-point baselines. To encourage further exploration and development of TriLMs, we will release the Spectra-1.1 suite and TriRun inference kernels. Overall, our work lays the foundation for building and deploying efficient LLMs, providing a valuable resource for the research community.
\end{abstract}

\section{Introduction}

Large language models (LLMs) \citep{gpt2, zhang2022automaticchainthoughtprompting, llama} have become increasingly pivotal in both research and industry. Beyond their broad utility, their capabilities during inference with additional compute demonstrate the potential to enable advancements in reasoning and agentic tasks \citep{sardana2024chinchillaoptimalaccountinginferencelanguage, singh2024humandatascalingselftraining, wei2023chainofthoughtpromptingelicitsreasoning}. As the demand for efficient and scalable inference grows \cite{zhou2024surveyefficientinferencelarge}, significant efforts have been directed toward reducing inference costs and latency \citep{dettmers2022llmint88bitmatrixmultiplication,gptq,flex}. While, the computational power of GPUs has improved rapidly, advancements in memory capacity and bandwidth have lagged behind \citep{gholami2024aimemorywall, spectra}. This disparity has made memory-related bottlenecks a predominant challenge during LLM inference, where memory usage and bandwidth (driven by model size in bits) increasingly outweigh computational (FLOPs) limitations. While post-training quantization, combined with custom kernels for inference acceleration, has become widely adopted, its effectiveness in mitigating these bottlenecks remains limited. Specifically, post-training quantization is typically restricted to 4-bits and results in significant performance degradation beyond this threshold \citep{dettmers2023case4bitprecisionkbit}.

\begin{figure*}[t] % or [!htbp] for more flexible positioning
    \centering
    \vspace{-2em}
    \begin{minipage}[b]{0.63\textwidth} % First figure
        \centering
        \includegraphics[width=1.1\textwidth]{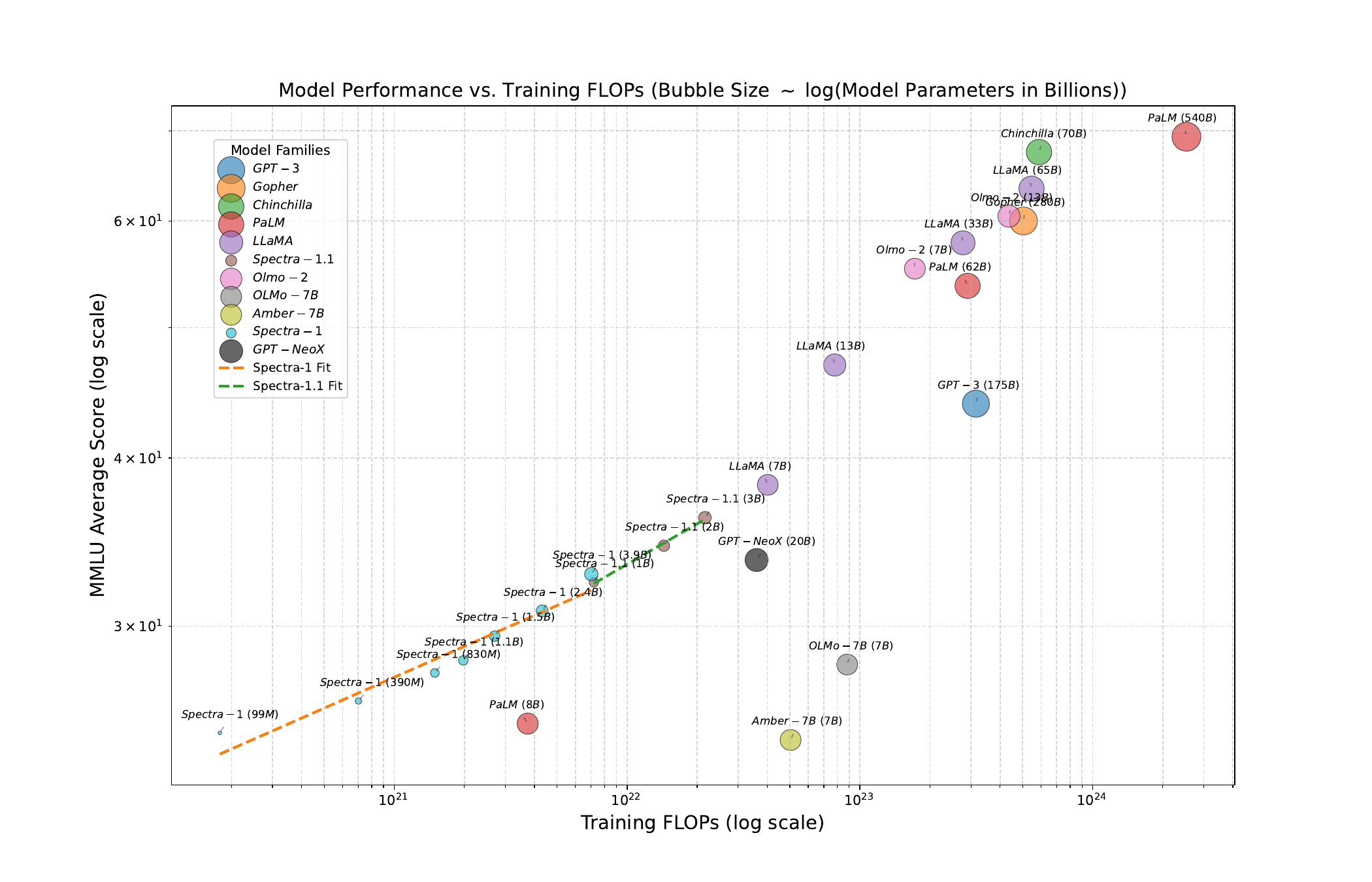}
        \label{fig:side_by_side_graphs1}
    \end{minipage}
    \begin{minipage}[b]{0.339\textwidth} % Second figure
        \centering
        \includegraphics[width=1.1\textwidth]{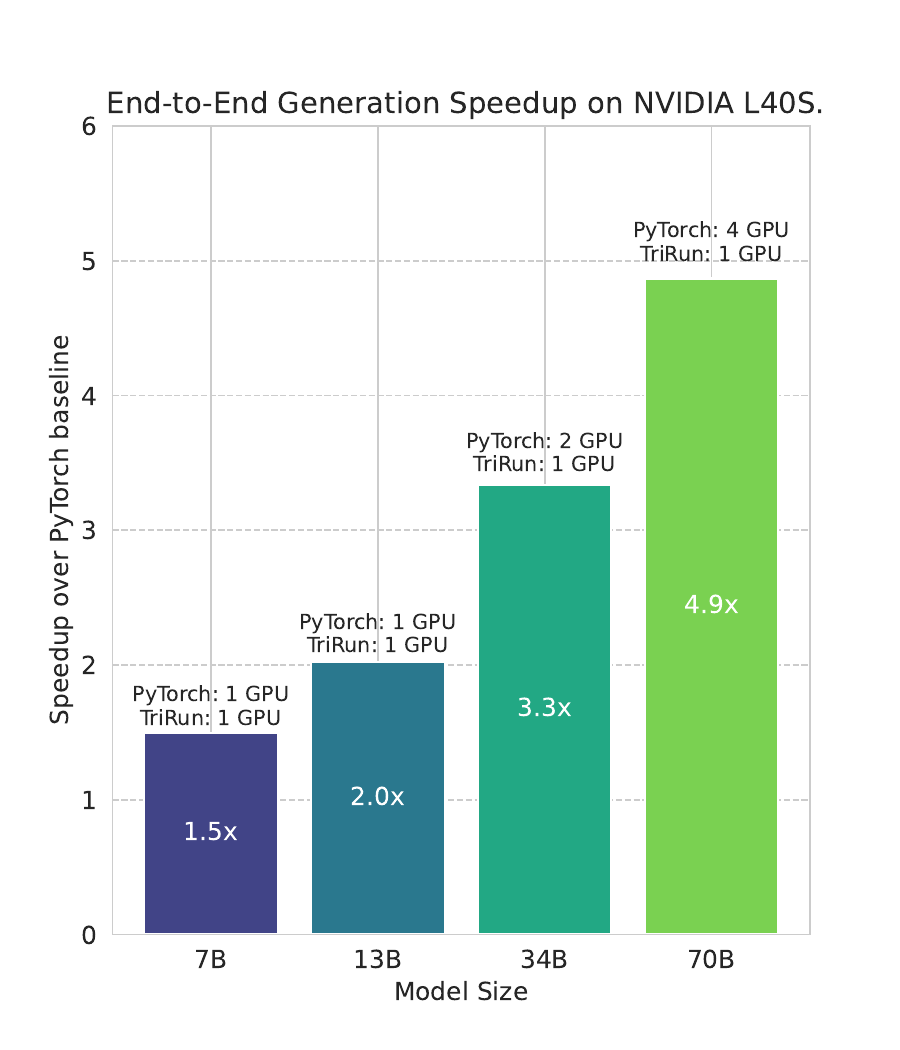}
        \label{fig:side_by_side_graphs2}
    \end{minipage}
    
    \caption{\footnotesize{Model performance (MMLU average accuracy) versus training FLOPs, considering only models with similar compute budgets and training tokens for a fair comparison (left); and end-to-end generation time speedup achieved by TriRun kernels over the PyTorch's FP16 baseline (64 Input Tokens, 64 Output Tokens) on the NVIDIA L40S (right).}}
    \label{fig:side_by_side_graphs}\vspace{-1.1 em}
\end{figure*}    

Recent advancements in extreme low-bit language models \citep{spectra, bitnet, bitnet_1_58} have shown that quantization-aware training allows ternary-weight models to achieve performance comparable to full-precision models (referred to as FloatLMs in this paper) at larger parameter scales. Additionally, ternary representations demonstrate superior bit-efficiency as they scale.
However, critical gaps persist in understanding the scaling laws governing Ternary Language Models (TriLMs)—specifically, \textit{how TriLM performance is affected by training on much larger datasets or with many more parameters} remains unanswered. Furthermore, the acceleration of inference in sub-4-bit models (e.g., ternary) remains unexplored, with most existing research limited to 4-bit quantization \citep{gptq, dettmers2022llmint88bitmatrixmultiplication, marlin, he2024inferenceperformanceoptimizationlarge}. These limitations are compounded by the absence of a comprehensive suite of strong open-source models, suppressing innovation in post-training and broader research on extreme quantization. In this work, we aim to address these foundational challenges through the following contributions:

\vspace{-0.1em}
\paragraph{Scaling law for ternary language models.} We conduct (in \cref{sec:scaling_ternary_models_to_1T_tokens}) a systematic study to explore the scaling properties of TriLMs, focusing on both the number of parameters and the volume of training tokens. Unlike previous works \citep{spectra, bitnet}, which primarily examine parameter scaling, we demonstrate that increasing the number of tokens leads to a greater reduction in validation loss compared to increasing the number of parameters (see \cref{sec:scaling_trilms}).

\vspace{-0.2em}
    
   \paragraph{Effect of scaling pretraining tokens.} We scale the TriLM models by pretraining them on 1.2T tokens (see \cref{subsec:benchmark_spectra}), refered to as the Spectra-1.1 family of models. Our results show that the 3B model continues to improve with up to 1.2T tokens, suggesting that TriLM remains effective even at higher token-to-parameter ratios. Additionally, it achieves competitive performance with FloatLMs for a given compute budget (see  \cref{fig:side_by_side_graphs}, left).
    
    \vspace{-0.2em}

\paragraph{Efficient packing mechanism for ternary weights.} In \cref{sec:packing}, we propose efficient 1.6-bit and 2-bit packing schemes for ternary weights. We provide a theoretical analysis of these packing methods, along with the implementation of efficient kernels and benchmarking on a CPU (see \cref{subsec:CPU_Inference_with_efficient_packing} and \cref{app:cpu_bench}), demonstrating a significant acceleration in inference speed.
    \vspace{-0.2em}
    
   \paragraph{Efficient GPU kernels for ternary models.} We introduce GPU kernels based on 2-bit packing schemes, which we call TriRun (in \cref{sec:trirun_main_section}). We extensively benchmark its performance in model serving settings across various model sizes and different NVIDIA hardware (see \cref{subsec:Performance_of_TriRuns_Kernels} and \cref{app:trirun_perfromance_benchmark}). Notably, we achieve up to a 7-8× speedup compared to PyTorch’s float16 kernels in high-batch settings (16-32 samples) for the ternary layer in transformer blocks of larger parameter (70B - 405B) models on the L40S GPU. Additionally, as shown in \Cref{fig:side_by_side_graphs} (right), our 70B model achieves a 4.9× end-to-end speedup (compared to float16) while running on a single L40S.

    % \item \textbf{Efficient GPU kernels for ternary models.} We introduce GPU kernels based on 2-bit packing schemes, which we call TriRun (\cref{sec:trirun_main_section}). We extensively benchmark its performance in model serving settings across various model sizes and different NVIDIA hardware (see \cref{subsec:Performance_of_TriRuns_Kernels} and \cref{app:trirun_perfromance_benchmark}). Notably, we achieve a 7-8× speedup compared to PyTorch’s float16 kernels in high-batch settings (16-32 batches) for ternary layers in transformer blocks of large-scale models (70B–405B parameters) on the L40S GPU.
    % We also develop kernels based on 2-bit packing schemes for NVIDIA GPUs, which we call TriRun, We extensively benchmark it's performance during serving across model size and different nvidia hardware. We achieve upto \textbf{7-8}$\mathbf{\times}$ speedup compared to PyTorch's float16 kernels in high-batch settings (16-32 batches) of higher parameter models in L40S nvidia-GPU.
    
    % primarily due to the FLOP-to-byte ratio of 100 to 200 \citep{nvidiaA10}.

\begin{figure*}[t!] 
  \centering
  \begin{minipage}{0.45\textwidth}
    \centering
    \includegraphics[width=\linewidth]{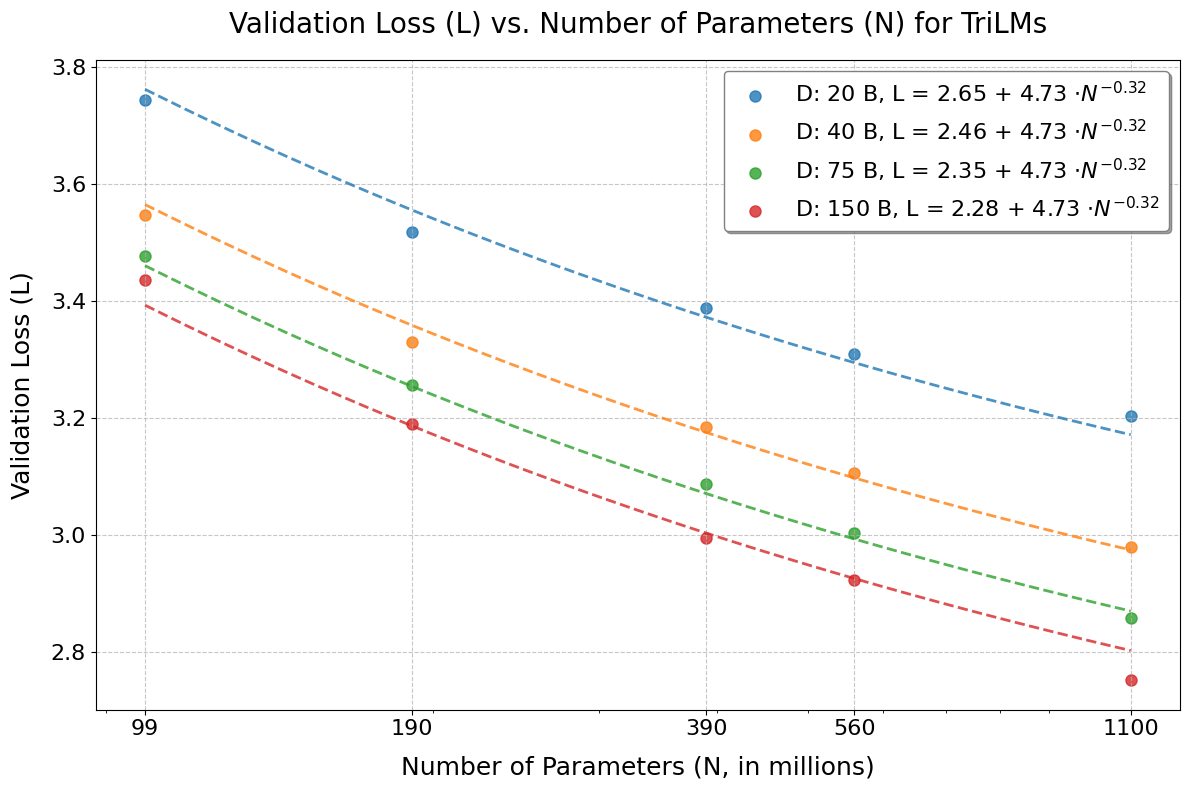} % Adjust the width as needed
    \label{fig:param_scaling_fig} 
  \end{minipage}
  \hfill
  \begin{minipage}{0.45\textwidth}
    \centering
    \includegraphics[width=\linewidth]{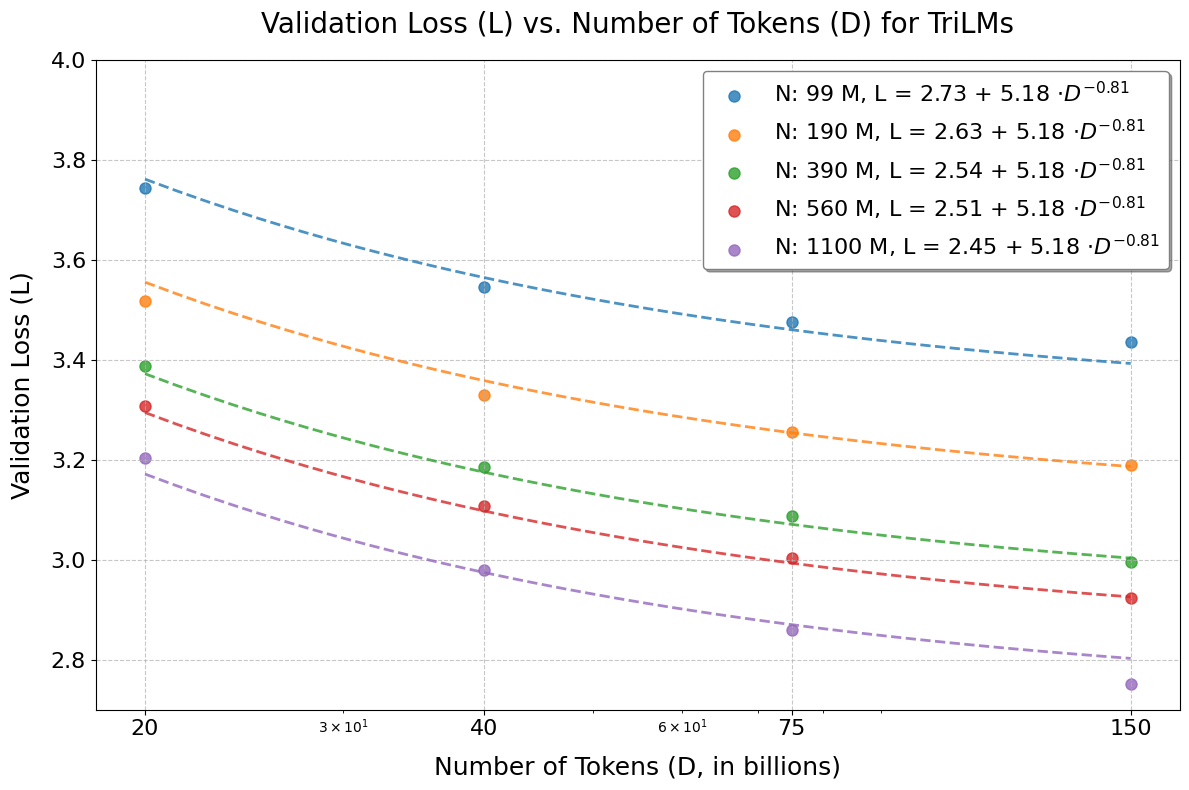} % Replace with your figure's path
    \label{fig:token_scaling_fig}
  \end{minipage}
  \vspace{-1.5em}
  \caption{\footnotesize{Effect of scaling number of parameters (left) and number of training tokens (right) on final validation loss for TriLMs. The dotted lines show the power law derived in \Cref{eq:power_law}}.}
  \vspace{-1.2em}
  \label{fig:scaling_fig}
\end{figure*}

\section{Scaling ternary models to 1T tokens}
\label{sec:scaling_ternary_models_to_1T_tokens}
In this section, we study the scalability of pretraining ternary models. We begin by outlining our training setup, including details about the data, hardware scaling, and model architecture (see \Cref{sec:training_details}). We then analyze the scaling properties of ternary language models with respect to both parameters and training tokens, deriving a scaling law in \Cref{sec:scaling_trilms}. Based on insights from our scaling studies, we train a suite of models on up to 1.2 trillion tokens, which we call Spectra-1.1, and benchmark their performance in \Cref{subsec:benchmark_spectra}.

\subsection{Training Details}
\label{sec:training_details}

\paragraph{Data.} Our training corpus comprises a diverse mix of data from publicly available sources. To scale TriLMs \cite{spectra}, we trained on approximately 1.2 trillion tokens from ArXiv \citep{clement2019arxiv}, Cosmopedia-v2 \citep{benallal2024cosmopedia}, PeS2o \citep{peS2o}, Zyda-StarCoder-Git-Commits,  Zyda-StarCoder-Languages \citep{tokpanov2024zyda}, FineWeb-Edu \citep{lozhkov2024fineweb-edu}. The dataset details are summarized in Table \ref{tab:datasets} and appendix \S\ref{pretraining_data}.  For tokenization, we employ the LLaMA tokenizer over previously used GPT-NeoX tokenizer. 

% \paragraph{Architecture.}  We adopt a decoder-only transformer architecture based on \citet{vaswani2023attentionneed}, closely aligned with the first iteration of TriLM \citep{spectra}, except for using parameterless RMSNorm for normalization. TriLM influenced by LLaMA \citep{llama}, incorporating key components such as  SwiGLU gated MLPs \citep{shazeer2020gluvariantsimprovetransformer}, Rotary Position Embedding (RoPE) \citep{su2023roformerenhancedtransformerrotary}, multi-head attention, and the omission of bias terms. The model's distinguishing feature lies in its representation of linear layers, which utilize ternary states ({-1, 0, 1}) in conjunction with a shared floating-point scale value. During the training phase, we maintain floating-point latent weights to accumulate updates, while implementing on-the-fly ternarization in the forward pass. The scale value is computed as the absolute mean of these weights. For gradient computation, we employ a straight-through estimator, as proposed by \citep{bengio2013estimatingpropagatinggradientsstochastic}. 

\paragraph{Architecture.} Our model follows a decoder-only transformer architecture \citep{vaswani2023attentionneed}, closely resembling the  TriLM  \citep{spectra}. Inspired by LLaMA \citep{llama}, it incorporates SwiGLU MLPs \citep{shazeer2020gluvariantsimprovetransformer}, RoPE \citep{su2023roformerenhancedtransformerrotary}, multi-head attention, and bias-free layers. A key distinction is its ternary-weighted linear layers ({-1, 0, 1}) with a shared floating-point scale. Training maintains latent floating-point weights, applying on-the-fly ternarization in the forward pass, and the scale set to their absolute mean. Hyperparameters and additional pretraining details are provided in \cref{app:pretraining_details}.

\begin{figure}[h]
    \centering
    \includegraphics[width=0.45\textwidth]{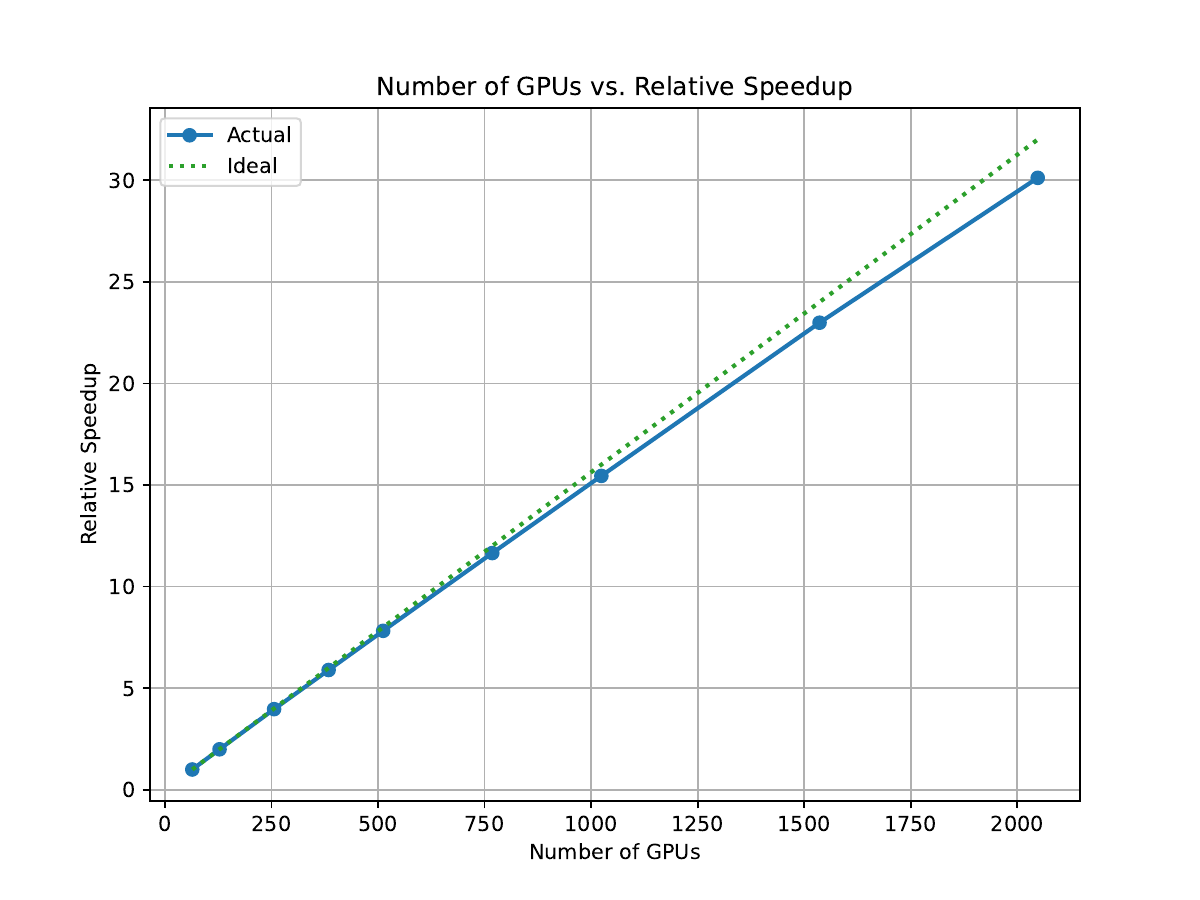}
    \caption{\footnotesize{Number of GPUs vs. Relative Speedup.}
    \label{fig:gpus_speedup}}
\end{figure}

\paragraph{Hardware and Scaling.} We conduct our training experiments on the Frontier\footnote{\url{https://en.wikipedia.org/wiki/Frontier_(supercomputer)}} cluster. Each node comprises four AMD MI250X accelerators \citep{amd_mi250x_datasheet}, where each MI250X contains two Graphics Compute Dies (GCDs) operating as separate GPUs \citep{amd_cdna2_whitepaper}. Within a node, GPUs are at most one hop away from one another, facilitating efficient intra-node communication. Our distributed training strategy is designed with this hardware architecture in mind. Similar to the ZeRO Stage 2 strategy \citep{Samyam_Rajbhandari_zero_2020}, we shard the AdamW optimizer states and gradients, synchronizing model parameters after each update step. However, due to slower inter-node connectivity, sharding is performed only across devices within a node. This approach enables near-linear scaling up to 2,048 GPUs, as shown in Figure~\ref{fig:gpus_speedup}.

% We carry out the training experiments on the Frontier supercomputing\footnote{https://en.wikipedia.org/wiki/Frontier\_(supercomputer)} cluster, where a single node consists of four AMD MI250X accelerators \citep{amd_mi250x_datasheet}. Each MI250X has two Graphics Compute Dies (GCDs), which function as separate GPUs \citep{amd_cdna2_whitepaper}. Within a node, each GPU is at most one hop away from every other GPU. The total bidirectional communication bandwidth varies between 100 GB/s and 400 GB/s. The nodes are connected via Ethernet-based HPE Slingshot interconnects, with four links per node, each providing a total directional bandwidth of 50 GB/s. We design our distributed training strategy accordingly. Similar to the ZeRO Stage 2 scaling strategy \citep{Samyam_Rajbhandari_zero_2020}, we shard the AdamW optimizer states and gradients, synchronizing the model parameters after each update step. However, due to the slow inter-node connectivity, we only shard across devices within a node rather than across all nodes. Our solution achieves near-linear scaling for up to 2,048 GPUs, as shown in \ref{fig:gpus_speedup}.

\subsection{Scaling Laws for TriLMs}
\label{sec:scaling_trilms}

\paragraph{Experimental Setup.} 

For this study ($\leq$150B tokens), we use a SlimPajama subset from \citet{shen2024slimpajamadcunderstandingdatacombinations}, while the 1.2T-token dataset incorporating additional sources (Appendix~\ref{tab:datasets}). All other aspects follow the procedures outlined in Sections \ref{sec:training_details} regarding the pretraining of the models. We train and evaluate a suite of TriLM models, conducting a series of language model pretraining experiments across parameter sizes $\in [99M, 190M, 390M, 560M, 1100M]$ (excluding embeddings) and dataset sizes $\in [20, 40, 75, 150]$ billion tokens. In Section~\ref{subsec:benchmark_spectra}, we expand the training dataset size to 1.2 trillion tokens for models with 1.5 B, 2.5 B, and 3.6 B parameters.

\paragraph{Parametric Scaling Law. } We derive the scaling law for ternary LLMs following the general form introduced in \citet{hoffmann2022trainingcomputeoptimallargelanguage}. In particular, we assume the following functional form for the validation loss $\hat{L}$ as a function of  model size $N$ (number of parameters, in millions) and training data $D$ (number of tokens, in billions),
\begin{equation}
    \hat{L}(N,D) \triangleq E + \frac{A}{N^\alpha} + \frac{B}{D^\beta},
\label{eq:scaling_params}
\end{equation}
where the constant term includes the irreducible loss due to entropy of natural text, plus the error introduced by quantization. Based on the validation losses of the converged models, we fit the parameters $\{E, A, \alpha, B, \beta\}$ (see \Cref{app:scaling} for evaluation of our fit). This provides a scaling law that describes how the validation loss of a ternary model changes with data and model size.
\begin{equation}
    \hat{L}(N,D) \approx 2.19 + \frac{4.73}{N^{0.32}} + \frac{5.18}{D^{0.81}}
\label{eq:power_law}
\end{equation}
\Cref{fig:scaling_fig} shows the final validation loss for different models against the number of parameters and training tokens. For each plot, we also substitute the corresponding value of $D$ or $N$ to get the scaling law equation for that setting. We discuss the implications of this law in more detail and compare with 16-bit models in \Cref{app:scaling}.

From \Cref{eq:power_law}, we observe that increasing the number of tokens lowers the validation loss more effectively than increasing the number of parameters. This suggests that TriLM remains effective at high training token-to-parameter ratios. Based on these observations, we focus primarily on increasing the number of tokens to train our new family of models in the following section.

% Extrapolating from these trends, our largest model (3.6B) could potentially train on 11.8 trillion tokens without convergence, comparable to the scale used for LLaMA-3 \citep{llama3modelcard}, though training at this scale remains a future goal due to resource constraints. 
\subsection{Effect of scaling training tokens}
\label{subsec:benchmark_spectra}
We pre-trained three TriLM models with 1.5 B, 2.5 B, and 3.6 B parameters (for simplicity, we refer to these models as 1B, 2B, and 3B throughout the paper) on a 1.2 trillion-token dataset (detailed in Appendix \ref{pretraining_data}), which we refer to as Spectra-1.1 suite in this paper. The details of the parameters are provided in Table \ref{tab:arch-summary}. Inspired by well-established model suites, such as those by \citep{groeneveld2024olmoacceleratingsciencelanguage, spectra, biderman2023pythiasuiteanalyzinglarge}, Spectra-1.1 aims to provide robust baseline models to advance scientific research on TriLMs.

\paragraph{LLM Benchmarks Performance.} We evaluate the Spectra-1.1 suite of models on a variety of tasks testing commonsense and reasoning abilities, general knowledge, and mathematical problem-solving. A full description of the tasks is given in \Cref{app:benchmark}.

\begin{figure}[h!]
    \centering
\includegraphics[width=0.95\columnwidth]{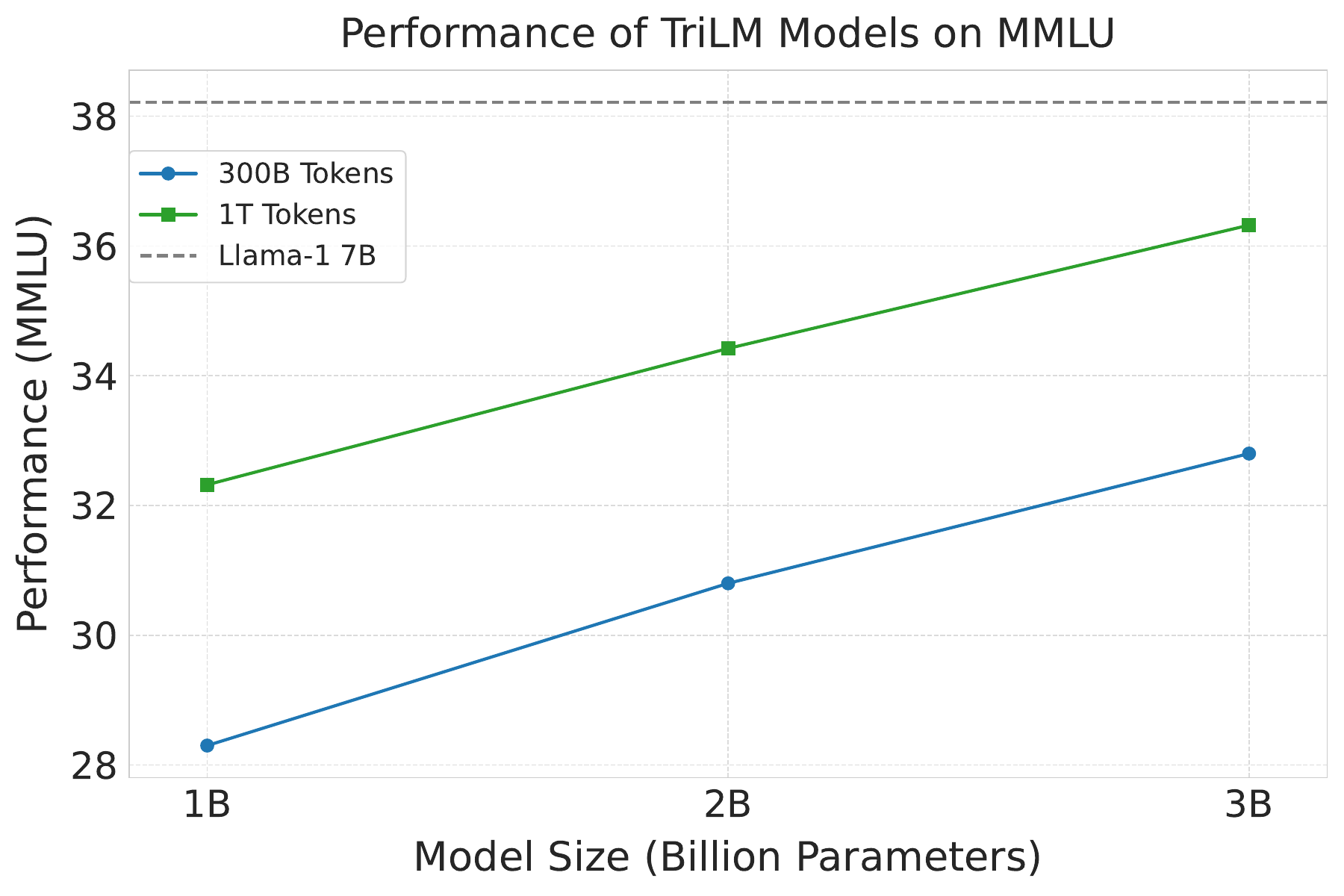}
    \caption{\footnotesize{Average MMLU accuracy for Spectra-1.1 and Spectra, with the dotted line representing LLaMA-1 7B (trained on 1.2T tokens). Note that LLaMA-3 \cite{llama3modelcard}, trained on over 15 trillion tokens, is not included.}}
    \label{fig:mmlu_performance}
\end{figure}

To understand the effect of scaling training tokens on downstream performance, we compared benchmark scores with the Spectra suite of models, which have comparable parameter sizes and were trained on 300B tokens. \Cref{fig:mmlu_performance} shows the average accuracy on the MMLU benchmark for both family of models, demonstrating consistently better performance across different parameter sizes. Full results on individual benchmarks are presented in  \Cref{tab:model-performance}.

\section{Efficient packing of ternary weights}
\label{sec:packing}
In this section, we propose weight-packing strategies and kernel implementations to enable the efficient deployment of ternary LLMs. We begin by formalizing the packing problem and then present two progressively optimized solutions. These solutions target effective 1.6-bit and 2-bit packing, supported by theoretical guarantees. Following this, we conduct a preliminary feasibility assessment on a CPU to evaluate the practicality of our approach.

% This section introduces a packing and unpacking mechanism tailored for ternary LLMs, accompanied by efficient kernel implementations. We first conduct a preliminary feasibility assessment on a CPU before presenting TriRun, a high-batch setting kernel \S\ref{sec:trirun_main_section}. Subsequently, we evaluate its performance across various hardware platforms, demonstrating its advantages in diverse application scenarios.

\paragraph{Definition (Lossless Packing and Unpacking).}
Let \( D = ( d_1, d_2, \dots, d_n ) \) represent a sequence of ternary numbers, where \( d_i \in \{-1, 0, 1\} \). The \textit{packing} process is a function \( P: \{-1, 0, 1\}^n \to B \), which maps the ternary sequence \( D \) to some binary representation \( B \), such that the original sequence \( D \) can be reconstructed from \( B \). The \textit{unpacking} process is the inverse function \( U: B \to \{-1, 0, 1\}^n \), which reconstructs the original sequence \( D \) from \( B \). These processes satisfy the property of losslessness:
\[
U(P(D)) = D, \quad \forall D \in \{-1, 0, 1\}^n.
\]

% \begin{figure*}[t!] % or [!htbp] for more flexible positioning
%     \centering
%     \includegraphics[width=0.44\textwidth]{images/cpu_total_tokens_per_second_64_output.pdf} % Replace with the first figure's path
%     \includegraphics[width=0.44\textwidth]{images/output_tokens_per_second} % Replace with the second figure's path
%     \caption{Comparison of output tokens for different model sizes running on a Mac M4 CPU laptop:
%     \textbf{(Left)} Output tokens (for a 256 prompt with 64 output tokens).
%     \textbf{(Right)} Output tokens per second versus model size.}
%     \label{fig:cpu_tq1_tq_2}
% \end{figure*}

\begin{figure*}[t!] % or [!htbp] for more flexible positioning
    \centering
    \includegraphics[width=0.32\textwidth]{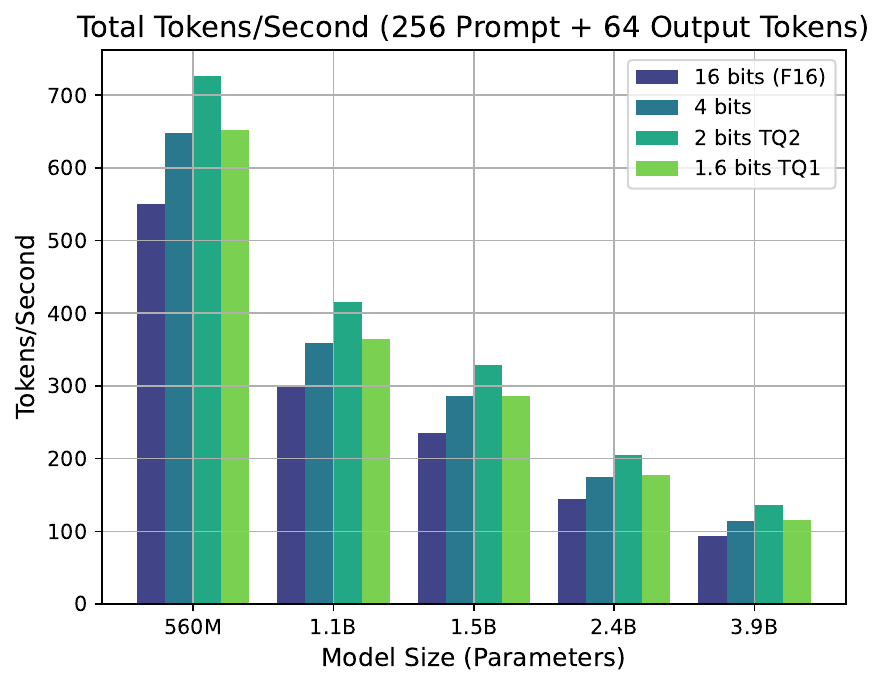} % Replace with the first figure's path
    \includegraphics[width=0.33\textwidth]{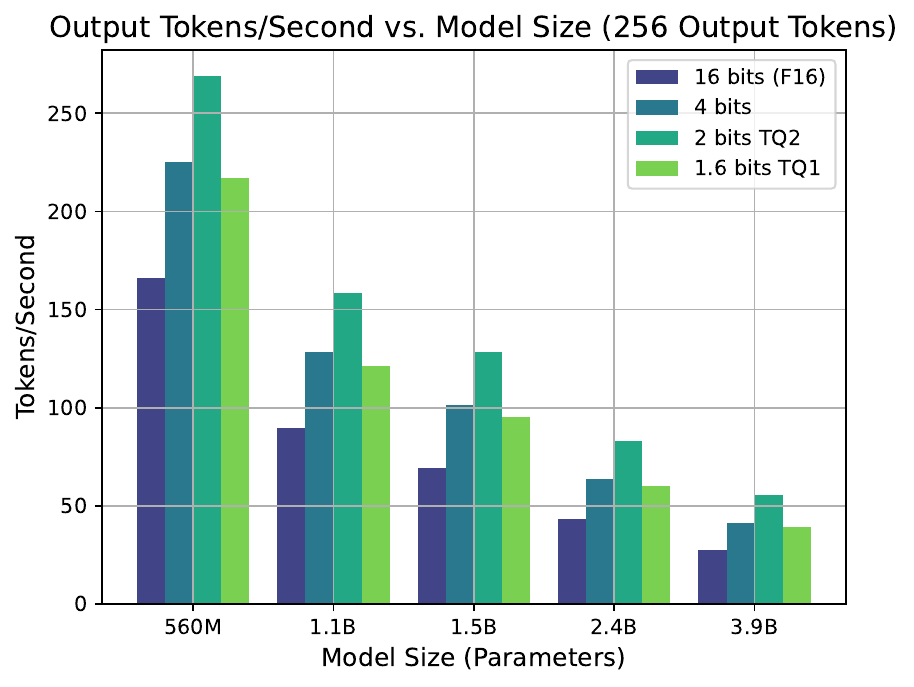} % Replace with the second figure's path
    \includegraphics[width=0.305\textwidth]{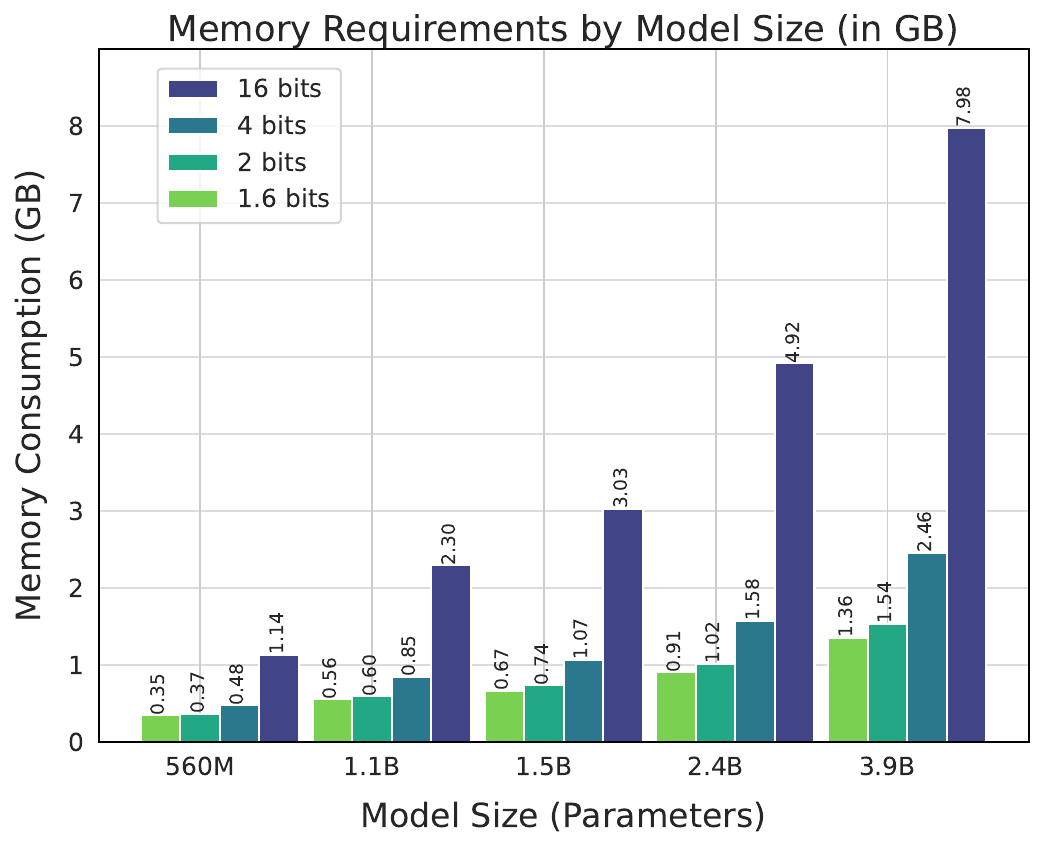} % Replace with the new figure's path
    \caption{\footnotesize{Comparison of output tokens for different model sizes running on a Mac M4 CPU laptop:
    \textbf{(Left)} Total output tokens (for a 256 prompt with 64 output tokens).
    \textbf{(Center)} Output tokens per second versus model size.
    \textbf{(Right)} Memory requirements by model size (in GB) with different Packing Strategies. For more details, refer to  \cref{tab:lM4_Max_14_CPU_Cores}}}
    \label{fig:cpu_tq1_tq_2}
\end{figure*}

\subsection{Packing Strategy with effective 2 bits.}
\label{sec:2bit}

\paragraph{Packing/Encoding}  
The packing process transforms each ternary value \( d_i \in \{-1, 0, 1\} \) into a digit \( d_i' \) by the mapping
\[
d_i' = d_i + 1,
\]
so that \( d_i' \in \{0, 1, 2\} \). These digits are then grouped into blocks of up to \( k \) values. Each block is encoded as a single integer using bitwise shifts. The packing function \( P(D) = ( b'_0, b'_1, \dots, b'_{m-1} ) \) (with \( m = \lceil n/k \rceil \) for an original sequence \( D = \{ d_0, d_1, \dots, d_{n-1} \} \)) is defined as:
\[
b'_i = \sum_{j=0}^{k-1} \left( d'_{ki+j} \cdot 2^{2j} \right),
\]If a block is not completely filled (when \( n \) is not a multiple of \( k \)), the remaining positions are padded with \( 0 \), which map to  \( 1 \)  after the shift. Since each \( d'_i \) is in \(\{0,1,2\}\) and fits within 2 bits, each block uses \(2k\) bits, giving an effective 2 bits per weight.

\paragraph{Unpacking/Decoding}  
The unpacking process \( U(P(D)) = ( d_0, d_1, \dots, d_{n-1} ) \) recovers the original ternary values from the packed representation. For each block, the decoding is defined as:
\[
d_{ki+j} = d'_{ki+j} - 1, \] \[ \quad \text{where} \quad d'_{ki+j} = \bigl( (b'_i \gg 2j) \,\&\, 0x03 \bigr),
\]
for \( i \ge 0 \) and \( 0 \le j < k \). Here, \(\gg\) denotes the bitwise right shift operation and \(\&\) denotes the bitwise AND operation (with \(0x03\) serving as a mask to extract 2 bits). This procedure ensures that each original ternary value \( d_i \in \{-1, 0, 1\} \) is accurately reconstructed from its packed form. Although each \( d_i' \) is constrained to three possible states, the packing allocates a total of \( 2k \) bits per block. However, the actual information content per block is only \(\log_2(3^k) = k \log_2(3) \text{ bits},\) which is strictly less than \( 2k \) bits (since \( \log_2(3) \approx 1.585 \)). In the following, we outline a general strategy for a better effective bit rate.

\subsection{Packing Strategy with 1.6 effective bits.} 

\paragraph{Packing/Encoding:} The packing process transforms each ternary value \( d_i \in \{-1, 0, 1\} \) into a base-3 digit (or trit) \( d_i' = d_i + 1 \), then groups the digits into blocks of up to k. Each block is encoded as a base-3 integer and normalized to fit within \([0, 2^{p}-1]\), where p is an integer representing the number of bits allocated for each encoded block. The packing $P(D) = ( b'_1, b'_2, \dots, b'_k )$ is defined as,
\begin{align*}
    &b'_i = \left\lfloor \frac{\left( \sum_{j=0}^{k-1} d'_{ki+j} \cdot 3^{k-1-j} \right) \cdot 2^{p} + (3^{k}-1)}{3^{k}} \right\rfloor,\\
    &\text{where} \; d'_i = d_i + 1, \; \text{and} \; d_i \in \{-1, 0, 1\}.
\end{align*}
Here, \( k \) is the number of digits in each block (which may be less than $k$ for the last block). The final packed byte array $B$ is then constructed from the $b_i'$ values. 
\paragraph{Unpacking/decoding:}

The unpacking process $U(P(D)) = ( d_1, d_2, \dots, d_n )$ is defined as:

\begin{align*}
    &d_{ki+j} = d'_{ki+j} - 1,\\
    &\text{where} \; 
    x_i = \left\lfloor\frac{b_i \,\times 3^k - \bigl(3^k - 1\bigr) + \bigl(2^p - 1\bigr)}{2^{p}}\right\rfloor,\\
    &\text{and} \; d'_{ki+j} = \left( \left\lfloor \frac{x_i}{3^{k-1-j}} \right\rfloor \right) \bmod 3.
\end{align*}

Here,  k  represents the number of digits in each block, typically equal to 5 for full blocks, though it may be fewer for the final block. For practical purposes, we recommend setting  p = 8  and  k = 5 , as this configuration results in an effective packing of 1.6 bits — very close to the theoretical optimum for ternary data.

% \vineet{Use consistent notation P / \texttt{P} and U / \texttt{U}}.
\paragraph{Theorem 1 (Correctness).}  Let \( D = ( d_1, d_2, \dots, d_n ) \) be a sequence of ternary digits \( d_i \in \{-1, 0, 1\} \). When \( D \) is partitioned into blocks of size \( k \) and each block is encoded into a \( p \)-bit integer, the encoding and decoding operations \( P \) and \( U \) are lossless if and only if \( 2^p > 3^k \); that is, \( U(P(D)) = D \). \textbf{Proof:} See Appendix \ref{subsec:theorem1_correctness}.

% For a sequence \( D = ( d_1, d_2, \dots, d_n ) \) of ternary digits \( d_i \in \{-1, 0, 1\} \), grouped into blocks of size \( k \) and packed into \( p \)-bit integers with \( 2^p > 3^k \), the $P$ and $U$ operations are exact inverses, ensuring \( U(P(D)) = D \). \textbf{Proof:} See appendix \ref{subsec:theorem1_correctness}

\paragraph{Corollary (Injectivity).} 
If \( 2^p > 3^k \), then the mapping \( P: \{ -1, 0, 1 \}^k \to [ 0, 2^p - 1 ] \) is injective i.e.
\( \{ d_j \} \neq \{ d_j' \} \) implies $P(\{ d_j \}) \neq P(\{ d_j' \})$.

% Suppose \( 2^p \geq 3^k \). If \( \{ d_j \} \neq \{ d_j' \} \), then  $
% \texttt{P}(\{ d_j \}) \neq \texttt{P}(\{ d_j' \}).
% $ Equivalently, the mapping from \( \{ -1, 0, 1 \}^k \to \{ 0, \dots, 2^p - 1 \} \) is injective. 

%\textbf{Proof}: See appendix \ref{subsec:lemma_injectivity}

\paragraph{Near‐Optimal Bits per Trit} 

\noindent
From an information‐theoretic perspective, each trit (with values in $\{-1,0,1\}$) requires
$ \log_2(3) \;\approx\; 1.58496 \quad \textit{bits of entropy}.$ To encode $k$ trits without collision, we need a $p$-bit container with $ 2^p \;>\; 3^k 
  \quad\Longrightarrow\quad
  p \;>\; k \,\log_2(3) $. Consequently, the \emph{bits‐per‐trit ratio} is bounded below by
$\frac{p}{k} \;>\; \log_2(3).$
When $p = \lceil k\,\log_2(3)\rceil$, this is effectively the smallest integer $p$ that still allows all $3^k$ trit patterns to be stored with no collisions. As $k \to \infty$, the ratio 
$ \frac{p}{k}
  \;\longrightarrow\;
  \log_2(3), $
making the scheme \emph{asymptotically optimal} in terms of bits used per trit.

% \paragraph{Corollary 1 (Error-Free Conversion):}

% Consequently, the scheme is both:

% Lossless (no collisions, exact recovery), and
% Nearly optimal in terms of compression ratio, because it matches the fundamental lower bound on bits needed to represent trit sequences.

% Given a ternary sequence $D = \{ d_1, d_2, \dots, d_n \}$ such that $n$ is divisible by 5, the packing and unpacking process guarantees error-free conversion, i.e., there is no loss or corruption of data during the transformation.
% Can you write this in the form of theorem

% \paragraph{Theorem 2 Optimal packing (Optional):} 1.6 is the optimal packing in practice for modern hardware? Can someone proof this for me.

% Instead by multiplying this representation by 256, we shift the most significant ternary digit into the top two bits of a 10-bit number, enabling extraction via bit-shifting and masking. Subsequent digits are extracted iteratively by multiplying the remaining value by 3 and isolating the top bits, avoiding division entirely. This method is computationally efficient and particularly suited for high-throughput tasks, where unpacking is significantly more frequent than packing. The only division in the pipeline occurs during packing, which is infrequent relative to unpacking, making this trade-off optimal for large-scale language model workloads. The algorithm is provided in appendix F see.

\subsection{CPU Inference with efficient packing}
\label{subsec:CPU_Inference_with_efficient_packing}
To assess the effectiveness of packings, we implemented both packing methods in \texttt{ggml.cpp}\footnote{https://github.com/ggerganov/llama.cpp}, a framework optimized for running large language models (LLMs) on CPUs. While further optimizations are possible, our primary focus is on reducing a model’s memory footprint and accelerating memory-bound workloads. This is achieved by statically compressing pretrained weights and decompressing them on-the-fly during inference. We begin by showing the efficiency of the CPU implementations for the 1.6-bit packing, referred to as TQ1, and the 2-bit packing, referred to as TQ2. 
%Subsequently, we explore the extension of these methods to high-batch GPU environments.

\paragraph{TQ2: Implementing effective 2 bit for TriLMs.} The quantization process begins with partitioning the input tensor into contiguous, non-overlapping blocks, each containing 256 elements. For each block, a scaling factor (floating-point numbers associated with TriLMs) \(d_i\) is calculated as the maximum absolute value of the elements within the block, i.e.,  \(d_i = \max( |b_{ij}| ),\) where \( b_{ij} \) denotes the \(j\)-th element in the \(i\)-th block. The inverse scaling factor \( \hat{d}_i \) is then defined as   \( \hat{d}_i = \frac{1}{d_i}.\)  
Each element \( b_{ij} \) in the block is quantized to a ternary value by multiplying it by the inverse scaling factor and rounding the result: \( q_{ij} = \text{round}(b_{ij} \cdot \hat{d}_i).\) To enable efficient storage, the quantized values are shifted, resulting in \( q_{ij} \in \{0, 1, 2\} \) and packed into 64 bytes per block of 256 elements using base-4 positional encoding. The scaling factor  \(d_i \)  is stored in 2 bytes (float16), leading to a total storage of 66 bytes per block.
% The quantized elements are then packed into a single byte, where every four quantized values are represented using base-4 positional encoding. 
% As a result, each block of 256 elements is stored in 64 bytes, with the scaling factor \( d_i \) stored in 2 bytes in float16 format, yielding a total of 66 bytes per block. 
The dequantization process begins by reversing the base-4 encoding to recover the four ternary elements (see \ref{app:tq_2_implementation}). The elements are then adjusted back to their signed values by subtracting 1. Finally, the original block is reconstructed by multiplying each quantized value by the corresponding scaling factor:  
\( \hat{b}_{ij} = d_i \cdot q_{ij}, \) where \( \hat{b}_{ij} \) denotes the dequantized approximation of \( b_{ij} \). This quantization scheme achieves significant memory efficiency by compressing 256 floating-point values into just 66 bytes.

\begin{figure*}[t!] % or [!htbp] for more flexible positioning
    \centering
    \includegraphics[width=0.94\textwidth]{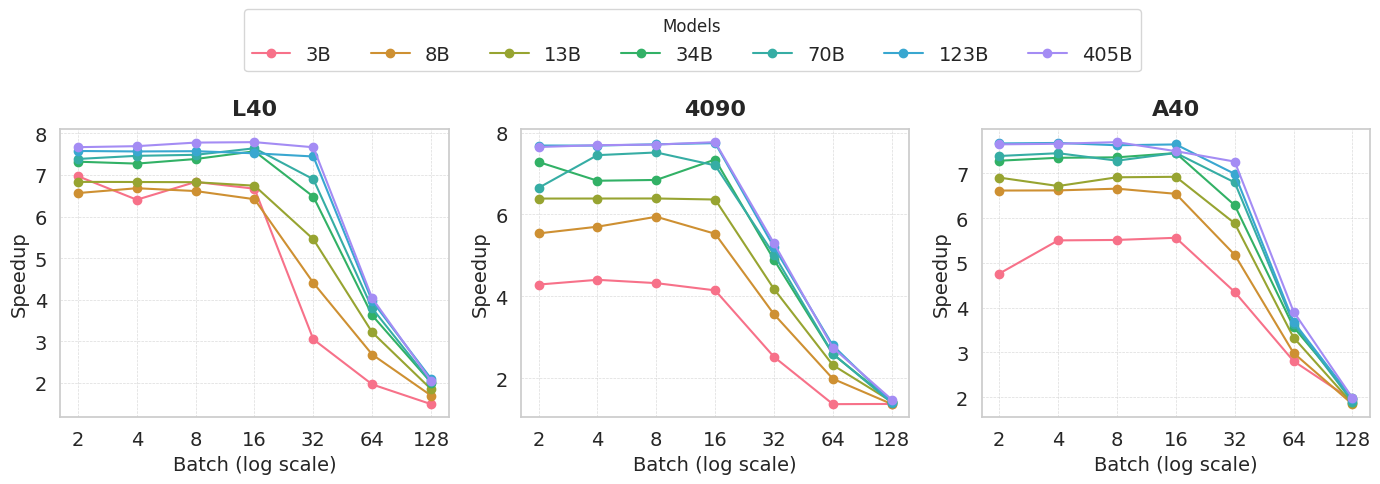} % Replace with your figure's path
    \caption{ \footnotesize {Performance evaluation of ternary layers in a transformer block, comparing TriRun with PyTorch FP16 (using CUTLASS), shows near-optimal inference speedup in high-batch settings for larger models. Each subplot corresponds to a specific Nvidia GPU. For additional results, refer to \cref{app:trirun_perfromance_benchmark}.}}
\label{fig:device_benchmark}
\end{figure*}
\vspace{0.5 em}

% \paragraph{TQ1: Implementing effective 1.6 bit for TriLMs.} 
% In our implementation, we encode \( k = 5 \) ternary digits (trits) into \( p = 8 \) bits, achieving an effective bit rate of 1.6 bits per trit. A key challenge arises in efficiently decoding these packed trits for SIMD-optimized operations. Traditional decoding methods rely on division and modulo operations, which are computationally expensive and ill-suited for vectorization. The conventional approach to decoding a packed byte \( b \) involves computing a base-3 integer \( x \) using the formula: \(
% x =  \left\lfloor\frac{ b \cdot 3^5 - (3^5 - 1) +(2^8-1) }{2^8}\right\rfloor.
% \) Each trit \( d_i \in \{-1, 0, 1\} \) is then extracted through the operation:  
% \(
% d_{i+1} = \left\lfloor \frac{x}{3^{4-i}} \right\rfloor \mod 3 \quad \text{for} \quad i = 0, \dots, 4.
% \) This method incurs high computational costs due to the repeated divisions and modulo operations, which hinder SIMD parallelism. To address these inefficiencies, we exploit the near-equivalence \( 3^5 \approx 2^8 \), enabling a multiplication-based scheme that iteratively extracts trits without explicit division or modulo operations. (See Appendix \ref{app:tq_1_implementation} for the detailed iterative procedure and its SIMD advantages.)  
\paragraph{TQ1: Implementing effective 1.6 bit for TriLMs.} 
In our implementation, we encode \( k = 5 \) ternary digits (trits) into \( p = 8 \) bits, achieving an effective bit rate of 1.6 bits per trit. A key challenge arises in efficiently decoding these packed trits for SIMD-optimized operations. Traditional decoding methods rely on division and modulo operations, which are computationally expensive and ill-suited for vectorization. The conventional approach to decoding a packed byte \( b \) involves computing a base-3 integer \( x \) using the formula: \(
x =  \left\lfloor\frac{ b \cdot 3^5 - (3^5 - 1) +(2^8-1) }{2^8}\right\rfloor.
\) Each trit \( d_i \in \{-1, 0, 1\} \) is then extracted through the operation:  
\(
d_{i+1} = \left\lfloor \frac{x}{3^{4-i}} \right\rfloor \mod 3 \quad \text{for} \quad i = 0, \dots, 4.
\)  
This method incurs high computational costs due to the repeated divisions and modulo operations, which hinder SIMD parallelism. To address these inefficiencies, we exploit the near-equivalence \( 3^5 \approx 2^8 \), enabling a multiplication-based scheme that iteratively extracts trits without explicit division or modulo operations. (See Appendix \ref{app:tq_1_implementation} for the detailed iterative procedure and its SIMD advantages.)  % \[

\paragraph{Results.} Figure ~\ref{fig:cpu_tq1_tq_2} compares token generation speeds (end-to-end and decoding) for models spanning 560M to 3.9B parameters on a Mac M4, highlighting end-to-end latency and output token generation speedup. Specifically, TQ2 outperforms other formats by utilizing 2-bit weight packing, surpassing both 4-bit quantization (as implemented in GGML) and TQ1 (1.6 bits per weight). While TQ1 requires additional fixed-point multiplication operations—resulting in slower inference compared to TQ2—it achieves a significantly smaller memory footprint (shown in \cref{fig:cpu_tq1_tq_2} on the right), making it advantageous for low-resource environments where memory storage constraints outweigh computational latency. Additional benchmarks are conducted on  AMD EPYC 7502 (see \cref{fig:cpu_tq1_tq_2_amd}) and  Apple M4 Max (14 CPU cores), with detailed results presented in ~\cref{tab:lM4_Max_14_CPU_Cores}, ~\ref{tab:benchmark_AMD_EPYC_7502} and \cref{app:cpu_bench}. These findings motivate our next step: in the following section, we introduce an optimized 2-bit packing variant designed for high-batch GPU workloads. While our current implementation demonstrates significant speedups, further refinements remain possible to enhance computational efficiency.

% \paragraph{Results.} Figure ~\ref{fig:cpu_tq1_tq_2} compares token generation speeds (end-to-end and decoding) for models spanning 560M to 3.9B parameters on a Mac M4, highlighting end-to-end latency and output token generation speedups. Specifically, TQ2 outperforms other formats by utilizing 2-bit weight packing, surpassing both 4-bit quantization (as implemented in GGML) and TQ1 (1.6 bits per weight). Although TQ1 requires additional fixed point multiplication operations - resulting in slower inference compared to TQ2, it achieves a significantly smaller memory footprint (shown in \cref{fig:cpu_tq1_tq_2} on the right). This makes TQ1 particularly beneficial in low-resource environments where memory storage constraints outweigh computational latency. Additional benchmarks are conducted on  AMD EPYC 7502 (see \cref{fig:cpu_tq1_tq_2_amd}) and  Apple M4 Max (14 CPU cores), with detailed results presented in ~\cref{tab:lM4_Max_14_CPU_Cores}, ~\ref{tab:benchmark_AMD_EPYC_7502} and \cref{app:cpu_bench}. These findings motivate our next step: in the following section, we introduce an optimized 2-bit packing variant designed for high-batch GPU workloads. While our current implementation demonstrates significant speedups, further refinements remain possible to enhance computational efficiency.

\section{TriRun: GPU Kernels for High-Batch Settings.}
\label{sec:trirun_main_section}

 LLM weight quantization leverages the fact that GPUs perform floating‐point operations much faster than they can fetch data from memory. For example, the NVIDIA L40 has a  FLOPs-to-Bytes of approximately 105 \citep{nvidia_l40s_datasheet}. In a typical matrix multiplication for mixed-precision inference in large language models, each input token requires about 2 FLOPs per weight. When weights are 2 bits, each weight occupies 0.25 bytes. During the time needed to load one such weight, the L40 can perform roughly 26 FLOPs. Since each token needs 2 FLOPs per weight, the GPU can support a critical input batch size of about $\approx$ 13 tokens. Thus, for the L40, if the input batch size is below roughly 13, memory loading becomes the bottleneck for the computation.

\subsection{Ternary Kernel Implementation}
\paragraph{Optimized Mixed-Precision Multiplication} In this work, we introduce an optimized mixed-precision matrix multiplication routine \cite{marlin} that performs FP16 × INT2 computations. In this scheme, an FP16 input matrix is multiplied by a weight matrix stored in a compact 2‐bit integer (INT2) format, wherein each 32‐bit integer encodes 16 distinct 2‐bit values. The central component of this approach is a dequantization function that employs carefully selected bit masks and a lookup‐based three‐input logical operation to extract the 2‐bit fields. This function applies a series of fused arithmetic operations to convert the packed 2‐bit data into FP16 values(see appendix \S\ref{additional_implementation_details_for_trirun}).
% Specifically, it isolates the low and high portions of the encoded integer and then uses intrinsic half‐precision subtraction and fused multiply–add operations that incorporate predetermined scaling constants and an offset tailored for symmetric quantization centered at –1. 
As a result, dequantized weight values are produced as fragments containing four FP16 numbers, which can subsequently be scaled using quantization scales stored in a separate buffer.

\begin{figure*}[ht] % or [!htbp] for more flexible positioning
    \centering
    \includegraphics[width=0.32\textwidth]{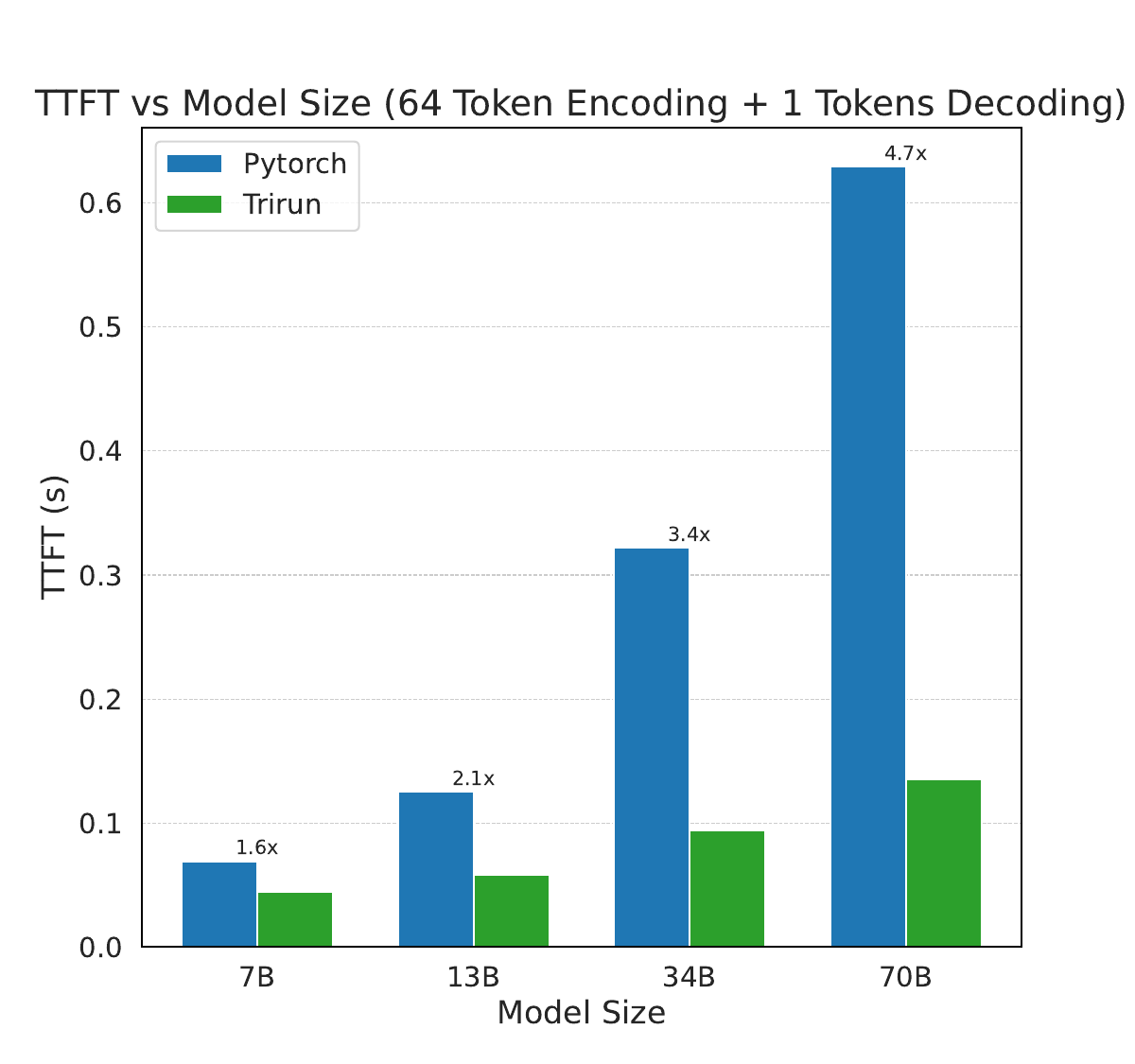} % Second figure (adjusted height)
    \includegraphics[width=0.32\textwidth]{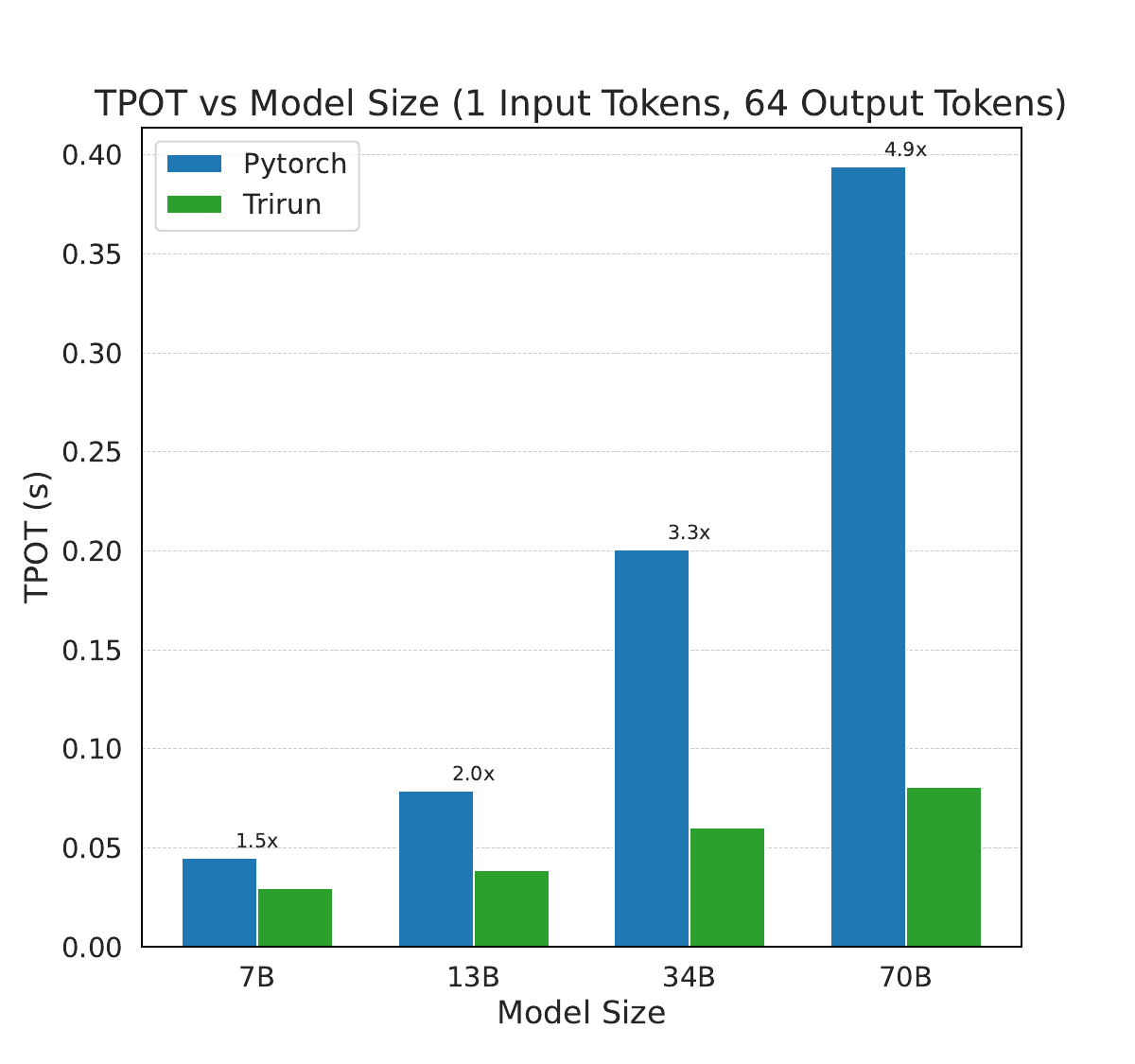} % Third figure (adjusted height)
    \includegraphics[width=0.30\textwidth]{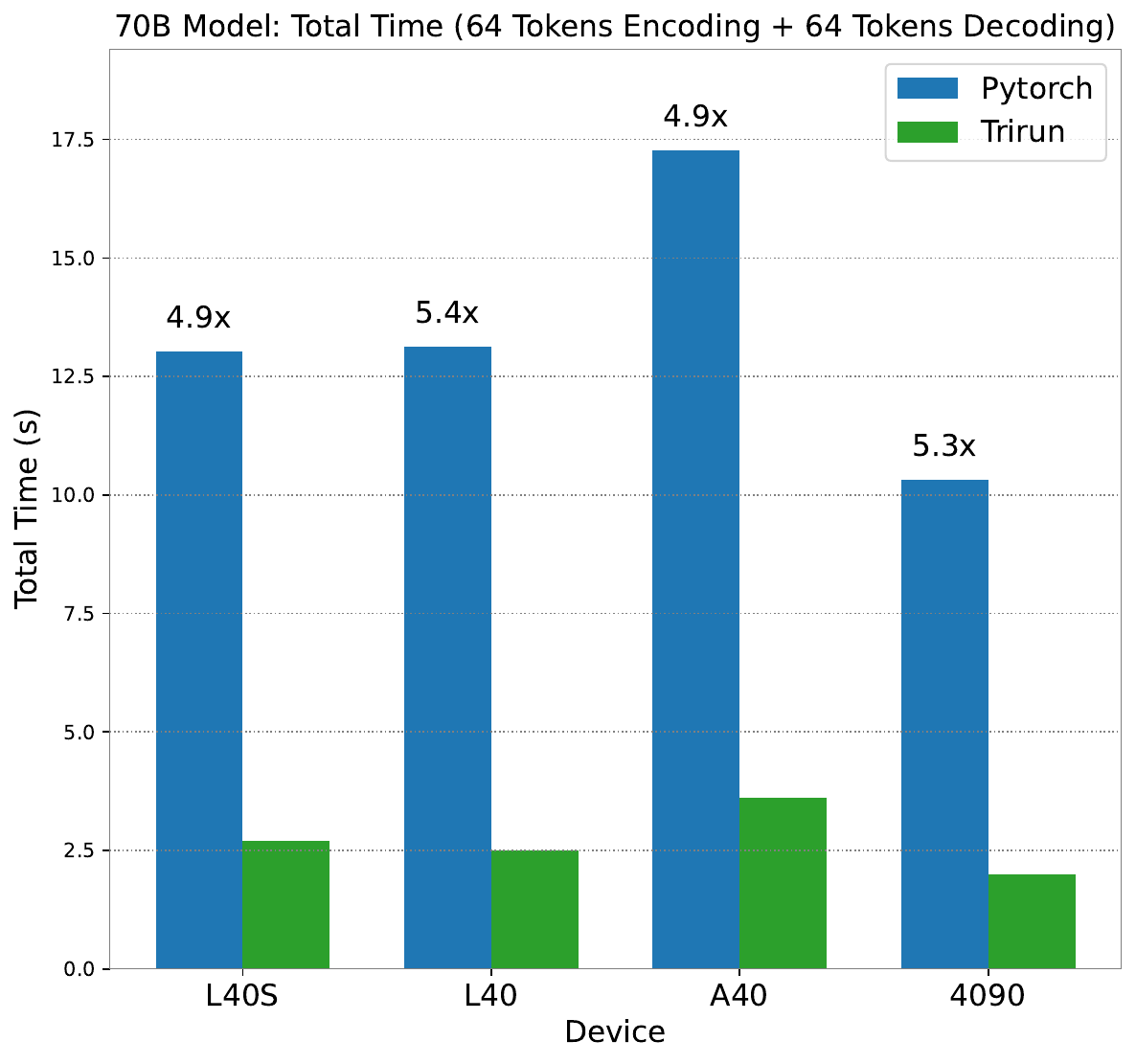} % Third figure (adjusted height)

    \caption{\footnotesize{Comparison of TriRun kernels with the FP16 PyTorch baseline on NVIDIA L40S (for more details see \cref{app:trirun_perfromance_benchmark}): \textbf{(a)} Left: Time to first token, \textbf{(b)} Center: Time per output token, \textbf{(c)} Right: Total time across different NVIDIA GPUs.}}
\label{fig:end-end_time-to-first-token_time_per_output_token}
\end{figure*}

\paragraph{GPU Performance Optimization.} On the GPU, the multiplication kernel is engineered for high performance by using asynchronous memory copy operations alongside specialized tensor core instructions available on modern NVIDIA hardware. Input fragments from the FP16 matrix are asynchronously loaded into shared memory via \textit{cp.async}\footnote{cp.async in CuPy refers to the support for asynchronous execution of GPU operations} instructions, allowing global memory accesses to overlap with computation. The kernel arranges these fragments in memory to minimize bank conflicts and maximize data reuse. Concurrently, the INT2 weight values are fetched using asynchronous copy operations that include cache hints, thereby reducing L2 cache pollution since these weights are used only once during each operation. Once the FP16 fragments and dequantized INT2 weights reside in registers, the kernel employs \textit{tensor core mma} instructions to perform efficient block‐wise multiplications. These operations accumulate the results in FP32 registers to maintain higher precision during the reduction phase before converting the final outputs back to FP16 for storage in global memory.

% \paragraph{Flexible Implementation and Data Movement.} The implementation is to allow flexible configuration of thread block dimensions, pipeline stages, and grouping parameters, ensuring adaptability to various problem sizes and hardware. Data movement from global to shared memory is managed through a double-buffering strategy, with synchronization achieved via asynchronous copy fences and explicit barrier instructions. Partial results accumulated across warps or thread blocks are reduced using a hierarchical reduction scheme that first operates within shared memory and then, if necessary, synchronizes globally across thread blocks. Finally, the computed results are reorganized into the appropriate layout and written back to global memory in FP16 format.  This approach—characterized by efficient data packing, effective asynchronous memory operations, and the exploitation of tensor core acceleration—yields a highly optimized FP16×INT2 matrix multiplication routine that is well-suited for Ternary large language models.

\paragraph{Flexible Implementation and Data Movement.} The implementation supports flexible configuration of thread block dimensions, pipeline stages, and grouping parameters for varying problem sizes and hardware. Data is moved from global to shared memory using double-buffering with asynchronous copy fences and explicit barriers. Partial results accumulated across warps or thread blocks are reduced using a hierarchical reduction scheme that first operates within shared memory and then, if necessary, synchronizes globally across thread blocks. Finally, results are reorganized and written back to global memory in FP16 format. This approach leverages efficient data packing, asynchronous memory operations, and tensor core acceleration, optimizing FP16×INT2 matrix multiplication for ternary large language models.

\subsection{Experimental Results}
\paragraph{Performance of TriRuns Kernels.}
\label{subsec:Performance_of_TriRuns_Kernels}
In  \cref{fig:device_benchmark} and  \ref{fig:TriRun_benchmark_transformer_block}, we evaluate the efficiency of TriRun kernels against PyTorch's FP16 kernels for the ternary layers within transformer modules. We benchmarked models ranging from 3 billion to 405 billion parameters across different hardwares (see \cref{tab:transformer_block_trirun performance-1}, and \ref{tab:transformer_block_trirun performance_2} for complete results). Our findings demonstrate that TriRun provides substantial performance improvements. Specifically, on an NVIDIA L40 GPU (optimized for inference) processing large matrices from a 405B parameter model, TriRun achieves a speedup of roughly 7.98x compared to FP16 when using batch sizes between 16 and 32. However, as batch sizes increase beyond this range, the speedup diminishes. This is because the computation becomes increasingly limited by the GPU's processing capabilities (compute-bound). This pattern of speedup reduction with batch size is observed across all tested GPUs. For more detailed analysis and results, refer to \cref{app:trirun_perfromance_benchmark}.

% \paragraph{End-to-End Serving Benchmark} \cref{fig:end-end_time-to-first-token_time_per_output_token} illustrates the time-to-first-token performance of TriRun with PyTorch, achieving a speedup of up to 4.7× for the 70B model (with 64 input tokens) on NVIDIA L40s. Additionally, it shows the time per output token (with 1 input and 64 output tokens), demonstrating a speedup of up to 4.9× in decoding. This trend is particularly evident with larger models, where the 70B model achieves a 4.9× end-to-end generation speedup compared to PyTorch. TriLms uses only one GPU, as opposed to the four GPUs used in the PyTorch FP16 configuration (see \cref{fig:side_by_side_graphs} on the right for more details). \cref{fig:end-end_time-to-first-token_time_per_output_token} (c) shows that these performance gains are consistent across different hardware, with a more detailed analysis provided in \cref{app:trirun_perfromance_benchmark}. Furthermore, Figure \ref{fig:speedup_plot} demonstrates that these speedups are particularly pronounced on newer consumer GPUs, as the FLOPs/byte ratio increases.

\paragraph{End-to-End Serving Benchmark} \cref{fig:end-end_time-to-first-token_time_per_output_token} illustrates the time-to-first-token performance of TriRun with PyTorch, achieving up to a 4.7× speedup on the 70B model with 64 input tokens when running on NVIDIA L40s. Additionally, it shows the time per output token (with 1 input and 64 output tokens), demonstrating a 4.9× improvement in decoding. This trend is particularly evident for larger models, where the 70B model achieves a 4.9× end-to-end generation speedup compared to PyTorch. TriRun uses only one GPU, as opposed to the four GPUs used in the PyTorch FP16 configuration (see \cref{fig:side_by_side_graphs} on the right for more details). Furthermore, \cref{fig:end-end_time-to-first-token_time_per_output_token} (c) shows that these performance gains are consistent across different NVIDIA hardware, with a more detailed analysis provided in \cref{app:trirun_perfromance_benchmark}. Finally, Figure \ref{fig:speedup_plot} demonstrates that these speedups are particularly pronounced on newer consumer GPUs, as the FLOPs/byte ratio increases.
% This comprehensive analysis encompasses both data-center-grade GPUs and prosumer-grade GPUs, providing insights into the scalability and efficiency of our approach across diverse hardware configurations. The hardware utilized for benchmarking includes NVIDIA’s A100 PCIe, A100 SMX, 3090, A40, A30, 4090, L40, L40S, and L4 GPUs.  Figure \ref{fig:device_benchmark} show that 

% Notably, a recent independent extension of MARLIN applied the method to cases where activations are quantized to 8 bits while weights are reduced to 4 bits (Zhang et al., 2024).

\section{Conclusion and Future Work}

% In this work, we addressed the growing memory bottlenecks in large language model inference by studying ternary language models (TriLMs) and proposed the strategies for efficient kernels implementation. We conducted a comprehensive scaling law analysis, revealing that TriLMs benefit significantly from scaling training data, achieving competitive performance with floating-point models for given compute budget despite their extreme quantization. Our experiments with the Spectra-1.1 family, trained on up to 1.2 trillion tokens, demonstrate sustained performance improvements, emphasizing the potential of ternary models for large-scale training and deployment. To further improve inference efficiency,  we introduced novel ternary weight packing schemes and developed optimized kernels. Our GPU kernel, TriRun, achieves up to an 8$\times$ speedup over float16 baselines in high-batch inference settings, making ternary models a viable solution for memory-constrained environments. By releasing the Spectra-1.1 models and optimized inference kernels, we aim to encourage further research on extreme low-bitwidth models and their deployment. Our results highlight the scalability and efficency of ternary models and establish a foundation for future advancements in efficient LLM inference.

In this work, we address the growing memory bottlenecks in large language model inference by studying ternary language models (TriLMs) and proposing strategies for efficient kernel implementation. We conduct a comprehensive scaling law analysis, revealing that TriLMs benefit significantly from scaling training data, achieving competitive performance with floating-point models for a given compute budget despite their extreme quantization. Our experiments with the Spectra-1.1 family, trained on up to 1.2 trillion tokens, demonstrate sustained performance improvements, emphasizing the potential of ternary models for large-scale training and deployment. To further improve inference efficiency, we introduce novel ternary weight packing schemes and develop optimized kernels. Our GPU kernel, TriRun, achieves up to an 8× speedup over float16 baselines in high-batch inference settings, making ternary models a viable solution for memory-constrained environments. By releasing the Spectra-1.1 models and optimized inference kernels, we aim to encourage further research on extreme low-bitwidth models and their deployment. Our results demonstrate the scalability and efficiency of ternary models, laying the groundwork for future advancements in efficient LLM research.

\section*{Limitations}
\label{sec:limitations}
We study the scaling law for TriLMs where we consider the dependence on number of parameters and training tokens, but do not explicitly account for the number of bits $b$ used to quantize the model. The various terms that appear in \Cref{eq:power_law} may depend non-linearly on $b$, which is an interesting direction for future work. Our pre-training scale was constrained by computational resources, and both the parameters and data need to be scaled up significantly to make TriLMs competitive with current state-of-the-art models \cite{llama3modelcard}. TriRun implements the 2-effective-bit packing scheme from \Cref{sec:2bit}. A more memory-efficient solution would involve implementing the 1.6-effective-bit packing scheme. However, due to the increased complexity of packing, the unpacking functions would require additional operations in the latter case, making it slower than TriRun. This is left as a direction for future work.

\section*{Ethics Statement}
The development of TriLMs represents a significant step toward making large-scale language models more efficient by reducing memory consumption and accelerating inference. These advancements enhance accessibility and sustainability in AI research. We advocate for openness in AI, as it drives scientific progress, fosters collaboration, and eliminates the need for re-training, which helps lower environmental impact. However, openness also presents challenges, including concerns related to privacy, security, and fairness. Despite these risks, we believe that transparency enables more effective risk mitigation by inviting diverse scrutiny and safeguards. By releasing the Spectra-1.1 suite and TriRun kernels, we aim to empower further innovation while ensuring that efficient language models serve a broad spectrum of stakeholders. As open model releases continue to gain momentum, we see this approach as the most effective way to balance progress with responsible AI development.

\section*{Acknowledgement}
We acknowledge the support from the Mozilla Responsible AI Grant, the Canada CIFAR AI Chair Program, Nolano AI and the Canada Excellence Research Chairs Program. This research was enabled by the computational resources provided by the Summit supercomputer, awarded through the Frontier DD allocation and INCITE 2023 program for the project "Scalable Foundation Models for Transferable Generalist AI" and SummitPlus allocation in 2024. These resources were supplied by the Oak Ridge Leadership Computing Facility at the Oak Ridge National Laboratory, with support from the Office of Science of the U.S. Department of Energy. 
%%%%%%%%%%%%%%%%%%%%%%% NOT INCLUDED IN PAGE LIMIT %%%%%%%%%%%%%%%%

% \newpage
% \begin{figure}[h]
%     \centering
%     \includegraphics[width=\columnwidth]{images/speedup_across_hardware.pdf}
%     \caption{Speedup Across Hardware}
%     \label{fig:speedup_plot}
% \end{figure}

%%%%%%%%%%%%%%%%%%%%%%%%%%%%%%%%%%%%%%%%%%%%%%%%%%%%%%%%%%%%

\bibliography{acl}

\begin{thebibliography}{55}
\providecommand{\natexlab}[1]{#1}

\bibitem[{{Advanced Micro Devices, Inc.}(2022)}]{amd_cdna2_whitepaper}
{Advanced Micro Devices, Inc.} 2022.
\newblock \href {https://www.amd.com/content/dam/amd/en/documents/instinct-business-docs/white-papers/amd-cdna2-white-paper.pdf} {{AMD CDNA\texttrademark 2 ARCHITECTURE}}.
\newblock White paper, Advanced Micro Devices, Inc.

\bibitem[{{Advanced Micro Devices, Inc.}(2025)}]{amd_mi250x_datasheet}
{Advanced Micro Devices, Inc.} 2025.
\newblock Amd instinct™ mi250x accelerators.
\newblock \url{https://www.amd.com/en/products/accelerators/instinct/mi200/mi250x.html}.
\newblock Accessed February 9, 2025.

\bibitem[{AI@Meta(2024)}]{llama3modelcard}
AI@Meta. 2024.
\newblock \href {https://github.com/meta-llama/llama3/blob/main/MODEL_CARD.md} {Llama 3 model card}.

\bibitem[{Ashkboos et~al.(2024)Ashkboos, Mohtashami, Croci, Li, Cameron, Jaggi, Alistarh, Hoefler, and Hensman}]{ashkboos2024quarotoutlierfree4bitinference}
Saleh Ashkboos, Amirkeivan Mohtashami, Maximilian~L. Croci, Bo~Li, Pashmina Cameron, Martin Jaggi, Dan Alistarh, Torsten Hoefler, and James Hensman. 2024.
\newblock \href {https://arxiv.org/abs/2404.00456} {Quarot: Outlier-free 4-bit inference in rotated llms}.
\newblock \emph{Preprint}, arXiv:2404.00456.

\bibitem[{Ben~Allal et~al.(2024)Ben~Allal, Lozhkov, Penedo, Wolf, and von Werra}]{benallal2024cosmopedia}
Loubna Ben~Allal, Anton Lozhkov, Guilherme Penedo, Thomas Wolf, and Leandro von Werra. 2024.
\newblock \href {https://huggingface.co/datasets/HuggingFaceTB/cosmopedia} {Cosmopedia}.

\bibitem[{Bengio et~al.(2013)Bengio, Léonard, and Courville}]{bengio2013estimatingpropagatinggradientsstochastic}
Yoshua Bengio, Nicholas Léonard, and Aaron Courville. 2013.
\newblock \href {https://arxiv.org/abs/1308.3432} {Estimating or propagating gradients through stochastic neurons for conditional computation}.
\newblock \emph{Preprint}, arXiv:1308.3432.

\bibitem[{Biderman et~al.(2023)Biderman, Schoelkopf, Anthony, Bradley, O'Brien, Hallahan, Khan, Purohit, Prashanth, Raff, Skowron, Sutawika, and van~der Wal}]{biderman2023pythiasuiteanalyzinglarge}
Stella Biderman, Hailey Schoelkopf, Quentin Anthony, Herbie Bradley, Kyle O'Brien, Eric Hallahan, Mohammad~Aflah Khan, Shivanshu Purohit, USVSN~Sai Prashanth, Edward Raff, Aviya Skowron, Lintang Sutawika, and Oskar van~der Wal. 2023.
\newblock \href {https://arxiv.org/abs/2304.01373} {Pythia: A suite for analyzing large language models across training and scaling}.
\newblock \emph{Preprint}, arXiv:2304.01373.

\bibitem[{Bisk et~al.(2019)Bisk, Zellers, Bras, Gao, and Choi}]{piqa}
Yonatan Bisk, Rowan Zellers, Ronan~Le Bras, Jianfeng Gao, and Yejin Choi. 2019.
\newblock \href {https://api.semanticscholar.org/CorpusID:208290939} {Piqa: Reasoning about physical commonsense in natural language}.
\newblock In \emph{AAAI Conference on Artificial Intelligence}.

\bibitem[{Clark et~al.(2019)Clark, Lee, Chang, Kwiatkowski, Collins, and Toutanova}]{boolq}
Christopher Clark, Kenton Lee, Ming-Wei Chang, Tom Kwiatkowski, Michael Collins, and Kristina Toutanova. 2019.
\newblock \href {https://doi.org/10.18653/v1/N19-1300} {{B}ool{Q}: Exploring the surprising difficulty of natural yes/no questions}.
\newblock In \emph{Proceedings of the 2019 Conference of the North {A}merican Chapter of the Association for Computational Linguistics: Human Language Technologies, Volume 1 (Long and Short Papers)}, pages 2924--2936, Minneapolis, Minnesota. Association for Computational Linguistics.

\bibitem[{Clark et~al.(2018)Clark, Cowhey, Etzioni, Khot, Sabharwal, Schoenick, and Tafjord}]{ARC}
Peter Clark, Isaac Cowhey, Oren Etzioni, Tushar Khot, Ashish Sabharwal, Carissa Schoenick, and Oyvind Tafjord. 2018.
\newblock \href {https://api.semanticscholar.org/CorpusID:3922816} {Think you have solved question answering? try arc, the ai2 reasoning challenge}.

\bibitem[{Clement et~al.(2019)Clement, Bierbaum, O'Keeffe, and Alemi}]{clement2019arxiv}
Colin~B. Clement, Matthew Bierbaum, Kevin~P. O'Keeffe, and Alexander~A. Alemi. 2019.
\newblock \href {https://arxiv.org/abs/1905.00075} {On the use of arxiv as a dataset}.
\newblock \emph{Preprint}, arXiv:1905.00075.

\bibitem[{Dettmers et~al.(2022{\natexlab{a}})Dettmers, Lewis, Belkada, and Zettlemoyer}]{dettmers2022llmint88bitmatrixmultiplication}
Tim Dettmers, Mike Lewis, Younes Belkada, and Luke Zettlemoyer. 2022{\natexlab{a}}.
\newblock \href {https://arxiv.org/abs/2208.07339} {Llm.int8(): 8-bit matrix multiplication for transformers at scale}.
\newblock \emph{Preprint}, arXiv:2208.07339.

\bibitem[{Dettmers et~al.(2022{\natexlab{b}})Dettmers, Lewis, Belkada, and Zettlemoyer}]{dettmers2022}
Tim Dettmers, Mike Lewis, Younes Belkada, and Luke Zettlemoyer. 2022{\natexlab{b}}.
\newblock \href {https://arxiv.org/abs/2208.07339} {Llm.int8(): 8-bit matrix multiplication for transformers at scale}.
\newblock \emph{Preprint}, arXiv:2208.07339.

\bibitem[{Dettmers and Zettlemoyer(2023)}]{dettmers2023case4bitprecisionkbit}
Tim Dettmers and Luke Zettlemoyer. 2023.
\newblock \href {https://arxiv.org/abs/2212.09720} {The case for 4-bit precision: k-bit inference scaling laws}.
\newblock \emph{Preprint}, arXiv:2212.09720.

\bibitem[{Frantar et~al.(2023)Frantar, Ashkboos, Hoefler, and Alistarh}]{gptq}
Elias Frantar, Saleh Ashkboos, Torsten Hoefler, and Dan Alistarh. 2023.
\newblock \href {https://arxiv.org/abs/2210.17323} {Gptq: Accurate post-training quantization for generative pre-trained transformers}.
\newblock \emph{Preprint}, arXiv:2210.17323.

\bibitem[{Frantar et~al.(2024)Frantar, Castro, Chen, Hoefler, and Alistarh}]{marlin}
Elias Frantar, Roberto~L. Castro, Jiale Chen, Torsten Hoefler, and Dan Alistarh. 2024.
\newblock \href {https://arxiv.org/abs/2408.11743} {Marlin: Mixed-precision auto-regressive parallel inference on large language models}.
\newblock \emph{Preprint}, arXiv:2408.11743.

\bibitem[{Gholami et~al.(2024)Gholami, Yao, Kim, Hooper, Mahoney, and Keutzer}]{gholami2024aimemorywall}
Amir Gholami, Zhewei Yao, Sehoon Kim, Coleman Hooper, Michael~W. Mahoney, and Kurt Keutzer. 2024.
\newblock \href {https://arxiv.org/abs/2403.14123} {Ai and memory wall}.
\newblock \emph{Preprint}, arXiv:2403.14123.

\bibitem[{Groeneveld et~al.(2024)Groeneveld, Beltagy, Walsh, Bhagia, Kinney, Tafjord, Jha, Ivison, Magnusson, Wang, Arora, Atkinson, Authur, Chandu, Cohan, Dumas, Elazar, Gu, Hessel, Khot, Merrill, Morrison, Muennighoff, Naik, Nam, Peters, Pyatkin, Ravichander, Schwenk, Shah, Smith, Strubell, Subramani, Wortsman, Dasigi, Lambert, Richardson, Zettlemoyer, Dodge, Lo, Soldaini, Smith, and Hajishirzi}]{groeneveld2024olmoacceleratingsciencelanguage}
Dirk Groeneveld, Iz~Beltagy, Pete Walsh, Akshita Bhagia, Rodney Kinney, Oyvind Tafjord, Ananya~Harsh Jha, Hamish Ivison, Ian Magnusson, Yizhong Wang, Shane Arora, David Atkinson, Russell Authur, Khyathi~Raghavi Chandu, Arman Cohan, Jennifer Dumas, Yanai Elazar, Yuling Gu, Jack Hessel, Tushar Khot, William Merrill, Jacob Morrison, Niklas Muennighoff, Aakanksha Naik, Crystal Nam, Matthew~E. Peters, Valentina Pyatkin, Abhilasha Ravichander, Dustin Schwenk, Saurabh Shah, Will Smith, Emma Strubell, Nishant Subramani, Mitchell Wortsman, Pradeep Dasigi, Nathan Lambert, Kyle Richardson, Luke Zettlemoyer, Jesse Dodge, Kyle Lo, Luca Soldaini, Noah~A. Smith, and Hannaneh Hajishirzi. 2024.
\newblock \href {https://arxiv.org/abs/2402.00838} {Olmo: Accelerating the science of language models}.
\newblock \emph{Preprint}, arXiv:2402.00838.

\bibitem[{He et~al.(2024)He, Zhou, Huang, Li, Wang, Guo, Meng, Gui, Yu, and Xie}]{he2024inferenceperformanceoptimizationlarge}
Pujiang He, Shan Zhou, Wenhuan Huang, Changqing Li, Duyi Wang, Bin Guo, Chen Meng, Sheng Gui, Weifei Yu, and Yi~Xie. 2024.
\newblock \href {https://arxiv.org/abs/2407.07304} {Inference performance optimization for large language models on cpus}.
\newblock \emph{Preprint}, arXiv:2407.07304.

\bibitem[{Hendrycks et~al.(2021)Hendrycks, Burns, Basart, Zou, Mazeika, Song, and Steinhardt}]{mmlu}
Dan Hendrycks, Collin Burns, Steven Basart, Andy Zou, Mantas Mazeika, Dawn Song, and Jacob Steinhardt. 2021.
\newblock Measuring massive multitask language understanding.
\newblock \emph{Proceedings of the International Conference on Learning Representations (ICLR)}.

\bibitem[{Hoffmann et~al.(2022)Hoffmann, Borgeaud, Mensch, Buchatskaya, Cai, Rutherford, de~Las~Casas, Hendricks, Welbl, Clark, Hennigan, Noland, Millican, van~den Driessche, Damoc, Guy, Osindero, Simonyan, Elsen, Rae, Vinyals, and Sifre}]{hoffmann2022trainingcomputeoptimallargelanguage}
Jordan Hoffmann, Sebastian Borgeaud, Arthur Mensch, Elena Buchatskaya, Trevor Cai, Eliza Rutherford, Diego de~Las~Casas, Lisa~Anne Hendricks, Johannes Welbl, Aidan Clark, Tom Hennigan, Eric Noland, Katie Millican, George van~den Driessche, Bogdan Damoc, Aurelia Guy, Simon Osindero, Karen Simonyan, Erich Elsen, Jack~W. Rae, Oriol Vinyals, and Laurent Sifre. 2022.
\newblock \href {https://arxiv.org/abs/2203.15556} {Training compute-optimal large language models}.
\newblock \emph{Preprint}, arXiv:2203.15556.

\bibitem[{Jiang et~al.(2024)Jiang, Sablayrolles, Roux, Mensch, Savary, Bamford, Chaplot, de~las Casas, Hanna, Bressand, Lengyel, Bour, Lample, Lavaud, Saulnier, Lachaux, Stock, Subramanian, Yang, Antoniak, Scao, Gervet, Lavril, Wang, Lacroix, and Sayed}]{jiang2024mixtralexperts}
Albert~Q. Jiang, Alexandre Sablayrolles, Antoine Roux, Arthur Mensch, Blanche Savary, Chris Bamford, Devendra~Singh Chaplot, Diego de~las Casas, Emma~Bou Hanna, Florian Bressand, Gianna Lengyel, Guillaume Bour, Guillaume Lample, Lélio~Renard Lavaud, Lucile Saulnier, Marie-Anne Lachaux, Pierre Stock, Sandeep Subramanian, Sophia Yang, Szymon Antoniak, Teven~Le Scao, Théophile Gervet, Thibaut Lavril, Thomas Wang, Timothée Lacroix, and William~El Sayed. 2024.
\newblock \href {https://arxiv.org/abs/2401.04088} {Mixtral of experts}.
\newblock \emph{Preprint}, arXiv:2401.04088.

\bibitem[{Joshi et~al.(2017)Joshi, Choi, Weld, and Zettlemoyer}]{triviaqa}
Mandar Joshi, Eunsol Choi, Daniel Weld, and Luke Zettlemoyer. 2017.
\newblock \href {https://doi.org/10.18653/v1/P17-1147} {{T}rivia{QA}: A large scale distantly supervised challenge dataset for reading comprehension}.
\newblock In \emph{Proceedings of the 55th Annual Meeting of the Association for Computational Linguistics (Volume 1: Long Papers)}, pages 1601--1611, Vancouver, Canada. Association for Computational Linguistics.

\bibitem[{Kalamkar et~al.(2019)Kalamkar, Mudigere, Mellempudi, Das, Banerjee, Avancha, Vooturi, Jammalamadaka, Huang, Yuen, Yang, Park, Heinecke, Georganas, Srinivasan, Kundu, Smelyanskiy, Kaul, and Dubey}]{kalamkar2019studybfloat16deeplearning}
Dhiraj Kalamkar, Dheevatsa Mudigere, Naveen Mellempudi, Dipankar Das, Kunal Banerjee, Sasikanth Avancha, Dharma~Teja Vooturi, Nataraj Jammalamadaka, Jianyu Huang, Hector Yuen, Jiyan Yang, Jongsoo Park, Alexander Heinecke, Evangelos Georganas, Sudarshan Srinivasan, Abhisek Kundu, Misha Smelyanskiy, Bharat Kaul, and Pradeep Dubey. 2019.
\newblock \href {https://arxiv.org/abs/1905.12322} {A study of bfloat16 for deep learning training}.
\newblock \emph{Preprint}, arXiv:1905.12322.

\bibitem[{Kaplan et~al.(2020)Kaplan, McCandlish, Henighan, Brown, Chess, Child, Gray, Radford, Wu, and Amodei}]{kaplan2020scalinglawsneurallanguage}
Jared Kaplan, Sam McCandlish, Tom Henighan, Tom~B. Brown, Benjamin Chess, Rewon Child, Scott Gray, Alec Radford, Jeffrey Wu, and Dario Amodei. 2020.
\newblock \href {https://arxiv.org/abs/2001.08361} {Scaling laws for neural language models}.
\newblock \emph{Preprint}, arXiv:2001.08361.

\bibitem[{Kaushal et~al.(2024)Kaushal, Vaidhya, Mondal, Pandey, Bhagat, and Rish}]{spectra}
Ayush Kaushal, Tejas Vaidhya, Arnab~Kumar Mondal, Tejas Pandey, Aaryan Bhagat, and Irina Rish. 2024.
\newblock \href {https://arxiv.org/abs/2407.12327} {Spectra: Surprising effectiveness of pretraining ternary language models at scale}.
\newblock \emph{Preprint}, arXiv:2407.12327.

\bibitem[{Lin et~al.(2024)Lin, Tang, Tang, Yang, Chen, Wang, Xiao, Dang, Gan, and Han}]{awq}
Ji~Lin, Jiaming Tang, Haotian Tang, Shang Yang, Wei-Ming Chen, Wei-Chen Wang, Guangxuan Xiao, Xingyu Dang, Chuang Gan, and Song Han. 2024.
\newblock \href {https://arxiv.org/abs/2306.00978} {Awq: Activation-aware weight quantization for llm compression and acceleration}.
\newblock \emph{Preprint}, arXiv:2306.00978.

\bibitem[{Liu et~al.(2021)Liu, Cui, Liu, Huang, Wang, and Zhang}]{logiqa}
Jian Liu, Leyang Cui, Hanmeng Liu, Dandan Huang, Yile Wang, and Yue Zhang. 2021.
\newblock Logiqa: a challenge dataset for machine reading comprehension with logical reasoning.
\newblock In \emph{Proceedings of the Twenty-Ninth International Joint Conference on Artificial Intelligence}, IJCAI'20.

\bibitem[{Lo et~al.(2020)Lo, Wang, Neumann, Kinney, and Weld}]{lo-etal-2020-s2orc}
Kyle Lo, Lucy~Lu Wang, Mark Neumann, Rodney Kinney, and Daniel Weld. 2020.
\newblock \href {https://doi.org/10.18653/v1/2020.acl-main.447} {{S}2{ORC}: The semantic scholar open research corpus}.
\newblock In \emph{Proceedings of the 58th Annual Meeting of the Association for Computational Linguistics}, pages 4969--4983, Online. Association for Computational Linguistics.

\bibitem[{Loshchilov and Hutter(2019)}]{loshchilov2019decoupledweightdecayregularization}
Ilya Loshchilov and Frank Hutter. 2019.
\newblock \href {https://arxiv.org/abs/1711.05101} {Decoupled weight decay regularization}.
\newblock \emph{Preprint}, arXiv:1711.05101.

\bibitem[{Lozhkov et~al.(2024)Lozhkov, Ben~Allal, von Werra, and Wolf}]{lozhkov2024fineweb-edu}
Anton Lozhkov, Loubna Ben~Allal, Leandro von Werra, and Thomas Wolf. 2024.
\newblock \href {https://doi.org/10.57967/hf/2497} {Fineweb-edu: the finest collection of educational content}.

\bibitem[{Ma et~al.(2024)Ma, Wang, Ma, Wang, Wang, Huang, Dong, Wang, Xue, and Wei}]{bitnet_1_58}
Shuming Ma, Hongyu Wang, Lingxiao Ma, Lei Wang, Wenhui Wang, Shaohan Huang, Li~Dong, Ruiping Wang, Jilong Xue, and Furu Wei. 2024.
\newblock \href {https://arxiv.org/abs/2402.17764} {The era of 1-bit llms: All large language models are in 1.58 bits}.
\newblock \emph{Preprint}, arXiv:2402.17764.

\bibitem[{Micikevicius et~al.(2018)Micikevicius, Narang, Alben, Diamos, Elsen, Garcia, Ginsburg, Houston, Kuchaiev, Venkatesh, and Wu}]{micikevicius2018mixedprecisiontraining}
Paulius Micikevicius, Sharan Narang, Jonah Alben, Gregory Diamos, Erich Elsen, David Garcia, Boris Ginsburg, Michael Houston, Oleksii Kuchaiev, Ganesh Venkatesh, and Hao Wu. 2018.
\newblock \href {https://arxiv.org/abs/1710.03740} {Mixed precision training}.
\newblock \emph{Preprint}, arXiv:1710.03740.

\bibitem[{Paperno et~al.(2016)Paperno, Kruszewski, Lazaridou, Pham, Bernardi, Pezzelle, Baroni, Boleda, and Fern{\'a}ndez}]{lambada}
Denis Paperno, Germ{\'a}n Kruszewski, Angeliki Lazaridou, Ngoc~Quan Pham, Raffaella Bernardi, Sandro Pezzelle, Marco Baroni, Gemma Boleda, and Raquel Fern{\'a}ndez. 2016.
\newblock \href {https://doi.org/10.18653/v1/P16-1144} {The {LAMBADA} dataset: Word prediction requiring a broad discourse context}.
\newblock In \emph{Proceedings of the 54th Annual Meeting of the Association for Computational Linguistics (Volume 1: Long Papers)}, pages 1525--1534, Berlin, Germany. Association for Computational Linguistics.

\bibitem[{Radford et~al.(2019)Radford, Wu, Child, Luan, Amodei, and Sutskever}]{gpt2}
Alec Radford, Jeff Wu, Rewon Child, David Luan, Dario Amodei, and Ilya Sutskever. 2019.
\newblock \href {https://api.semanticscholar.org/CorpusID:160025533} {Language models are unsupervised multitask learners}.

\bibitem[{Rajbhandari et~al.(2020)Rajbhandari, Rasley, Ruwase, and He}]{Samyam_Rajbhandari_zero_2020}
Samyam Rajbhandari, Jeff Rasley, Olatunji Ruwase, and Yuxiong He. 2020.
\newblock \href {https://arxiv.org/abs/1910.02054} {Zero: Memory optimizations toward training trillion parameter models}.
\newblock \emph{Preprint}, arXiv:1910.02054.

\bibitem[{Sakaguchi et~al.(2021)Sakaguchi, Bras, Bhagavatula, and Choi}]{winogrande}
Keisuke Sakaguchi, Ronan~Le Bras, Chandra Bhagavatula, and Yejin Choi. 2021.
\newblock \href {https://doi.org/10.1145/3474381} {Winogrande: an adversarial winograd schema challenge at scale}.
\newblock \emph{Commun. ACM}, 64(9):99–106.

\bibitem[{Sardana et~al.(2024)Sardana, Portes, Doubov, and Frankle}]{sardana2024chinchillaoptimalaccountinginferencelanguage}
Nikhil Sardana, Jacob Portes, Sasha Doubov, and Jonathan Frankle. 2024.
\newblock \href {https://arxiv.org/abs/2401.00448} {Beyond chinchilla-optimal: Accounting for inference in language model scaling laws}.
\newblock \emph{Preprint}, arXiv:2401.00448.

\bibitem[{Shazeer(2020)}]{shazeer2020gluvariantsimprovetransformer}
Noam Shazeer. 2020.
\newblock \href {https://arxiv.org/abs/2002.05202} {Glu variants improve transformer}.
\newblock \emph{Preprint}, arXiv:2002.05202.

\bibitem[{Shen et~al.(2024)Shen, Tao, Ma, Neiswanger, Liu, Wang, Tan, Hestness, Vassilieva, Soboleva, and Xing}]{shen2024slimpajamadcunderstandingdatacombinations}
Zhiqiang Shen, Tianhua Tao, Liqun Ma, Willie Neiswanger, Zhengzhong Liu, Hongyi Wang, Bowen Tan, Joel Hestness, Natalia Vassilieva, Daria Soboleva, and Eric Xing. 2024.
\newblock \href {https://arxiv.org/abs/2309.10818} {Slimpajama-dc: Understanding data combinations for llm training}.
\newblock \emph{Preprint}, arXiv:2309.10818.

\bibitem[{Sheng et~al.(2023)Sheng, Zheng, Yuan, Li, Ryabinin, Fu, Xie, Chen, Barrett, Gonzalez, Liang, Ré, Stoica, and Zhang}]{flex}
Ying Sheng, Lianmin Zheng, Binhang Yuan, Zhuohan Li, Max Ryabinin, Daniel~Y. Fu, Zhiqiang Xie, Beidi Chen, Clark Barrett, Joseph~E. Gonzalez, Percy Liang, Christopher Ré, Ion Stoica, and Ce~Zhang. 2023.
\newblock \href {https://arxiv.org/abs/2303.06865} {Flexgen: High-throughput generative inference of large language models with a single gpu}.
\newblock \emph{Preprint}, arXiv:2303.06865.

\bibitem[{Singh et~al.(2024)Singh, Co-Reyes, Agarwal, Anand, Patil, Garcia, Liu, Harrison, Lee, Xu, Parisi, Kumar, Alemi, Rizkowsky, Nova, Adlam, Bohnet, Elsayed, Sedghi, Mordatch, Simpson, Gur, Snoek, Pennington, Hron, Kenealy, Swersky, Mahajan, Culp, Xiao, Bileschi, Constant, Novak, Liu, Warkentin, Qian, Bansal, Dyer, Neyshabur, Sohl-Dickstein, and Fiedel}]{singh2024humandatascalingselftraining}
Avi Singh, John~D. Co-Reyes, Rishabh Agarwal, Ankesh Anand, Piyush Patil, Xavier Garcia, Peter~J. Liu, James Harrison, Jaehoon Lee, Kelvin Xu, Aaron Parisi, Abhishek Kumar, Alex Alemi, Alex Rizkowsky, Azade Nova, Ben Adlam, Bernd Bohnet, Gamaleldin Elsayed, Hanie Sedghi, Igor Mordatch, Isabelle Simpson, Izzeddin Gur, Jasper Snoek, Jeffrey Pennington, Jiri Hron, Kathleen Kenealy, Kevin Swersky, Kshiteej Mahajan, Laura Culp, Lechao Xiao, Maxwell~L. Bileschi, Noah Constant, Roman Novak, Rosanne Liu, Tris Warkentin, Yundi Qian, Yamini Bansal, Ethan Dyer, Behnam Neyshabur, Jascha Sohl-Dickstein, and Noah Fiedel. 2024.
\newblock \href {https://arxiv.org/abs/2312.06585} {Beyond human data: Scaling self-training for problem-solving with language models}.
\newblock \emph{Preprint}, arXiv:2312.06585.

\bibitem[{Soldaini and Lo(2023)}]{peS2o}
Luca Soldaini and Kyle Lo. 2023.
\newblock {peS2o (Pretraining Efficiently on S2ORC) Dataset}.
\newblock Technical report, {Allen Institute for AI}.
\newblock ODC-By, \url{https://github.com/allenai/pes2o}.

\bibitem[{Su et~al.(2023)Su, Lu, Pan, Murtadha, Wen, and Liu}]{su2023roformerenhancedtransformerrotary}
Jianlin Su, Yu~Lu, Shengfeng Pan, Ahmed Murtadha, Bo~Wen, and Yunfeng Liu. 2023.
\newblock \href {https://arxiv.org/abs/2104.09864} {Roformer: Enhanced transformer with rotary position embedding}.
\newblock \emph{Preprint}, arXiv:2104.09864.

\bibitem[{Technologies(2023)}]{nvidia_l40s_datasheet}
PNY Technologies. 2023.
\newblock \href {https://www.pny.com/en-eu/File%20Library/Professional/DATASHEET/DATA%20CENTER%20CARDS/PNY-NVIDIA-L40S-Datasheet.pdf} {Nvidia l40s datasheet}.
\newblock Accessed: 2025-02-12.

\bibitem[{Tokpanov et~al.(2024)Tokpanov, Millidge, Glorioso, Pilault, Ibrahim, Whittington, and Anthony}]{tokpanov2024zyda}
Yury Tokpanov, Beren Millidge, Paolo Glorioso, Jonathan Pilault, Adam Ibrahim, James Whittington, and Quentin Anthony. 2024.
\newblock \href {https://arxiv.org/abs/2406.01981} {Zyda: A 1.3t dataset for open language modeling}.
\newblock \emph{Preprint}, arXiv:2406.01981.

\bibitem[{Touvron et~al.(2023)Touvron, Lavril, Izacard, Martinet, Lachaux, Lacroix, Rozière, Goyal, Hambro, Azhar, Rodriguez, Joulin, Grave, and Lample}]{llama}
Hugo Touvron, Thibaut Lavril, Gautier Izacard, Xavier Martinet, Marie-Anne Lachaux, Timothée Lacroix, Baptiste Rozière, Naman Goyal, Eric Hambro, Faisal Azhar, Aurelien Rodriguez, Armand Joulin, Edouard Grave, and Guillaume Lample. 2023.
\newblock \href {https://arxiv.org/abs/2302.13971} {Llama: Open and efficient foundation language models}.
\newblock \emph{Preprint}, arXiv:2302.13971.

\bibitem[{Vaswani et~al.(2023)Vaswani, Shazeer, Parmar, Uszkoreit, Jones, Gomez, Kaiser, and Polosukhin}]{vaswani2023attentionneed}
Ashish Vaswani, Noam Shazeer, Niki Parmar, Jakob Uszkoreit, Llion Jones, Aidan~N. Gomez, Lukasz Kaiser, and Illia Polosukhin. 2023.
\newblock \href {https://arxiv.org/abs/1706.03762} {Attention is all you need}.
\newblock \emph{Preprint}, arXiv:1706.03762.

\bibitem[{Wang et~al.(2023)Wang, Ma, Dong, Huang, Wang, Ma, Yang, Wang, Wu, and Wei}]{bitnet}
Hongyu Wang, Shuming Ma, Li~Dong, Shaohan Huang, Huaijie Wang, Lingxiao Ma, Fan Yang, Ruiping Wang, Yi~Wu, and Furu Wei. 2023.
\newblock \href {https://arxiv.org/abs/2310.11453} {Bitnet: Scaling 1-bit transformers for large language models}.
\newblock \emph{Preprint}, arXiv:2310.11453.

\bibitem[{Wei et~al.(2023)Wei, Wang, Schuurmans, Bosma, Ichter, Xia, Chi, Le, and Zhou}]{wei2023chainofthoughtpromptingelicitsreasoning}
Jason Wei, Xuezhi Wang, Dale Schuurmans, Maarten Bosma, Brian Ichter, Fei Xia, Ed~Chi, Quoc Le, and Denny Zhou. 2023.
\newblock \href {https://arxiv.org/abs/2201.11903} {Chain-of-thought prompting elicits reasoning in large language models}.
\newblock \emph{Preprint}, arXiv:2201.11903.

\bibitem[{Welbl et~al.(2017)Welbl, Liu, and Gardner}]{sciq}
Johannes Welbl, Nelson~F. Liu, and Matt Gardner. 2017.
\newblock \href {https://api.semanticscholar.org/CorpusID:1553193} {Crowdsourcing multiple choice science questions}.
\newblock \emph{ArXiv}, abs/1707.06209.

\bibitem[{Xiao et~al.(2024)Xiao, Lin, Seznec, Wu, Demouth, and Han}]{xiao2024smoothquant}
Guangxuan Xiao, Ji~Lin, Mickael Seznec, Hao Wu, Julien Demouth, and Song Han. 2024.
\newblock \href {https://arxiv.org/abs/2211.10438} {Smoothquant: Accurate and efficient post-training quantization for large language models}.
\newblock \emph{Preprint}, arXiv:2211.10438.

\bibitem[{Zellers et~al.(2019)Zellers, Holtzman, Bisk, Farhadi, and Choi}]{hellaswag}
Rowan Zellers, Ari Holtzman, Yonatan Bisk, Ali Farhadi, and Yejin Choi. 2019.
\newblock \href {https://doi.org/10.18653/v1/P19-1472} {{H}ella{S}wag: Can a machine really finish your sentence?}
\newblock In \emph{Proceedings of the 57th Annual Meeting of the Association for Computational Linguistics}, pages 4791--4800, Florence, Italy. Association for Computational Linguistics.

\bibitem[{Zhang et~al.(2022)Zhang, Zhang, Li, and Smola}]{zhang2022automaticchainthoughtprompting}
Zhuosheng Zhang, Aston Zhang, Mu~Li, and Alex Smola. 2022.
\newblock \href {https://arxiv.org/abs/2210.03493} {Automatic chain of thought prompting in large language models}.
\newblock \emph{Preprint}, arXiv:2210.03493.

\bibitem[{Zhou et~al.(2024)Zhou, Ning, Hong, Fu, Xu, Li, Lou, Wang, Yuan, Li, Yan, Dai, Zhang, Dong, and Wang}]{zhou2024surveyefficientinferencelarge}
Zixuan Zhou, Xuefei Ning, Ke~Hong, Tianyu Fu, Jiaming Xu, Shiyao Li, Yuming Lou, Luning Wang, Zhihang Yuan, Xiuhong Li, Shengen Yan, Guohao Dai, Xiao-Ping Zhang, Yuhan Dong, and Yu~Wang. 2024.
\newblock \href {https://arxiv.org/abs/2404.14294} {A survey on efficient inference for large language models}.
\newblock \emph{Preprint}, arXiv:2404.14294.

\end{thebibliography}
% \bibliographystyle{icml2025}

%%%%%%%%%%%%%%%%%%%%%%%%%%%%%%%%%%%%%%%%%%%%%%%%%%%%%%%%%%%%%%%%%%%%%%%%%%%%%%%
%%%%%%%%%%%%%%%%%%%%%%%%%%%%%%%%%%%%%%%%%%%%%%%%%%%%%%%%%%%%%%%%%%%%%%%%%%%%%%%
% APPENDIX
%%%%%%%%%%%%%%%%%%%%%%%%%%%%%%%%%%%%%%%%%%%%%%%%%%%%%%%%%%%%%%%%%%%%%%%%%%%%%%%
%%%%%%%%%%%%%%%%%%%%%%%%%%%%%%%%%%%%%%%%%%%%%%%%%%%%%%%%%%%%%%%%%%%%%%%%%%%%%%%
\appendix
% \onecolumn

\section{Related Work}
\paragraph{Training LLMs in low precision}

Large language models such as GPT \citep{gpt2}, OLMo \citep{groeneveld2024olmoacceleratingsciencelanguage}, and the LLaMA family \citep{llama} have traditionally relied on mixed precision (FP32/FP16 or FP32/BF16) \citep{micikevicius2018mixedprecisiontraining} and half-precision (FP16/BF16) \citep{kalamkar2019studybfloat16deeplearning} to optimize computational efficiency. More recent advancements in extreme quantization have introduced ternary and binary network paradigms \citep{spectra, bitnet}, which leverage quantization-aware training (QAT) for efficient low-bitwidth model representations. These models maintain higher-precision latent (or master) weights, such as FP16, to stabilize training while dynamically binarizing or ternarizing weights during inference. The straight-through estimator (STE) \citep{bengio2013estimatingpropagatinggradientsstochastic} is commonly employed to facilitate gradient-based updates. The Spectra suite \cite{spectra} provides a comprehensive study of ternary, quantized, and FP16 language models, offering insights into the performance and scaling trends of low-bitwidth models.

\paragraph{Advancements in Post-Training Quantization}
Post-training quantization (PTQ) remains a crucial approach for reducing LLM memory and compute requirements without requiring retraining. Techniques such as SmoothQuant \citep{xiao2024smoothquant} and QuaRot \citep{ashkboos2024quarotoutlierfree4bitinference} address challenges associated with activation quantization, particularly mitigating large activation outliers \citep{dettmers2022}. While these methods improve compression, they often rely on 8-bit quantization to preserve numerical stability. Continued research into activation-aware quantization techniques is vital for further enhancing LLM deployment in resource-constrained environments.

\begin{table*}[!t]
\centering
\small
\renewcommand{\arraystretch}{1.2}
\begin{tabular}{l c c}
\toprule
\textbf{Dataset Name} & \textbf{Number of Tokens (Billion)} & \textbf{Percentage} \\ 
\midrule
\textbf{ArXiv} \cite{clement2019arxiv} & 3.67 & 0.31\% \\ 
\textbf{Cosmopedia-v2} \cite{benallal2024cosmopedia} & 22.36 & 1.86\% \\ 
\textbf{PeS2o} \cite{peS2o} & 42.70 & 3.56\% \\ 
% \textbf{Zyda-StarCoder-Git-Commits} \cite{tokpanov2024zyda} & N/A & N/A \\ 
% \textbf{Zyda-StarCoder-Languages} & N/A & N/A \\ 
\textbf{FineWeb-Edu} \cite{lozhkov2024fineweb-edu} & 960.42 & 80.04\% \\ 
\textbf{Zyda - StarCoder \cite{tokpanov2024zyda}} & 170.85 & 14.24\% \\ 
\midrule
\textbf{Total} & 1200.00 & 100.00\% \\ 
\bottomrule
\end{tabular}
\caption{\footnotesize Pretraining datasets and token counts for Spectra-1.1 models.}
\label{tab:datasets}
\end{table*}

\paragraph{Optimizing Inference Efficiency}
To improve LLM deployment efficiency, frameworks like MARLIN \citep{marlin} initially implemented GPTQ-based quantization, enabling accelerated inference.MARLIN kernels combine various techniques, ranging from advanced
task scheduling, partitioning, and pipeplining techniques to quantization-specific layout and compute optimizations. More recently, MARLIN has been extended to incorporate Activation-Weight Quantization (AWQ) \citep{awq}, a technique that jointly quantizes both weights and activations to mitigate accuracy degradation in low-bitwidth settings. 

\section{Pretraining Details}
\label{app:pretraining_details}
\subsection{Quantized Linear Layer: Forward, Backward, and Inference Stages}
We now present the mathematical formulation for a linear layer employing the TriLM quantization scheme \cite{spectra}, outlining the processes for the forward pass, backward pass, and inference stages.

\paragraph{Forward Pass.} In the forward pass, we begin by calculating the scaling factor $\gamma$ to normalize the weight matrix $W$. The scaling factor is given by:

\[
\gamma = \epsilon + \frac{1}{nm} \sum_{i=1}^{n} \sum_{j=1}^{m} |W_{ij}|
\]

where $n$ and $m$ denote the dimensions of the weight matrix $W$, and $\epsilon$ is a small constant added for numerical stability.

Subsequently, the weight matrix $W$ is quantized by rounding its entries to the nearest value in the set $\{-1, 0, 1\}$, scaled by $\gamma$:

\[
\widehat{W_{ij}} = \text{round}\left(\min\left(\max\left(\frac{W_{ij}}{\gamma}, -1\right), 1\right)\right)
\]

The quantized weight matrix $\widetilde{W}$ is then obtained by scaling the rounded weights: $ \widetilde{W_{ij}} = \gamma \widehat{W_{ij}} $

Finally, the output $Y$ is computed as the product of the input $X$ and the transposed quantized weight matrix: $ Y = X \widetilde{W}^T$

\paragraph{Backward Pass.} During the backward pass, the gradients of the loss function $L$ with respect to the input $X$ and the weight matrix $W$ are computed. These gradients are given by:

\[
\frac{\partial L}{\partial X} = \frac{\partial L}{\partial Y} \widetilde{W}
\]

\[
\frac{\partial L}{\partial W} = \frac{\partial L}{\partial Y}^T X
\]

\paragraph{Inference.}

For inference, the quantized weight matrix $\widehat{W}$ and the scaling factor $\gamma$ are precomputed and cached to reduce computation during prediction. The steps are as follows:
\begin{enumerate}
    \item Compute $\widehat{W}$ and $\gamma$ once and store them.
    \item Use the precomputed values to calculate the quantized weight matrix:
    $
    \widetilde{W_{ij}} = \gamma \widehat{W_{ij}}
   $
    \item Finally, the output $Y$ is computed as:
    $
    Y = X \widetilde{W}^T
   $
\end{enumerate}
By caching the scaling factor and the quantized weights, the inference process is significantly accelerated, as it eliminates the need for redundant recalculations.

\subsection{Dataset}
\label{pretraining_data}

Our training corpus comprises a diverse mix of data from publicly available sources. To scale TriLMs, we trained on approximately 1.2 trillion tokens, up-sampling the most factual sources to enhance the model’s knowledge while reducing hallucinations. The details of the datasets used are summarized in Table \ref{tab:datasets}. Each dataset was preprocessed and tokenized using llama2 tokenizer \citep{llama3modelcard}.

\begin{itemize}

\item \textbf{ArXiv} \cite{clement2019arxiv}: The dataset comprises 1.5 million arXiv preprint articles from fields such as Physics, Mathematics, and Computer Science, encompassing text, figures, authors, citations, and metadata.

\item \textbf{Cosmopedia-v2} \cite{benallal2024cosmopedia}: A synthetic dataset of over 30 million documents and 25 billion tokens. The dataset was generated using the Mixtral-8x7B-Instruct-v0.1 model, a multi-expert language model introduced in \citep{jiang2024mixtralexperts}, designed for high-quality content generation. It is one of the largest publicly available synthetic datasets.

\item \textbf{PeS2o} \cite{peS2o}: It comprises 40 million open-access academic papers that have been cleaned, filtered, and formatted specifically for the pre-training of language models. It is derived from the Semantic Scholar Open Research Corpus \citep{lo-etal-2020-s2orc}.

% \begin{table}[ht]
% \centering
% \small
% \renewcommand{\arraystretch}{1.2}
% \begin{tabular}{l c c}
% \toprule
% \textbf{Dataset Name} & \textbf{Number of Tokens (Billion)} & \textbf{Percentage} \\ 
% \midrule
% \textbf{ArXiv} \cite{clement2019arxiv} & 3.67 & 0.306\% \\ 
% \textbf{Cosmopedia-v2} \cite{benallal2024cosmopedia} & 18.73 & 1.561\% \\ 
% \textbf{PeS2o} \cite{peS2o} & 42.70 & 3.558\% \\ 
% \textbf{Zyda-StarCoder-Git-Commits} \cite{tokpanov2024zyda} & 6.53 & 0.544\% \\ 
% \textbf{Zyda-StarCoder-Languages} & 503.27 & 41.939\% \\ 
% \textbf{FineWeb-Edu} \cite{lozhkov2024fineweb-edu} & 960.42 & 80.035\% \\ 
% \midrule
% \textbf{Total} & 1200.00 & 100.000\% \\ 
% \bottomrule
% \end{tabular}
% \caption{\footnotesize Pretraining datasets and token counts for Spectra-1.1 models.}
% \label{tab:datasets}
% \end{table}

% \begin{table}[ht]
% \centering
% \small
% \renewcommand{\arraystretch}{1.2}
% \begin{tabular}{l c}
% \toprule
% \textbf{Dataset Name} & \textbf{Number of Tokens (Billion)} \\ 
% \midrule
% \textbf{ArXiv} \cite{clement2019arxiv} & N/A \\ 
% \textbf{Cosmopedia-v2} \cite{benallal2024cosmopedia} & N/A \\ 
% \textbf{PeS2o} \cite{peS2o} & N/A \\ 
% \textbf{Zyda-StarCoder-Git-Commits} \cite{tokpanov2024zyda} & N/A \\ 
% \textbf{Zyda-StarCoder-Languages} & N/A \\ 
% \textbf{FineWeb-Edu} \cite{lozhkov2024fineweb-edu} & N/A \\ 
% \bottomrule
% \end{tabular}
% \caption{\footnotesize Pretraining datasets and token counts for Spectra-2 models.}
% \label{tab:datasets}
% \end{table}

\item \textbf{Zyda-StarCoder} 
\textbf{Git-Commits} \cite{tokpanov2024zyda}: For our models, we exclusively utilize the GitHub-Issues and Jupyter-Structured subsets of the Zyda-Starcoder dataset.

\item \textbf{Zyda-StarCoder-Languages}: A dataset encompassing multiple programming languages, enabling the model to perform well across diverse coding tasks.

\item \textbf{FineWeb-Edu} \cite{lozhkov2024fineweb-edu}: A subset of high quality dataset consists of 1.3T tokens of educational web pages filtered from FineWeb dataset.

\end{itemize}

\subsection{Hyperparameter Choices}
    
We adopt a single learning rate with a warmup followed by a cosine decay schedule, replacing the dual learning rate approach used in TriLMs \citep{spectra}. Additionally, we eliminate the use of weight decay, consistent with the modifications.

\begin{table}[h]
\centering
\begin{tabular}{@{}lr@{}}
\toprule
\textbf{Feature}                & \textbf{Spectra-1.1}      \\ 
\midrule
Biases                          & None                    \\
Activation                      & SwiGLU                  \\
RoPE ($\theta$)                 & $5 \cdot 10^5$          \\
QKV Normalization               & QK-Norm                 \\
Layer Norm                      & RMSNorm                 \\
Layer Norm Applied to           & Outputs                 \\
Z-Loss Weight                   & $10^{-5}$               \\
Weight Decay on Embeddings      & No                      \\
\bottomrule
\end{tabular}
\caption{Configuration Details for Spectra-1.1}
\label{tab:spectra1.1}
\vspace{-0.5cm}
\end{table}

\begin{table*}[h]
\centering
    \begin{tabular}{l c c c }
        \toprule
        \textbf{Parameter} & \textbf{Spectra-1.1-1B} & \textbf{Spectra-1.1-2B} & \textbf{Spectra-1.1-3B} \\
        \midrule
        \textbf{Number of Parameters} & 1.526B  & 2.5547B & 3.6680B \\
        \midrule
        \textbf{Hidden Size} & 2048 & 2560 & 3072 \\
        \midrule
        \textbf{Number of Layers} & 24 & 26 & 28 \\
        \midrule
        \textbf{Attention Heads} & 16 & 20 & 24 \\
        \midrule
        \textbf{MLP Hidden Size} & 8192 & 10240 & 11264 \\
        \midrule
        \textbf{Number of KV Heads} & 4 & 5 & 6 \\
        \midrule
        \textbf{Embedding Size} & 32768 & 32768 & 32768 \\
        \midrule
        \textbf{Max Sequence Length} & 2048 & 2048 & 2048 \\
        \midrule
        \textbf{Activation Function} & SiLU & SiLU & SiLU \\
        \midrule
        \textbf{Optimizer} & AdamW & AdamW & AdamW \\
        \midrule
        \textbf{Learning Rate} & 0.0015 & 0.0015 & 0.0015 \\
        \midrule
        \textbf{Weight Decay} & 0.1 & 0.1 & 0.1 \\
        \midrule
        % \textbf{Batch Size} & 2048 & 2048 & 2048 \\
        % \midrule
        \textbf{Gradient Clipping} & 1.0 & 1.0 & 1.0 \\
        \bottomrule
    \end{tabular}
    \caption{\footnotesize{Architecture summary for Spectra-1.1 1B, 2B, and 3B models based on revised configurations.}}
    \label{tab:arch-summary}
\end{table*}

    % \textbf{1B Configuration:} \href{https://github.com/Ayushk4/OLMo/blob/main/configs/official/TriLM-1B_Spectra-1.1_data.yaml}{TriLM-1B Configuration}$} \\
    % \textbf{2B Configuration:} \href{https://github.com/Ayushk4/OLMo/blob/main/configs/official/TriLM-2B_Spectra-1.1_data.yaml}{TriLM-2B Configuration}$} \\
    % \textbf{3B Configuration:} \href{https://github.com/Ayushk4/OLMo/blob/main/configs/official/TriLM-3B_Spectra-1.1_data.yaml}{TriLM-3B Configuration}$} \\

% \subsection{Known Implementation Artifacts}
% Similar to Spectra (TriLMs) \citep{spectra}, our models exhibit artifacts due to model parallelism, particularly during scale computation across sharded weight matrices. To mitigate this, we compute scales locally on each device, minimizing communication overhead. This modification has a negligible impact on bits per parameter (less than $10^{-5}$) even with high model parallelism. 

\subsection{Hyperparameters.} All the models are randomly initialized from a truncated normal distribution with a mean of 0 and a standard deviation of 0.02. We trained using the AdamW optimizer \citep{loshchilov2019decoupledweightdecayregularization}, with $\beta_1 = 0.9$, $\beta_2 = 0.95$, and $\epsilon = 10^{-5}$. The weight decay was applied with a value of 0.1. A cosine learning rate schedule was employed, with a warmup of 2000 steps, followed by a decay of the final learning rate to 10\% of the peak learning rate. We used gradient clipping with a threshold of 1.0. Metrics were logged every 10 steps. For simplicity during training, we adopt a single learning rate with a warmup followed by a cosine decay schedule, replacing the dual learning rate approach used in Spectra. Additionally, we eliminate the use of weight decay, consistent with the modifications. Table \ref{tab:arch-summary} summarizes the hyperparameters for our largest models.

\subsection{Hardware and Training Setups}

Each node in the Frontier cluster includes four AMD MI250X accelerators, with each accelerator featuring two GCDs that function as independent GPUs. The total bidirectional communication bandwidth within a node ranges between 100~GB/s and 400~GB/s. The nodes are connected via Ethernet-based HPE Slingshot interconnects. Each node is equipped with four links, each providing a total directional bandwidth of 50~GB/s. Our approach scales near-linearly up to 2048 GPUs, as shown in \Cref{fig:gpus_speedup}

\begin{figure*}[t] 
  \centering
  \begin{minipage}{0.45\textwidth}
    \centering
    \includegraphics[width=\linewidth]{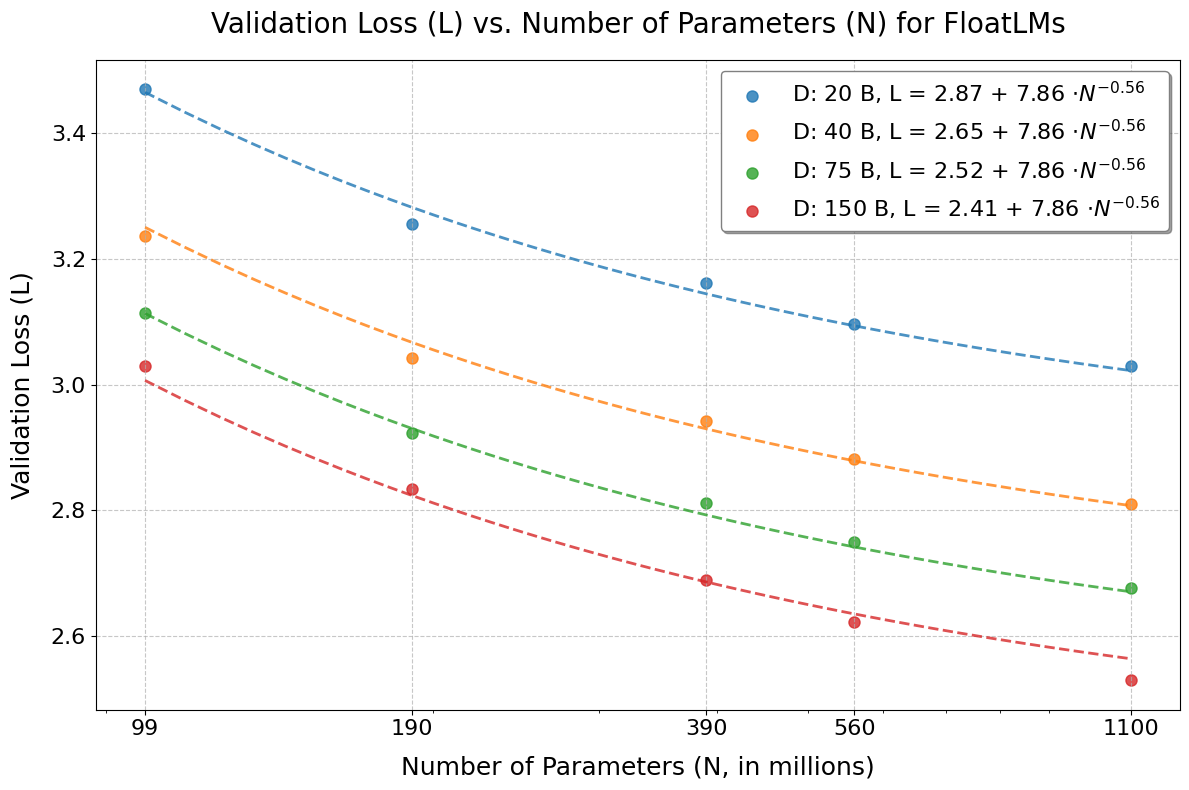} % Adjust the width as needed
    \label{fig:param_scaling_fig_float} 
  \end{minipage}
  \hfill
  \begin{minipage}{0.45\textwidth}
    \centering
    \includegraphics[width=\linewidth]{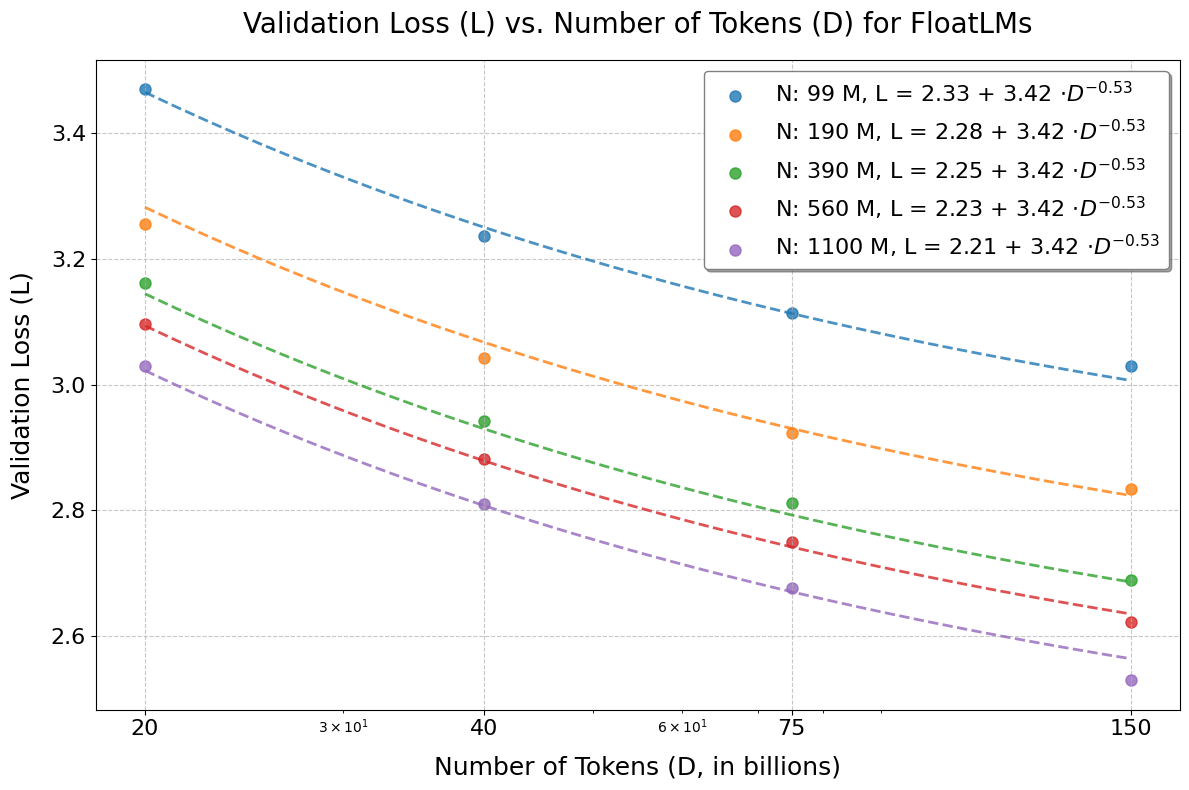} % Replace with your figure's path
    \label{fig:token_scaling_fig_float}
  \end{minipage}
  \caption{\footnotesize{Effect of scaling number of parameters (left) and number of training tokens (right) on final validation loss for FloatLMs. The dotted lines show the power law derived in \Cref{eq:power_law_float}}.}
  \label{fig:scaling_fig_float}
\end{figure*}

\section{Scaling Laws}
\label{app:scaling}

\subsection{Scaling Laws of TriLMs and FloatLMs}

In \Cref{sec:scaling_trilms}, we derived the scaling law for TriLMs as a function of the number of parameters ($N$) and the number of training tokens used ($D$) by assuming the parametric form defined in \citet{kaplan2020scalinglawsneurallanguage, hoffmann2022trainingcomputeoptimallargelanguage}. We apply the same procedure to derive the scaling law for FloatLMs which use 16-bit precision to facilitate direct comparison and understand the effect of compute on performance. 

\begin{figure*}[!ht] 
  \centering
  \begin{minipage}{0.45\textwidth}
    \centering
    \includegraphics[width=\linewidth]{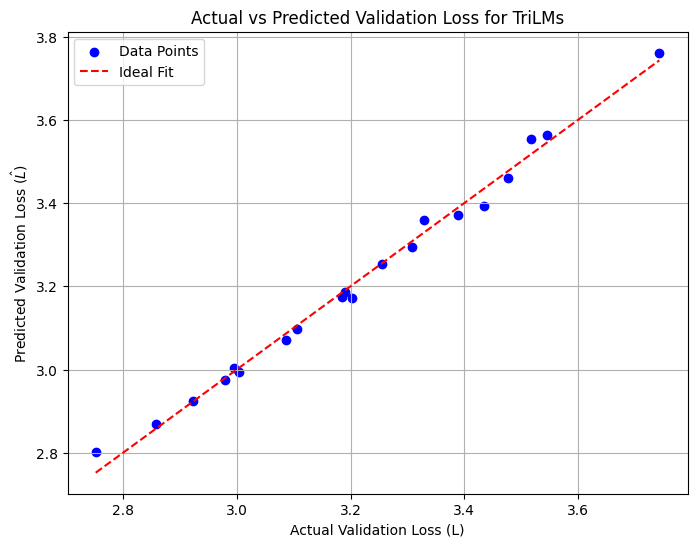} % Adjust the width as needed
    \label{fig:fit_trilm} 
  \end{minipage}
  \hfill
  \begin{minipage}{0.45\textwidth}
    \centering
    \includegraphics[width=\linewidth]{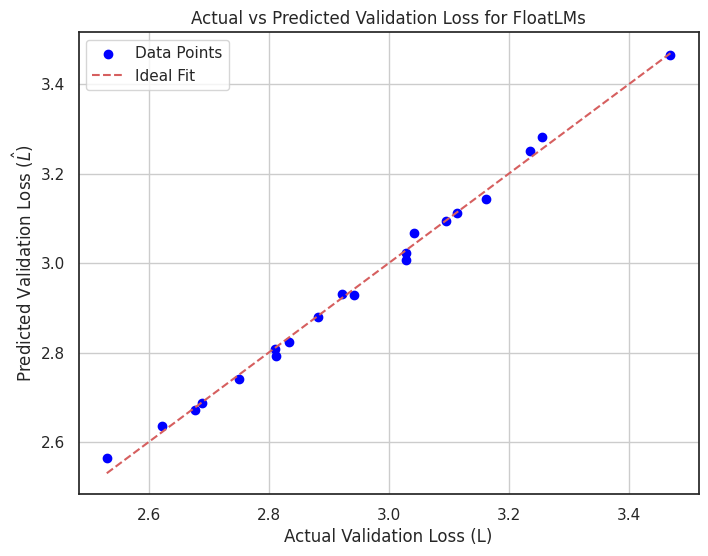} % Replace with your figure's path
    \label{fig:fit_floatlm}
  \end{minipage}
  \caption{\small{Predicted versus actual values of the final validation loss based on the parametric fit of the scaling law for TriLMs (left) and FloatLMs (right).}}
  \label{fig:scale_fit}
\end{figure*}

In addition to the ternary LLMs described in \Cref{sec:scaling_trilms}, we train corresponding 16-bit models which we refer to as FloatLMs across parameters sizes $\in [990 M, 1900 M, 3900 M, 5600 M, 11000 M]$ (excluding embeddings) and dataset sizes $\in [20 B, 40 B, 75 B, 150 B]$ tokens. We follow the same procedure as for TriLMs and obtain the following power law for FloatLMs,
\begin{equation}
    \hat{L}(N,D) \approx 2.17 + \frac{7.86}{N^{0.56}} + \frac{3.42}{D^{0.53}}.
\label{eq:power_law_float}
\end{equation}
Comparing this with \Cref{eq:power_law}, we make two interesting observations. First, the constant term and the coefficients are markedly different for ternary and float LMs, indicating that these terms might be dependent on the level of quantization. Second, the terms involving $N$ and $D$ have almost the same exponents for FloatLMs, which means that increasing either parameters and training tokens has a similar effect on improving LLM performance. This is in contrast to TriLMs, where the term involving training tokens decays much more rapidly than term involving number of parameters.

\Cref{fig:scaling_fig_float} shows the final validation loss for different FloatLM models against the number of parameters and the number of training tokens, along with the scaling law fit.

\subsection{Parametric Fit for Scaling Law}

We obtain the coefficients for the parametric scaling law in \Cref{eq:scaling_params} by finding the least squares fit on the the final validation losses of the suite of models trained across different parameter and training token values.

To evaluate our fit, we calculate the coefficient of determination, or $R^{2}$, which is a statistical measure that indicates how well a model fits a set of data, with $R^2 = 1.0$ indicating a perfect fit. Our fitted power laws have $R^2 = 0.9921$ for TriLMs and $R^2 = 0.9958$ for FloatLMs. \Cref{fig:scale_fit} plots the predicted validation loss following our derived scaling law versus the actual empirical values.

\begin{table*}[!t]
\centering
\small
\renewcommand{\arraystretch}{1.3} % Increase row height for better readability
\begin{tabular}{@{}llcccc@{}}
\toprule
\textbf{Dataset} & \textbf{Metric} & \textbf{Spectra-1.1 1B} & \textbf{Spectra-1.1 2B} & \textbf{Spectra-1.1 3B} & \textbf{Llama-1 7B} \\ \toprule
\multirow{2}{*}{Arc Challenge} & acc & 33.45\scriptsize{$\pm 1.38$} & 37.29\scriptsize{$\pm 1.41$} & 40.61\scriptsize{$\pm 1.44$} & 41.81\scriptsize{$\pm 1.44$} \\
& acc\_norm & 36.43\scriptsize{$\pm 1.41$} & 39.69\scriptsize{$\pm 1.43$} & 42.58\scriptsize{$\pm 1.44$} & 44.80\scriptsize{$\pm 1.45$} \\ \hline
\multirow{2}{*}{Arc Easy} & acc & 69.82\scriptsize{$\pm 0.94$} & 72.60\scriptsize{$\pm 0.92$} & 75.97\scriptsize{$\pm 0.88$} & 75.25\scriptsize{$\pm 0.89$} \\
& acc\_norm & 62.54\scriptsize{$\pm 0.99$} & 67.42\scriptsize{$\pm 0.96$} & 71.93\scriptsize{$\pm 0.92$} & 72.81\scriptsize{$\pm 0.91$} \\ \hline
\multirow{1}{*}{BoolQ} & acc & 62.57\scriptsize{$\pm 0.85$} & 56.70\scriptsize{$\pm 0.87$} & 66.15\scriptsize{$\pm 0.83$} & 75.11\scriptsize{$\pm 0.76$} \\ \hline
\multirow{2}{*}{HellaSwag} & acc & 43.20\scriptsize{$\pm 0.49$} & 46.44\scriptsize{$\pm 0.50$} & 49.65\scriptsize{$\pm 0.50$} & 56.95\scriptsize{$\pm 0.49$} \\
& acc\_norm & 56.61\scriptsize{$\pm 0.49$} & 61.37\scriptsize{$\pm 0.49$} & 66.28\scriptsize{$\pm 0.47$} & 76.21\scriptsize{$\pm 0.42$} \\ \hline
\multirow{1}{*}{LAMBADA (OpenAI)} & acc & 47.31\scriptsize{$\pm 0.70$} & 48.85\scriptsize{$\pm 0.70$} & 54.22\scriptsize{$\pm 0.89$} & 73.53\scriptsize{$\pm 0.61$} \\ \hline
\multirow{1}{*}{LAMBADA (Standard)} & acc & 34.81\scriptsize{$\pm 0.66$} & 38.58\scriptsize{$\pm 0.68$} & 47.04\scriptsize{$\pm 0.70$} & 67.82\scriptsize{$\pm 0.65$} \\ \hline
\multirow{2}{*}{LogiQA} & acc & 22.12\scriptsize{$\pm 1.63$} & 22.27\scriptsize{$\pm 1.63$} & 22.00\scriptsize{$\pm 1.66$} & 22.73\scriptsize{$\pm 1.64$} \\
& acc\_norm & 27.04\scriptsize{$\pm 1.75$} & 29.65\scriptsize{$\pm 1.79$} & 30.57\scriptsize{$\pm 1.81$} & 30.11\scriptsize{$\pm 1.80$} \\ \hline
\multirow{2}{*}{OpenBookQA} & acc & 28.60\scriptsize{$\pm 2.02$} & 30.00\scriptsize{$\pm 2.05$} & 32.20\scriptsize{$\pm 2.09$} & 34.20\scriptsize{$\pm 2.12$} \\
& acc\_norm & 38.80\scriptsize{$\pm 2.18$} & 41.00\scriptsize{$\pm 2.20$} & 41.80\scriptsize{$\pm 2.21$} & 44.40\scriptsize{$\pm 2.22$} \\ \hline
\multirow{2}{*}{PIQA} & acc & 71.98\scriptsize{$\pm 1.05$} & 73.67\scriptsize{$\pm 1.03$} & 76.01\scriptsize{$\pm 1.00$} & 78.67\scriptsize{$\pm 0.96$} \\
& acc\_norm & 72.47\scriptsize{$\pm 1.04$} & 75.41\scriptsize{$\pm 1.00$} & 76.33\scriptsize{$\pm 0.99$} & 79.16\scriptsize{$\pm 0.95$} \\ \hline
\multirow{1}{*}{WinoGrande} & acc & 58.09\scriptsize{$\pm 1.39$} & 58.56\scriptsize{$\pm 1.38$} & 62.43\scriptsize{$\pm 1.36$} & 69.93\scriptsize{$\pm 1.29$}\\ \midrule
% \multirow{1}{*}{\textbf{C\&R Avg.}} & & & & & \\ \midrule
\multirow{2}{*}{SciQ} & acc & 89.60\scriptsize{$\pm 0.97$} & 90.80\scriptsize{$\pm 0.91$} & 92.80\scriptsize{$\pm 0.82$} & 94.60\scriptsize{$\pm 0.72$} \\
& acc\_norm & 84.10\scriptsize{$\pm 1.16$} & 87.00\scriptsize{$\pm 1.06$} & 88.40\scriptsize{$\pm 1.01$} & 93.00\scriptsize{$\pm 0.81$}\\ \hline
% \multirow{1}{*}{TriviaQA} & & & & & \\ \midrule
% \multirow{1}{*}{\textbf{Knowledge Avg.}} & & & & &\\ \midrule
\multirow{1}{*}{MMLU (cont.): Humanities} & acc & 29.16\scriptsize{$\pm 0.65$} & 30.33\scriptsize{$\pm 0.66$} & 30.90\scriptsize{$\pm 0.65$} & 33.28\scriptsize{$\pm 0.67$} \\
\multirow{1}{*}{MMLU (cont.): Other} & acc & 38.46\scriptsize{$\pm 0.86$} & 40.42\scriptsize{$\pm 0.86$} & 49.39\scriptsize{$\pm 0.87$} & 46.31\scriptsize{$\pm 0.86$} \\
\multirow{1}{*}{MMLU (cont.): Social Sciences} & acc & 35.81\scriptsize{$\pm 0.86$} & 38.97\scriptsize{$\pm 0.87$} & 40.92\scriptsize{$\pm 0.87$} & 42.44\scriptsize{$\pm 0.88$} \\
\multirow{1}{*}{MMLU (cont.): STEM} & acc & 27.62\scriptsize{$\pm 0.79$} & 30.23\scriptsize{$\pm 0.80$} & 32.06\scriptsize{$\pm 0.82$} & 33.43\scriptsize{$\pm 0.83$} \\
\multirow{1}{*}{MMLU (cont.) Average} & acc & 32.34\scriptsize{$\pm 0.39$} & 34.43\scriptsize{$\pm 0.39$} & 36.12\scriptsize{$\pm 0.39$} & 38.21\scriptsize{$\pm 0.40$} \\ \hline
\multirow{1}{*}{GSM8K} & exact\_match & 2.05\scriptsize{$\pm 0.39$} & 2.12\scriptsize{$\pm 0.40$} & 3.03\scriptsize{$\pm 0.47$} & 9.70\scriptsize{$\pm 0.82$} \\ \hline
\multirow{2}{*}{MathQA} & acc & 23.22\scriptsize{$\pm 0.77$} & 24.22\scriptsize{$\pm 0.78$} & 24.69\scriptsize{$\pm 0.79$} & 27.07\scriptsize{$\pm 0.81$} \\ 
& acc\_norm & 23.12\scriptsize{$\pm 0.77$} & 24.52\scriptsize{$\pm 0.79$} & 24.63\scriptsize{$\pm 0.79$} & 26.50\scriptsize{$\pm 0.81$} \\ \bottomrule
\end{tabular}
\caption{\footnotesize{Model performance across various datasets.}}
\label{tab:model-performance}
\end{table*}

\section{Benchmark Details}
\label{app:benchmark}
We benchmark TriLM across knowledge, commonsense, and reasoning benchmarks. We average our scores across three different 'seeds'. 

\subsection{Commonsense and Reasoning} \label{appendix:Benchmark_Details::subsection:Commonsense_and_Reasoning}
    We report commonsense and reasoning benchmark scores across 6 benchmarks in Table \ref{tab:model-performance}. Each is considered in a zero-shot setting. Following are the details of each of the benchmarks considered:

\begin{itemize}
    \item \textbf{ARC Challenge and Easy}: \citep{ARC} The ARC dataset consists of 7,787 multiple-choice science questions, split into two categories: Challenge and Easy. We compute both the accuracy and normalized accuracy for these two sets.
    \item \textbf{BoolQ}: \citep{boolq} BoolQ is a reading comprehension dataset featuring naturally occurring yes/no questions. We evaluate the model's performance by calculating its accuracy on this task.
    \item \textbf{HellaSwag}: \citep{hellaswag} HellaSwag is a dataset for testing grounded commonsense through multiple-choice questions. Incorrect answer choices are generated using Adversarial Filtering (AF), designed to deceive machines but not humans. Accuracy and normalized accuracy are reported for this dataset.
    \item \textbf{WinoGrande}: \citep{winogrande} WinoGrande is a dataset of 44,000 questions designed to assess commonsense reasoning via a fill-in-the-blank task with binary options. We report the model's accuracy on this dataset.
    \item \textbf{PIQA}: \citep{piqa} The Physical Interaction Question Answering (PIQA) dataset evaluates physical commonsense reasoning. We compute accuracy and normalized accuracy for this task.
    \item \textbf{LAMBADA OpenAI}: \citep{lambada} LAMBADA is a dataset used to test text understanding through next-word prediction, containing narrative passages from BooksCorpus. To perform well on LAMBADA, models must leverage broad discourse information rather than just local context. We report both perplexity and accuracy for this dataset.
    \item \textbf{LogiQA}: \citep{logiqa} LogiQA focuses on testing human-like logical reasoning across multiple types of deductive reasoning tasks. We measure both accuracy and normalized accuracy for this dataset.
\end{itemize}

\subsection{Knowledge}
            We report performance on SciQ, TriviaQA in Tables \ref{tab:model-performance}. Each is considered in a zero-shot setting. Following are the details of each of the benchmarks considered:

            The knowledge-based evaluation included the following tasks:
            \begin{itemize}
                \item \textbf{SciQ}: \citep{sciq} The SciQ dataset contains multiple-choice questions with 4 answer options from crowd-sourced science exams. The questions range from Physics, Chemistry and Biology and several other fields. We calculate the accuracy and length normalized accuracy on this task.

                \item \textbf{TriviaQA}: \citep{triviaqa} TriviaQA is a reading comprehension dataset containing question-answer-evidence triples. We calculate the exact match accuracy on this task.
                
                \item \textbf{MMLU} \citep{mmlu}: The benchmark aims to assess the knowledge gained during pretraining by evaluating models solely in zero-shot and few-shot scenarios. It spans 57 subjects, including STEM fields, humanities, social sciences, and more.

            \end{itemize}

\subsection{Serving benchmark for inference}
We report the following serving benchmark for our TriRun kernels.

\begin{itemize}
    \item \textbf{Time to First Token.} The time taken from the start of the inference process until the model generates its first token. This metric is used to measure the latency before the model begins producing outputs.

    \item \textbf{Time per Output Token.} The average time taken by the model to generate each subsequent token after the first. This metric reflects the efficiency of the model in producing tokens once the inference process has started.
\item \textbf{Total Tokens per Second.} The overall rate at which the model generates tokens, including both the initial and subsequent tokens. This metric accounts for the entire sequence generation process and provides an aggregate measure of inference speed.

\item \textbf{Output Tokens per Second.} The rate at which the model generates tokens after the first token has been produced. This metric focuses on sustained generation speed, reflecting the model’s efficiency once the decoding process has started.

\end{itemize}

\section{Formal Proofs}
\subsection{Notations and Theorem}
\label{subsec:theorem1_correctness}
% \paragraph{Theorem 1 (Correctness).} For a sequence \( D = \{ d_1, d_2, \dots, d_n \} \) of ternary digits \( d_i \in \{-1, 0, 1\} \), grouped into blocks of size \( k \) and packed into \( p \)-bit integers with \( 2^p > 3^k \), the \texttt{P} and \texttt{U} operations are exact inverses, ensuring \( \texttt{U(P(D))} = D \).
\paragraph{Theorem 1 (Correctness).}  Let \( D = ( d_1, d_2, \dots, d_n ) \) be a sequence of ternary digits \( d_i \in \{-1, 0, 1\} \). When \( D \) is partitioned into blocks of size \( k \) and each block is encoded into a \( p \)-bit integer, the encoding and decoding operations \( P \) and \( U \) are lossless if and only if  \( 2^p > 3^k \); that is, \( U(P(D)) = D \). 

Let  
\[
D = \{d_1, d_2, \dots, d_n\}, \qquad d_i \in \{-1,0,1\},
\]
be a sequence of balanced ternary digits. We partition \(D\) into blocks of \(k\) digits (with the last block possibly shorter). For a given block, define the shifted digits by
\[
d'_j = d_j + 1,\quad j = 0,1,\dots,k-1,
\]
so that \(d'_j \in \{0,1,2\}\). Then define the integer
\[
N = \sum_{j=0}^{k-1} d'_j \cdot 3^{\,k-1-j}.
\]
Since each \(d'_j\) is in \(\{0,1,2\}\), we have
\[
0 \le N \le 3^k - 1.
\]

Assume we choose an integer \(p\) such that
\[
2^p > 3^k.
\]
The \emph{packing} function is defined by
\[
b = \left\lfloor \frac{N \cdot 2^p + (3^k - 1)}{3^k} \right\rfloor.
\]
This mapping is one-to-one on the set \(\{0,1,\dots,3^k-1\}\) and yields an integer \(b\) in the range \([0, 2^p - 1]\).

The \emph{unpacking} function recovers a number \(x\) via
\[
x = \left\lfloor \frac{b \cdot 3^k - (3^k - 1) + (2^p - 1)}{2^p} \right\rfloor.
\]
The recovery of the shifted digits is given by:
\[
d'_j = \left( \left\lfloor \frac{x}{3^{\,k-1-j}} \right\rfloor \right) \bmod 3,\quad j = 0,1,\dots,k-1.
\]

\subsection{Proof of Theorem 1 (Correctness).}

\paragraph{Step 1. Necessity of the Condition.}  
Notice that the mapping \(P\) takes an input \(N \in \{0,1,\dots,3^k-1\}\) (a total of \(3^k\) values) and produces an output \(b \in \{0,1,\dots,2^p-1\}\) (a total of \(2^p\) values). If
\[
2^p \le 3^k,
\]
then by the pigeonhole principle the mapping \(P\) cannot be injective, and therefore lossless recovery is impossible. Thus, a necessary condition for \(U(P(D)) = D\) is that
\[
2^p > 3^k.
\]

We now show that \(U(P(D)) = D\) if and only if \(2^p > 3^k\). We start by showing that \(x = N\), and then we recover the original digits.
\paragraph{Step 2. Expressing the Packing Equation via the Division Algorithm.}

By the division algorithm, there exists a unique remainder integer \( r \) with \( 0 \le r \le 3^k-1 \) such that
\[
N\cdot2^p + (3^k-1) = b\cdot3^k + r.
\]
Rearranging, we obtain
\[
N\cdot2^p = b\cdot3^k - (3^k-1) + r.
\]
Dividing both sides by \( 2^p \) yields
\[
N = \frac{b\cdot3^k - (3^k-1)}{2^p} + \frac{r}{2^p}.
\]

Because \( 0 \le r \le 3^k-1 \), the term
\[
\frac{r}{2^p}
\]
satisfies \[
0 \le \frac{r}{2^p} < \frac{3^k}{2^p}.
\]
Thus, the requirement \(2^p > 3^k\) is equivalent to having
\[
0 \le \frac{r}{2^p} < 1.
\]
If \(2^p \le 3^k\) the fractional part might reach or exceed 1, and the mapping would fail to be one-to-one. Hence, the lossless property holds \emph{if and only if} \(2^p > 3^k\).
% Since \( 0 \le r \le 3^k-1 \) and \( 2^p > 3^k \), we have
% \[
% 0 \le \frac{r}{2^p} < 1.
% \]
% Thus, \( N \) is expressed as the sum of an exact rational number and a fractional part strictly less than \( 1 \).

\paragraph{Step 3. Recovery of \( N \) via the Decoding Operation.}

Examine the decoding formula:
\[
x = \left\lfloor \frac{b\cdot3^k - (3^k-1) + (2^p-1)}{2^p} \right\rfloor.
\]
We rewrite the expression inside the floor as
\[
\begin{aligned}
\frac{b\cdot3^k - (3^k-1) + (2^p-1)}{2^p} \\
= \frac{b\cdot3^k - (3^k-1)}{2^p} + \frac{2^p-1}{2^p} \\
= N - \frac{r}{2^p} + \frac{2^p-1}{2^p} \\
= N + \frac{(2^p-1)-r}{2^p}.
\end{aligned}
\]
Since \( 0 \le r \le 3^k-1 \) and \(3^k < 2^p\), the correction term
\[
\frac{(2^p-1)-r}{2^p}
\]
satisfies
\[
0 \le \frac{(2^p-1)-r}{2^p} < 1.
\]
Thus,
\[
N \le N + \frac{(2^p-1)-r}{2^p} < N+1.
\]
Taking the floor gives
\[
x = N.
\]

\subsection*{Step 4. Recovery of the Original Ternary Digits}

Since \( N \) represents the base-3 number with shifted digits \( d_j' \in \{0,1,2\} \), we recover each \( d_j' \) by writing \( N \) in base 3. It is important to note that if \( N \) has a “short” base-3 representation (i.e., fewer than \( k \) digits), we must pad the representation on the left with zeros so that it has exactly \( k \) digits.  In other words, we interpret the expansion of \( x \) as

\[
x = \sum_{j=0}^{k-1} d_j' \cdot 3^{k-1-j},
\]

where the digits \( d_j' \) include leading zeros as needed. Then, for each \( j = 0,1,\dots,k-1 \), we have

\[
d_j' = \left( \left\lfloor \frac{x}{3^{k-1-j}} \right\rfloor \right) \mod 3.
\]

Finally, reversing the initial shift,

\[
d_j = d_j' - 1, \quad j = 0,1,\dots,k-1,
\]

retrieves the original balanced ternary digits.

% Since \( N \) represents the base-3 number with shifted digits \( d'_j \in \{0,1,2\} \), we recover each \( d'_j \) by writing \( N \) in base 3. In particular, for each \( j = 0,1,\dots,k-1 \) we have
% \[
% d'_j = \left( \left\lfloor \frac{x}{3^{\,k-1-j}} \right\rfloor \right) \bmod 3.
% \]
% Finally, reversing the initial shift,
% \[
% d_j = d'_j - 1,\quad j = 0,1,\dots,k-1,
% \]
% retrieves the original balanced ternary digits.

\subsection*{Conclusion}

The decoding operation precisely recovers \( N \), and therefore the original sequence of digits. In other words,
\[
U(P(D)) = D.
\]
This completes the corrected proof that the packing and unpacking functions are exact inverses.

\hfill\(\Box\)

\section{Inference implementation on CPUs and benchmarking across hardware.}  
\subsection{Additional Implementation details of TQ2.}
\label{app:tq_2_implementation}
For quantization, the packed value calculation and detailed encoding steps are as follows:  

\begin{itemize}
    \item \textbf{Packing}: \( q_{\text{packed}} = q_0 + 4q_1 + 16q_2 + 64q_3 \).  
    \item \textbf{Storage}: 64 bytes for quantized elements + 2 bytes for the float16 scaling factor \( d_i \), totaling 66 bytes per block.  
\end{itemize}

For dequantization, the explicit unpacking procedure involves:  
$
q_0 = q_{\text{packed}} \mod 4, \quad 
q_1 = \left\lfloor \frac{q_{\text{packed}}}{4} \right\rfloor \mod 4,$ $ \quad
q_2 = \left\lfloor \frac{q_{\text{packed}}}{16} \right\rfloor \mod 4, $
$\quad
q_3 = \left\lfloor \frac{q_{\text{packed}}}{64} \right\rfloor \mod 4.
$
The ternary storage method uses only 2 bits per element, with minimal overhead from the float16 scale per block. The process relies on hardware-friendly bitwise operations for fast packing and unpacking, making it suitable for large-scale deployments in memory-constrained environments while maintaining a balance between numerical fidelity and storage efficiency.

\subsection{Additional Implementation details of TQ1.}
\label{app:tq_1_implementation}

\begin{figure*}[th!] % or [!htbp] for more flexible positioning
    \centering
    \includegraphics[width=0.41\textwidth]{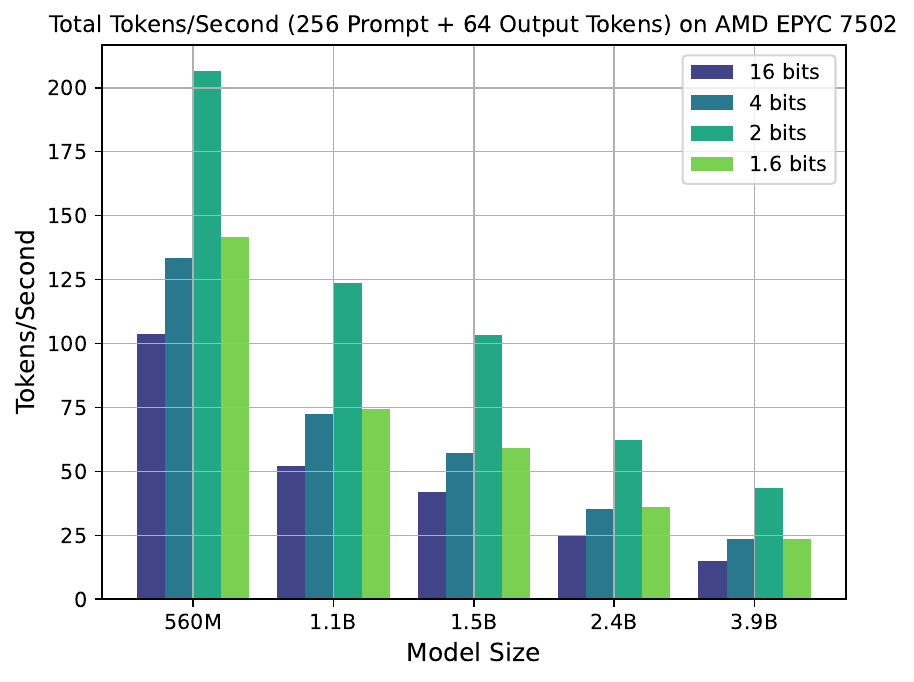} % Replace with the first figure's path
    \includegraphics[width=0.45\textwidth]{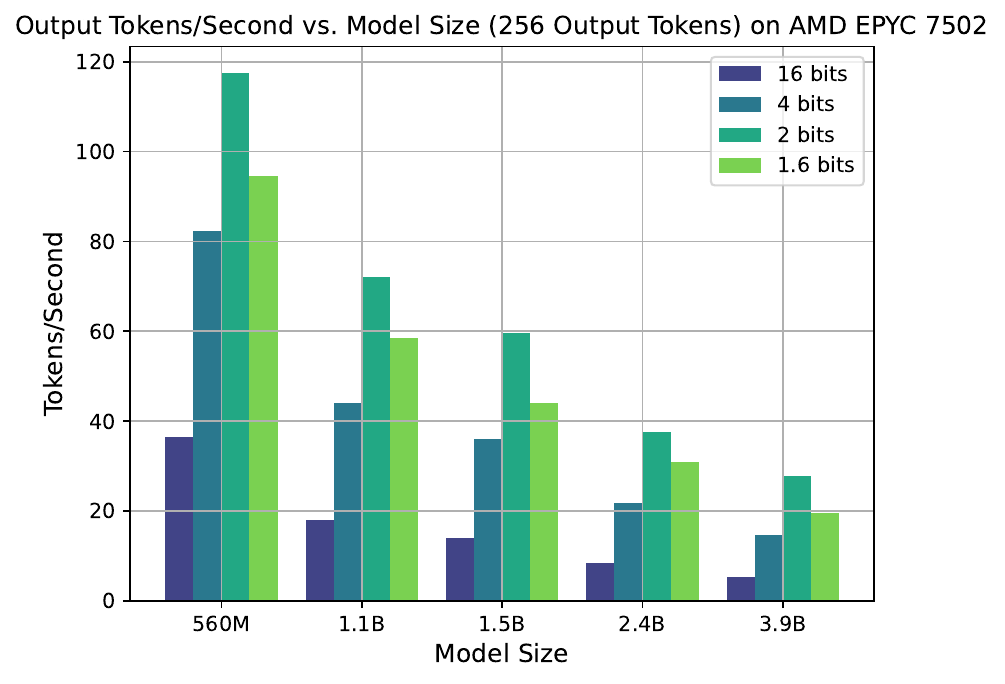} % Replace with the second figure's path
    \caption{\footnotesize{Comparison of output tokens for different model sizes running on a AMD EPYC 750 laptop:
    \textbf{(Left)} Output tokens (for a 256 prompt with 64 output tokens).
    \textbf{(Right)} Output tokens per second versus model size. For more details, refer to  \cref{tab:benchmark_AMD_EPYC_7502}}}
    \label{fig:cpu_tq1_tq_2_amd}
\end{figure*}

\paragraph{Ternary digit extraction using fixed-point and bitwise operations.} In this optimized decoding approach, we define \( i \) as the index variable, which represents the iteration counter for extracting each trit. The index \( i \) ranges from 0 to 4, as we are extracting \( k = 5 \) trits from a packed byte \( b \). The procedure begins by setting \( b_0 = b \). For each iteration \( i \), we multiply the current value \( b_i \) by 3, yielding a 10-bit intermediate value. The high byte of this value is then extracted to obtain the ternary digit \( d_i' = \left\lfloor \frac{b_i \cdot 3}{2^8} \right\rfloor \), where \( d_i' \in \{0, 1, 2\} \). After this, the remainder is updated for the next iteration using the operation \( b_{i+1} = (b_i \cdot 3) \& 0xFF \). This process repeats for all iterations \( i = 0, 1, 2, 3, 4 \), extracting the corresponding ternary digits. Once all the trits \( d_i' \) are extracted, they are normalized by subtracting 1, mapping the values from \( \{0, 1, 2\} \) to \( \{-1, 0, 1\} \). 

\begin{algorithm}[htbp]
\caption{Ternary Digit Extraction Using Fixed-Point and Bitwise Operations}
\label{alg:ternary_extraction}
\begin{algorithmic}[1]
\State \textbf{Input:} Packed byte \( b \)
\State \textbf{Output:} Extracted ternary digits \( d_0, d_1, d_2, d_3, d_4 \)

\State Initialize \( b_0 = b \)
\For{each iteration \( i = 0, 1, 2, 3, 4 \)}
    \State Multiply \( b_i \) by 3 to get a 10-bit intermediate value
    \State Extract high byte to get ternary digit \( d_i' = \left\lfloor \frac{b_i \cdot 3}{2^8} \right\rfloor \)
    \State Update remainder for next iteration: \( b_{i+1} = (b_i \cdot 3) \& 0xFF \)
\EndFor
\State Normalize extracted digits by subtracting 1, mapping \( \{0, 1, 2\} \) to \( \{-1, 0, 1\} \)
\State \textbf{Return:} \( d_0, d_1, d_2, d_3, d_4 \)
\end{algorithmic}
\end{algorithm}

This iterative method replaces the costly division and modulo operations with fixed-point arithmetic and bitwise masking, both of which are highly optimized for SIMD implementations. The structure of this approach minimizes data dependencies, enabling the parallel extraction of trits across multiple packed bytes. Furthermore, by leveraging the near equivalence of \( 3^5 \) and \( 2^8 \), it achieves efficient decoding of ternary values with computational complexity that scales linearly with \( k \). The avoidance of costly arithmetic operations and compatibility with SIMD architectures make this approach particularly well-suited for high-performance applications involving ternary arithmetic.

% In this optimized decoding approach, for each packed byte \( b \), the procedure begins by setting \( b_0 = b \). The trits are then extracted iteratively by multiplying \( b_i \) by 3, yielding a 9-bit intermediate value. The high byte of this value is then extracted to obtain \( d_i' = \left\lfloor \frac{b_i \cdot 3}{2^8} \right\rfloor \), which corresponds to a ternary digit from the set \( \{0, 1, 2\} \). The remainder is then updated for the next iteration with the operation:  
% \[
% b_{i+1} = (b_i \cdot 3) \& 0xFF.
% \]  
% Finally, the values \( d_i' \) are normalized by subtracting 1, mapping \( \{0, 1, 2\} \) to \( \{-1, 0, 1\} \). 

% This iterative decoding method offers significant advantages for SIMD implementation. It replaces non-vectorizable division and modulo operations with fixed-point arithmetic and bitwise masking, both of which are highly SIMD-friendly. The iterative structure minimizes data dependencies, enabling the parallel extraction of trits across multiple packed bytes. By leveraging the numerical proximity of \( 3^5 \) and \( 2^8 \), this method achieves efficient decoding of ternary values with a computational complexity linear in \( k \). 

\subsection{Benchmarking across various hardware.}
\label{app:cpu_bench}

We present a comprehensive benchmarking analysis of quantization kernels: TQ1 (1.6 bits), TQ2 (2 bits), Q4 (4 bits, implementation provided in ggml), and FP16 (16 bits), across various model sizes ranging from 560M to 3.9B parameters and different token configurations. The benchmarks are executed on the AMD EPYC 7502 and Apple M4 Max (14 CPU cores). Detailed results are presented in Tables \ref{tab:lM4_Max_14_CPU_Cores} and \ref{tab:benchmark_AMD_EPYC_7502}. It should be noted that further hardware-level optimizations are possible and will be addressed in future work.

\paragraph{Prompt Encoding Performance:} On the Apple M4 Max, TQ1 and TQ2 outperformed FP16 and Q4, particularly for longer prompts, indicating efficient utilization of lower-bit quantization for prompt processing on this architecture. In contrast, benchmarks on the AMD EPYC 7502 showed that FP16 achieved superior throughput for shorter prompts (32–128 tokens), while Q4 and TQ2 gained an advantage at 256 tokens, highlighting a precision vs. memory bandwidth trade-off. TQ1 underperformed FP16 and Q4 on this platform. Across both architectures, larger models led to reduced prompt encoding throughput.

\paragraph{Autoregressive Decoding Performance.}
Autoregressive decoding consistently demonstrated quantization's performance benefits on both platforms. Quantized kernels (Q4, TQ2, TQ1) outperformed FP16 in output tokens per second. AMD EPYC 7502 showed substantial Q4 gains over FP16, with TQ2 further improving throughput, highlighting reduced precision benefits for decoding. Apple M4 Max showed even greater quantization improvements; TQ2 achieved highest throughput, followed by TQ1 and Q4, all exceeding FP16. Output token length (8-256) minimally impacted decoding throughput, suggesting independence within this range. Larger models reduced decoding throughput, consistent with prompt encoding.

\paragraph{Combined Prompt Encoding and Autoregressive Decoding Performance.}
The combined benchmark confirmed quantization advantages. For both platforms, quantized kernels, especially TQ2 and Q4, delivered higher overall tokens per second than FP16 in combined scenarios. Prompt/decode token ratio (256/8, 256/64, 256/128) influenced overall throughput. Increased decoding token proportion decreased overall tokens per second, reflecting lower decoding throughput relative to prompt encoding. Apple M4 Max demonstrated highest combined throughput with TQ1 and TQ2, particularly at higher decoding token ratios, indicating optimization for end-to-end generation on this architecture.

\begin{table*}[]
\centering
\small
\begin{tabular}{lrllllll}
\toprule
\multicolumn{3}{c}{Configuration} & \multicolumn{5}{c}{Model Size} \\
\cmidrule(lr){1-3} \cmidrule(lr){4-8}
Tokens & Bits & Kernel & 560M & 1.1B & 1.5B & 2.4B & 3.9B \\
\midrule
\multicolumn{7}{c}{\textbf{Prompt Encoding Benchmark (Prompt Tokens/seconds})} \\
\midrule
32 & 16 & FP16 & 223.1 ± 0.3 & 110.7 ± 0.1 & 77.2 ± 0.2 & 49.1 ± 0.0 & 30.7 ± 0.1 \\
32 & 4 & Q4 & 182.4 ± 0.3 & 92.0 ± 0.1 & 70.2 ± 0.9 & 44.2 ± 0.8 & 27.8 ± 0.0 \\
32 & 2 & TQ2 & 315.2 ± 0.6 & 165.1 ± 0.4 & 130.1 ± 0.1 & 83.7 ± 0.0 & 51.3 ± 0.1 \\
32 & 1.6 & TQ1 & 181.1 ± 0.2 & 89.2 ± 0.1 & 70.0 ± 0.1 & 40.3 ± 0.1 & 25.4 ± 0.0 \\
64 & 16 & FP16 & 233.0 ± 0.3 & 113.5 ± 0.1 & 80.2 ± 0.1 & 49.7 ± 0.6 & 31.0 ± 0.0 \\
64 & 4 & Q4 & 182.9 ± 0.5 & 98.7 ± 0.1 & 69.3 ± 0.2 & 44.2 ± 0.7 & 27.6 ± 0.0 \\
64 & 2 & TQ2 & 320.5 ± 0.5 & 179.2 ± 0.3 & 130.6 ± 0.1 & 83.9 ± 0.3 & 51.1 ± 0.1 \\
64 & 1 & TQ1 & 180.9 ± 0.5 & 89.6 ± 0.1 & 66.6 ± 0.4 & 40.3 ± 0.0 & 25.3 ± 0.0 \\
128 & 16 & FP16 & 228.6 ± 1.1 & 116.2 ± 0.2 & 81.1 ± 0.2 & 51.5 ± 0.1 & 32.1 ± 0.0 \\
128 & 4 & Q4 & 179.2 ± 0.7 & 97.0 ± 0.6 & 69.3 ± 1.0 & 44.8 ± 0.1 & 27.9 ± 0.0 \\
128 & 2 & TQ2 & 305.2 ± 2.2 & 174.8 ± 0.1 & 124.2 ± 0.1 & 81.9 ± 0.1 & 51.5 ± 0.1 \\
128 & 1.6 & TQ1 & 177.2 ± 0.3 & 90.4 ± 0.1 & 65.8 ± 0.1 & 40.5 ± 0.1 & 25.5 ± 0.0 \\
256 & 16 & FP16 & 220.3 ± 0.6 & 105.9 ± 1.0 & 79.9 ± 0.1 & 50.0 ± 0.1 & 31.3 ± 0.1 \\
256 & 4 & Q4 & 170.1 ± 0.1 & 90.0 ± 2.2 & 69.3 ± 0.1 & 42.9 ± 0.1 & 27.3 ± 0.0 \\
256 & 2 & TQ2 & 287.1 ± 2.2 & 176.5 ± 1.7 & 122.4 ± 0.1 & 77.5 ± 0.3 & 49.9 ± 0.1 \\
256 & 1.6 & TQ1 & 169.5 ± 0.5 & 88.0 ± 1.0 & 64.3 ± 0.0 & 39.4 ± 0.0 & 24.9 ± 0.0 \\
\midrule
\multicolumn{8}{c}{\textbf{Autoregressive Decoding Benchmark (Output Tokens/seconds})} \\
\midrule
8 & 16 & FP16 & 37.6 ± 0.2 & 16.9 ± 0.0 & 14.1 ± 0.0 & 8.6 ± 0.0 & 5.4 ± 0.0 \\
8 & 4 & Q4 & 83.0 ± 0.0 & 47.1 ± 0.2 & 35.5 ± 0.0 & 22.6 ± 0.0 & 15.1 ± 0.0 \\
8 & 2 & TQ2 & 135.1 ± 0.3 & 84.6 ± 0.1 & 62.0 ± 0.1 & 42.0 ± 0.0 & 29.2 ± 0.0 \\
8 & 1.6 & TQ1 & 102.0 ± 0.8 & 62.5 ± 0.1 & 48.6 ± 0.0 & 30.1 ± 0.1 & 20.7 ± 0.0 \\
64 & 16 & FP16 & 37.4 ± 0.0 & 17.8 ± 0.0 & 14.0 ± 0.0 & 8.7 ± 0.0 & 5.4 ± 0.0 \\
64 & 4 & Q4 & 83.1 ± 0.6 & 46.3 ± 0.7 & 35.2 ± 0.1 & 23.1 ± 0.0 & 15.0 ± 0.1 \\
64 & 2 & TQ2 & 126.3 ± 0.0 & 83.2 ± 0.1 & 60.8 ± 0.5 & 44.1 ± 0.1 & 28.9 ± 0.3 \\
64 & 1.6 & TQ1 & 105.4 ± 3.6 & 56.9 ± 0.3 & 45.1 ± 0.1 & 29.8 ± 0.0 & 20.5 ± 0.0 \\
128 & 16 & FP16 & 37.2 ± 0.1 & 18.7 ± 0.3 & 14.3 ± 0.0 & 8.7 ± 0.0 & 5.4 ± 0.0 \\
128 & 4 & Q4 & 85.6 ± 0.3 & 47.9 ± 0.1 & 37.5 ± 0.1 & 23.6 ± 0.0 & 15.0 ± 0.0 \\
128 & 2 & TQ2 & 131.4 ± 0.3 & 82.2 ± 0.5 & 64.1 ± 0.0 & 43.1 ± 0.1 & 28.8 ± 0.0 \\
128 & 1.6 & TQ1 & 104.4 ± 0.2 & 60.5 ± 0.0 & 47.7 ± 0.1 & 31.8 ± 0.1 & 20.8 ± 0.0 \\
256 & 16 & FP16 & 36.4 ± 0.2 & 18.0 ± 0.0 & 14.0 ± 0.1 & 8.4 ± 0.0 & 5.3 ± 0.0 \\
256 & 4 & Q4 & 82.3 ± 0.7 & 44.1 ± 0.7 & 36.1 ± 0.1 & 21.9 ± 0.1 & 14.6 ± 0.0 \\
256 & 2 & TQ2 & 117.5 ± 3.9 & 72.0 ± 0.3 & 59.7 ± 0.5 & 37.6 ± 0.1 & 27.7 ± 0.2 \\
256 & 1.6 & TQ1 & 94.5 ± 0.4 & 58.6 ± 0.6 & 44.0 ± 0.1 & 31.0 ± 0.0 & 19.6 ± 0.3 \\
\midrule
\multicolumn{8}{c}{\textbf{Prompt Encoding + Autoregressive Decoding Benchmark (Tokens/seconds})} \\
\midrule
256/8 & 16 & FP16 & 190.2 ± 0.1 & 90.3 ± 0.3 & 73.7 ± 0.1 & 43.4 ± 0.1 & 26.6 ± 0.1 \\
256/8 & 4 & Q4 & 167.6 ± 0.2 & 86.0 ± 0.2 & 70.2 ± 0.2 & 41.1 ± 0.1 & 27.7 ± 0.3 \\
256/8 & 2 & TQ2 & 271.7 ± 1.0 & 149.3 ± 0.4 & 118.4 ± 0.7 & 73.6 ± 0.1 & 50.7 ± 0.1 \\
256/8 & 1.6 & TQ1 & 164.8 ± 0.6 & 82.4 ± 0.1 & 65.7 ± 0.1 & 38.4 ± 0.1 & 25.2 ± 0.0 \\
256/64 & 16 & FP16 & 103.8 ± 0.7 & 52.3 ± 0.0 & 41.8 ± 0.1 & 25.1 ± 0.1 & 15.1 ± 0.6 \\
256/64 & 4 & Q4 & 133.4 ± 0.8 & 72.4 ± 0.0 & 57.3 ± 0.3 & 35.2 ± 0.1 & 23.5 ± 0.0 \\
256/64 & 2 & TQ2 & 206.4 ± 2.7 & 123.5 ± 0.1 & 103.3 ± 0.4 & 62.1 ± 0.2 & 43.7 ± 0.2 \\
256/64 & 1.6 & TQ1 & 141.6 ± 0.1 & 74.5 ± 0.1 & 59.2 ± 0.3 & 35.9 ± 0.1 & 23.7 ± 0.3 \\
256/128 & 16 & FP16 & 79.0 ± 1.1 & 39.1 ± 0.1 & 31.0 ± 0.0 & 18.8 ± 0.0 & 11.9 ± 0.0 \\
256/128 & 4 & Q4 & 117.0 ± 0.3 & 63.2 ± 0.8 & 51.8 ± 2.1 & 31.7 ± 0.1 & 20.5 ± 0.0 \\
256/128 & 2 & TQ2 & 177.2 ± 1.0 & 104.6 ± 1.0 & 85.7 ± 0.1 & 54.5 ± 0.3 & 37.9 ± 0.2 \\
256/128 & 1.6 & TQ1 & 125.5 ± 0.5 & 68.5 ± 0.2 & 54.6 ± 0.2 & 33.7 ± 0.1 & 22.4 ± 0.0 \\
\bottomrule
\end{tabular}
\caption{Tokens per second for different model sizes and quantization kernels with varying prompt lengths on AMD EPYC 7502. Values represent mean ± standard deviation.}
\label{tab:benchmark_AMD_EPYC_7502}
\end{table*}

\begin{table*}[]
\centering
\small
\begin{tabular}{lrllllll}
\toprule
\multicolumn{3}{c}{Configuration} & \multicolumn{5}{c}{Model Size} \\
\midrule
Tokens & Bits & Kernel & 560M & 1.1B & 1.5B & 2.4B & 3.9B \\
\midrule
\multicolumn{7}{c}{\textbf{Prompt Encoding Benchmark (Prompt Tokens/seconds})} \\
\midrule
32 & 16 & FP16 & 730.0 ± 6.5 & 417.6 ± 17.3 & 295.9 ± 2.4 & 152.4 ± 0.2 & 91.2 ± 0.7 \\
32 & 4 & Q4 & 490.4 ± 3.2 & 270.0 ± 1.6 & 195.2 ± 0.9 & 106.1 ± 0.6 & 61.8 ± 0.3 \\
32 & 2 & TQ2 & 543.0 ± 3.5 & 305.1 ± 1.1 & 221.9 ± 0.6 & 118.3 ± 0.9 & 68.6 ± 0.3 \\
32 & 1.6 & TQ1 & 617.9 ± 2.8 & 362.7 ± 2.5 & 276.5 ± 5.7 & 145.5 ± 1.4 & 84.0 ± 0.0 \\
64 & 16 & FP16 & 1223.8 ± 21.7 & 640.8 ± 2.1 & 440.4 ± 0.8 & 247.1 ± 0.5 & 144.7 ± 0.4 \\
64 & 4 & Q4 & 886.1 ± 7.1 & 451.0 ± 1.3 & 320.7 ± 0.5 & 180.0 ± 1.1 & 105.5 ± 0.4 \\
64 & 2 & TQ2 & 951.1 ± 10.6 & 501.7 ± 1.1 & 363.0 ± 1.2 & 198.7 ± 0.3 & 116.2 ± 0.3 \\
64 & 1.6 & TQ1 & 1104.4 ± 11.5 & 578.3 ± 9.2 & 435.7 ± 0.4 & 235.2 ± 0.7 & 136.1 ± 0.4 \\
128 & 16 & FP16 & 1256.1 ± 4.2 & 788.2 ± 6.4 & 564.8 ± 1.7 & 328.0 ± 1.7 & 205.8 ± 0.6 \\
128 & 4 & Q4 & 1081.5 ± 4.6 & 629.7 ± 2.4 & 460.6 ± 0.6 & 266.6 ± 0.5 & 162.5 ± 0.3 \\
128 & 2 & TQ2 & 1143.2 ± 7.2 & 677.9 ± 2.7 & 494.3 ± 2.7 & 285.8 ± 1.0 & 175.8 ± 0.3 \\
128 & 1.6 & TQ1 & 1223.2 ± 11.1 & 769.6 ± 4.6 & 556.8 ± 0.6 & 320.4 ± 2.1 & 197.9 ± 0.2 \\
256 & 16 & FP16 & 1485.2 ± 3.3 & 785.5 ± 2.8 & 611.6 ± 1.3 & 367.5 ± 0.7 & 243.6 ± 3.6 \\
256 & 4 & Q4 & 1350.2 ± 3.4 & 710.3 ± 1.4 & 545.2 ± 2.1 & 325.8 ± 0.5 & 209.3 ± 2.2 \\
256 & 2 & TQ2 & 1398.6 ± 2.6 & 738.9 ± 2.4 & 561.7 ± 1.9 & 339.2 ± 0.7 & 225.9 ± 0.5 \\
256 & 1.6 & TQ1 & 1468.0 ± 3.5 & 786.7 ± 1.7 & 601.8 ± 2.2 & 361.6 ± 0.2 & 242.9 ± 0.5 \\
\midrule
\multicolumn{8}{c}{\textbf{Autoregressive Decoding Benchmark (Output Tokens/seconds})} \\
\midrule
8 & 16 & FP16 & 170.9 ± 0.8 & 92.6 ± 0.1 & 71.5 ± 0.0 & 44.5 ± 0.0 & 28.1 ± 0.0 \\
8 & 4 & Q4 & 237.8 ± 0.3 & 134.8 ± 0.3 & 107.2 ± 0.3 & 64.8 ± 0.6 & 43.3 ± 0.7 \\
8 & 2 & TQ2 & 278.7 ± 0.6 & 167.4 ± 0.1 & 134.0 ± 0.2 & 86.6 ± 0.0 & 57.6 ± 0.1 \\
8 & 1.6 & TQ1 & 228.8 ± 0.5 & 125.7 ± 0.1 & 99.6 ± 0.1 & 62.3 ± 0.1 & 40.0 ± 0.1 \\
64 & 16 & FP16 & 169.4 ± 0.3 & 92.4 ± 0.1 & 71.3 ± 0.0 & 44.5 ± 0.0 & 28.0 ± 0.2 \\
64 & 4 & Q4 & 236.8 ± 0.7 & 134.0 ± 0.0 & 106.1 ± 0.1 & 66.2 ± 0.5 & 42.6 ± 0.2 \\
64 & 2 & TQ2 & 279.7 ± 0.1 & 166.5 ± 0.1 & 132.2 ± 0.1 & 86.1 ± 0.0 & 57.5 ± 0.0 \\
64 & 1.6 & TQ1 & 227.4 ± 0.5 & 125.4 ± 0.1 & 98.6 ± 0.1 & 62.1 ± 0.0 & 38.3 ± 0.0 \\
128 & 16 & FP16 & 171.5 ± 0.1 & 91.6 ± 0.1 & 71.0 ± 0.1 & 44.5 ± 0.1 & 28.1 ± 0.1 \\
128 & 4 & Q4 & 232.7 ± 0.3 & 132.2 ± 0.1 & 105.5 ± 0.1 & 67.0 ± 0.1 & 43.5 ± 0.3 \\
128 & 2 & TQ2 & 281.3 ± 1.0 & 164.0 ± 0.2 & 132.0 ± 0.1 & 85.1 ± 0.0 & 56.9 ± 0.0 \\
128 & 1.6 & TQ1 & 225.3 ± 1.6 & 124.5 ± 0.1 & 97.8 ± 0.1 & 61.6 ± 0.0 & 39.7 ± 0.0 \\
256 & 16 & FP16 & 166.0 ± 0.4 & 89.4 ± 0.4 & 69.1 ± 0.2 & 43.3 ± 0.2 & 27.7 ± 0.0 \\
256 & 4 & Q4 & 225.1 ± 0.2 & 128.4 ± 0.1 & 101.6 ± 0.6 & 63.7 ± 0.5 & 41.1 ± 0.4 \\
256 & 2 & TQ2 & 268.9 ± 0.2 & 158.6 ± 1.0 & 128.5 ± 0.2 & 82.9 ± 0.2 & 55.6 ± 0.0 \\
256 & 1.6 & TQ1 & 217.2 ± 1.1 & 121.1 ± 0.6 & 95.5 ± 0.0 & 60.3 ± 0.1 & 39.0 ± 0.0 \\

\midrule
\multicolumn{8}{c}{\textbf{Prompt Encoding + Autoregressive Decoding Benchmark (Tokens/seconds})} \\
\midrule
256/8 & 16 & FP16 & 1142.4 ± 7.0 & 628.9 ± 1.4 & 489.9 ± 1.5 & 296.4 ± 0.7 & 194.6 ± 0.3 \\
256/8 & 4 & Q4 & 1165.6 ± 2.6 & 620.8 ± 2.2 & 481.8 ± 1.4 & 288.8 ± 0.5 & 186.6 ± 0.9 \\
256/8 & 2 & TQ2 & 1234.7 ± 0.9 & 668.9 ± 1.3 & 513.5 ± 1.1 & 308.6 ± 0.3 & 202.1 ± 0.5 \\
256/8 & 1.6 & TQ1 & 1241.5 ± 4.7 & 665.5 ± 4.9 & 517.6 ± 1.2 & 311.7 ± 0.5 & 205.4 ± 0.3 \\
256/64 & 16 & FP16 & 550.0 ± 2.9 & 298.9 ± 0.5 & 234.8 ± 0.2 & 144.1 ± 0.3 & 93.6 ± 0.1 \\
256/64 & 4 & Q4 & 648.3 ± 1.1 & 358.6 ± 1.7 & 286.8 ± 0.3 & 175.2 ± 0.7 & 114.6 ± 0.4 \\
256/64 & 2 & TQ2 & 726.3 ± 1.9 & 415.5 ± 0.5 & 328.3 ± 0.3 & 205.2 ± 0.2 & 135.9 ± 0.4 \\
256/64 & 1.6 & TQ1 & 652.7 ± 2.4 & 364.1 ± 0.6 & 285.7 ± 0.3 & 177.0 ± 0.2 & 115.7 ± 0.4 \\
256/128 & 16 & FP16 & 385.8 ± 3.5 & 210.4 ± 1.2 & 165.1 ± 0.5 & 102.1 ± 0.2 & 66.1 ± 0.2 \\
256/128 & 4 & Q4 & 475.3 ± 0.7 & 270.2 ± 0.3 & 216.6 ± 0.4 & 133.1 ± 0.6 & 87.5 ± 0.4 \\
256/128 & 2 & TQ2 & 540.8 ± 0.6 & 317.9 ± 0.6 & 255.4 ± 0.2 & 160.8 ± 0.1 & 108.2 ± 0.1 \\
256/128 & 1.6 & TQ1 & 466.8 ± 0.5 & 265.8 ± 0.4 & 209.9 ± 0.1 & 130.9 ± 0.1 & 86.0 ± 0.3 \\
\bottomrule
\end{tabular}
\caption{Tokens per Second for Different Model Sizes and Quantization Kernels M4 Max (14CPU Coresz). Values represent mean ± standard deviation.}
\label{tab:lM4_Max_14_CPU_Cores}
\end{table*}

\section{TriRun Kernel Design for Accelerated Matrix Multiplication}
\label{additional_implementation_details_for_trirun}
This section presents the design of the TriRun kernel, which accelerates matrix multiplication \( A \times B \to C \), where \( A \) is stored in half-precision (16-bit floating point), \( B \) is quantized to 2 bits per element, and \( C \) is accumulated in single-precision (32-bit floating point) before optional conversion to half-precision. The kernel optimizes memory efficiency and computational throughput through specialized data layouts, dequantization strategies, and tensor-core utilization. Key components include:

\subsection{Data Organization and Quantization}
\paragraph{2-Bit Weight Matrix (\( B \) Storage)} 
The 2-bit quantized elements of \( B \) are packed into 64-bit \texttt{int2} vectors, where each 32-bit integer contains 16 quantized weights. During loading, 64-bit global memory transactions retrieve 32 weights per \texttt{int2}, minimizing memory bandwidth. To align with tensor-core requirements, these packed values are asynchronously copied to shared memory via \texttt{cp.async} instructions, then unpacked into 16-bit fragments for computation.

\paragraph{Half-Precision Matrix (\( A \) Access)} 
Matrix \( A \) is stored in half-precision and loaded via 128-bit \texttt{int4} vectors, fetching eight elements per transaction. This aligns with the 16-byte memory alignment optimal for GPU global memory accesses. Subsequent stages repack these into 16$\times$16 submatrices compatible with tensor-core operations.

\subsection{Dequantization and Tensor-Core Computation}

The \texttt{dequant} function performs dequantization of 2-bit integer values into half-precision floating-point representations, employing hardware-optimized bitwise operations and fused arithmetic to enable efficient tensor core execution. Rather than relying on conventional shift-and-mask techniques, the implementation decomposes each 32-bit word—which encodes sixteen 2-bit weights—using a specialized bitwise operation that leverages a tailored mask to both isolate the individual weight segments and embed a predetermined FP16 exponent. Following this, an integrated arithmetic fusion stage applies a zero-point adjustment, effectively adding 1.0 to the extracted values, and performs dynamic range scaling through a fused multiply-add operation. This approach diverges from the traditional \( \text{scale} \cdot (w - \text{zero\_point}) \) formulation by consolidating multiple arithmetic steps into a single, hardware-specific sequence. Subsequently, per-group FP16 scales are applied to the dequantized values, which are then stored in register-based fragments (\texttt{FragB}) to minimize shared memory contention. 
% The kernel further exploits tensor core capabilities by loading matrix A into fragment registers using specialized instructions and by retaining dequantized matrix B fragments directly in registers, thus enabling each tensor core operation—executed via matrix multiply-accumulate instructions—to compute submatrix products with FP16 inputs and FP32 accumulations for enhanced numerical stability. This tightly integrated design maximizes throughput by unrolling computations across submatrix tiles while preserving warp-level synchronization.
% The \texttt{dequant} function performs dequantization of 2-bit integer values into floating-point representations optimized for tensor core operations. This process involves bitwise manipulations and half-precision arithmetic to reconstruct symmetric quantized values centered at a zero point of $-1$. The implementation consists of three key phases: bitwise decomposition and FP16 embedding, symmetric zero-point adjustment, and dynamic range scaling.    Each 2-bit weight \( w \in \{ 0, 1, 2, 3 \} \) is dequantized to a 16-bit floating-point value. For symmetric quantization, the mapping follows:$
% w_{\text{dequant}} = \text{scale} \cdot (w - \text{zero\_point})$ where \textit{scale} and \textit{zero\_point} are per-group learnable parameters. Bit extraction employs shift-and-mask operations, while sign correction (if required) uses \texttt{lop3} logic for efficient sign-bit propagation. Dequantized weights are cached in register-based fragments (\texttt{FragB}) to minimize shared memory contention. 
Later, the kernel employs \texttt{ldmatrix.sync.aligned.m8n8.x4} to load \( A \) and \( B \) fragments into tensor-core registers. Each \texttt{mma.sync.aligned.m16n8k16} operation computes a 16$\times$8$\times$16 submatrix product, accumulating results into 32-bit floating-point fragments (\texttt{FragC}) for numerical stability. By unrolling across submatrix tiles, the kernel fully utilizes tensor-core throughput while maintaining warp-level synchronization.

\subsection{Memory Latency Hiding via Asynchronous Pipelines}
To overlap computation with memory transfers, the kernel implements a four-stage software pipeline with double buffering. Key mechanisms include:
\begin{itemize}
    \item \textbf{Asynchronous Data Copies}: \texttt{cp.async} instructions prefetch \( A \) and \( B \) tiles into shared memory without stalling computation threads.
    \item \textbf{Double Buffering}: Two shared memory buffers alternate between data ingestion (from global memory) and consumption (by tensor cores), ensuring continuous utilization of memory and compute units.
    \item \texttt{cp.async} Synchronization: Warps issue \texttt{cp.async.commit\_group} to batch memory transactions and \texttt{cp.async.wait\_group} to enforce dependencies, preventing read-after-write hazards.
\end{itemize}

\begin{figure}[h]
    \centering
    \includegraphics[width=0.4\textwidth]{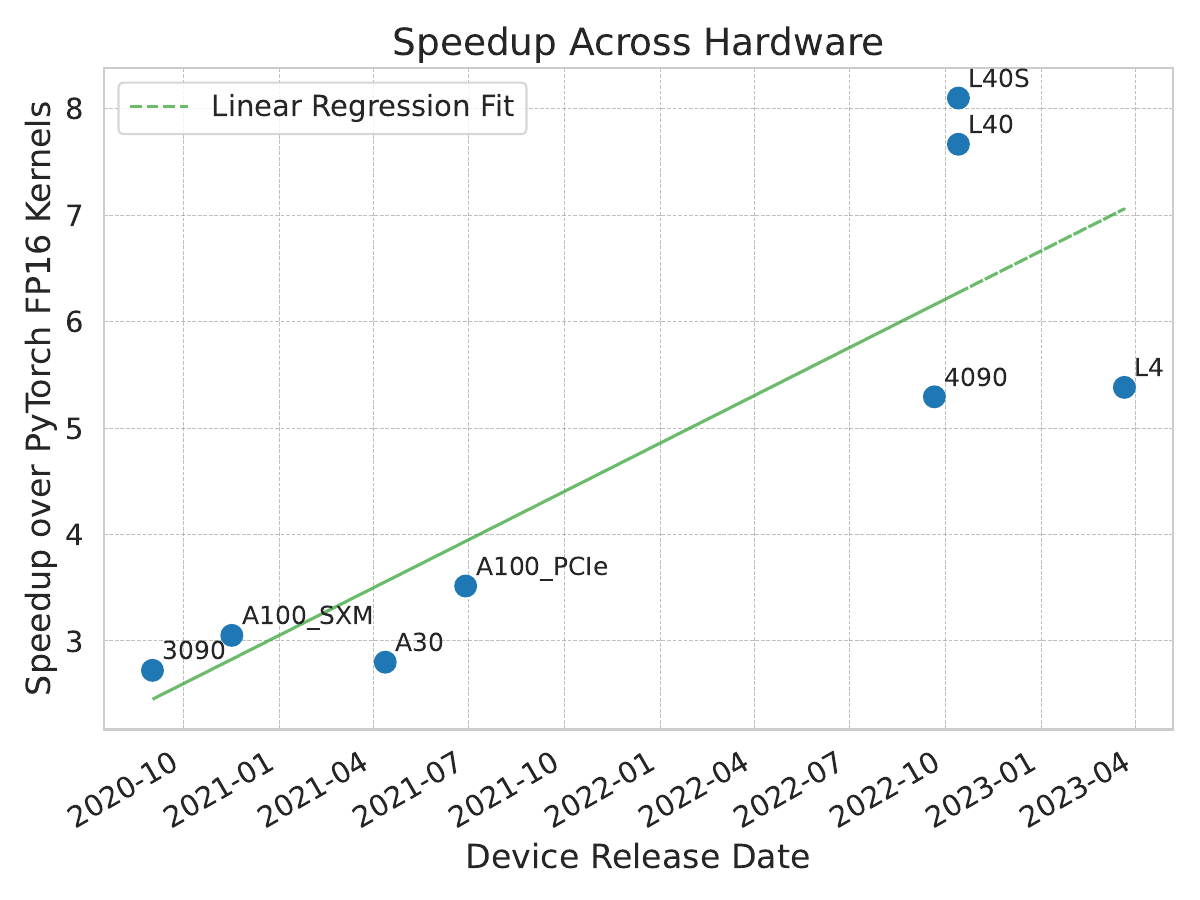}
    \caption{Speedup across hardware over the years using TriRun kernels}
    \label{fig:speedup_plot}
\end{figure}

\subsection{Precision-Preserving Accumulation}
\paragraph{Intra-Warp Reduction} 
Partial sums within a thread block are reduced across warps using shared memory. A tree-based summation merges per-warp \texttt{FragC} outputs, minimizing shared memory bank conflicts through staggered access patterns.

\paragraph{Global Memory Atomic Reduction} 
For outputs spanning multiple thread blocks, atomic 32-bit floating-point additions ensure correct inter-block accumulation. Final results are converted to half-precision (if specified) using round-to-nearest-even mode, balancing precision and storage efficiency.

\subsection{Performance Configuration}

Thread blocks (256 threads) balance register pressure (128/thread) and occupancy (8 warps/block). Tile dimensions adapt to problem size: 128×128 tiles for small batches (m $\leq$ 16) and 64×256 tiles for larger workloads. The 96 KB shared memory budget supports four concurrent pipeline stages, sustaining 98\% tensor core utilization across varied workloads. This implementation demonstrates that 2-bit quantized inference can achieve near-FP16 throughput while maintaining numerical fidelity, providing a practical solution for deploying compressed deep learning models on modern GPUs.

\subsection{TriRun performance benchmark across various Nvidia hardware.}
\label{app:trirun_perfromance_benchmark}

\begin{table*}[h!]
\centering
\begin{tabular}{llrrrrrrrr}
\toprule
Model Size & GPU Type  & \multicolumn{8}{c}{Batch Size} \\
 (Parameters)   &    & 1 & 2 & 4 & 8 & 16 & 32 & 64 & 128 \\
\midrule

\multicolumn{10}{c}{\textbf{A40}} \\
405B & A40 & 7.85 & 7.65 & 7.66 & 7.70 & 7.50 & 7.27 & 3.91 & 1.98 \\
123B & A40 & 7.67 & 7.67 & 7.68 & 7.63 & 7.65 & 6.99 & 3.69 & 1.97 \\
70B & A40 & 7.36 & 7.39 & 7.45 & 7.29 & 7.46 & 6.80 & 3.64 & 1.90 \\
34B & A40 & 7.22 & 7.29 & 7.35 & 7.36 & 7.46 & 6.29 & 3.57 & 1.99 \\
13B & A40 & 6.76 & 6.91 & 6.72 & 6.92 & 6.93 & 5.89 & 3.33 & 1.86 \\
8B & A40 & 6.42 & 6.62 & 6.62 & 6.66 & 6.55 & 5.18 & 2.99 & 1.84 \\
3B & A40 & 5.39 & 4.76 & 5.50 & 5.51 & 5.56 & 4.35 & 2.81 & 1.93 \\
\midrule
\multicolumn{10}{c}{\textbf{3090}} \\
405B & 3090 & 4.92 & 4.70 & 4.70 & 4.98 & 5.04 & 2.72 & 1.41 & 1.09 \\
123B & 3090 & 4.83 & 4.75 & 4.75 & 5.46 & 5.50 & 2.63 & 1.34 & 1.14 \\
70B & 3090 & 4.51 & 4.69 & 4.70 & 4.72 & 4.74 & 2.48 & 1.35 & 1.10 \\
34B & 3090 & 4.42 & 4.70 & 4.76 & 4.75 & 4.65 & 2.53 & 1.62 & 1.23 \\
13B & 3090 & 4.25 & 4.39 & 4.47 & 4.41 & 4.39 & 2.40 & 1.26 & 1.15 \\
8B & 3090 & 4.19 & 4.25 & 4.30 & 4.32 & 4.17 & 2.37 & 1.34 & 1.15 \\
3B & 3090 & 4.05 & 3.53 & 3.71 & 3.71 & 3.68 & 2.32 & 1.32 & 1.24 \\
\midrule
\multicolumn{10}{c}{\textbf{A30}} \\
405B & A30 & 3.96 & 3.98 & 3.99 & 4.00 & 4.01 & 2.80 & 1.77 & 1.35 \\
123B & A30 & 3.89 & 3.90 & 3.90 & 3.91 & 3.90 & 2.66 & 1.68 & 1.12 \\
70B & A30 & 3.81 & 3.82 & 3.91 & 3.94 & 3.90 & 2.70 & 1.64 & 1.19 \\
34B & A30 & 3.62 & 3.62 & 3.67 & 3.81 & 3.78 & 2.76 & 1.71 & 1.29 \\
13B & A30 & 3.36 & 3.44 & 3.45 & 3.48 & 3.38 & 2.47 & 1.59 & 1.48 \\
8B & A30 & 3.28 & 3.30 & 3.32 & 3.33 & 3.31 & 2.28 & 1.61 & 1.13 \\
3B & A30 & 2.66 & 2.90 & 2.90 & 2.94 & 2.87 & 2.08 & 1.35 & 1.39 \\
\midrule
\multicolumn{10}{c}{\textbf{L4}} \\
405B & L4 & 5.98 & 6.10 & 6.12 & 6.24 & 6.46 & 5.38 & 3.21 & 1.00 \\
123B & L4 & 6.34 & 6.05 & 6.05 & 6.08 & 6.07 & 5.24 & 3.12 & 1.63 \\
70B & L4 & 5.91 & 5.97 & 5.96 & 5.90 & 5.86 & 5.22 & 3.15 & 1.65 \\
34B & L4 & 6.75 & 5.87 & 5.87 & 5.84 & 5.76 & 5.40 & 3.20 & 1.80 \\
13B & L4 & 7.75 & 5.64 & 5.64 & 5.63 & 6.16 & 5.03 & 3.13 & 1.66 \\
8B & L4 & 5.31 & 5.25 & 5.27 & 5.23 & 5.30 & 4.42 & 2.79 & 1.56 \\
3B & L4 & 7.74 & 5.43 & 5.43 & 5.38 & 5.40 & 4.49 & 2.83 & 1.63 \\
\bottomrule
\end{tabular}
\caption{Speedup over FP16 PyTorch (using CUTLASS) across different batch sizes for all ternary linear layers in a transformer block, accounting for the matrix structures in models ranging from 3B to 405B on A40, 3090, A30 and L4 GPUs.}
\label{tab:transformer_block_trirun performance-1}
\end{table*}

% \begin{table}[h!]
% \centering
% \begin{tabular}{llrrrrrrrr}
% \toprule
% Model Size & GPU & 1 & 2 & 4 & 8 & 16 & 32 & 64 & 128 \\

\begin{table*}[h!]
\centering
\begin{tabular}{llrrrrrrrr}
\toprule
Model Size & GPU Type  & \multicolumn{8}{c}{Batch Size} \\
 (Parameters)   &    & 1 & 2 & 4 & 8 & 16 & 32 & 64 & 128 \\
\midrule
\multicolumn{10}{c}{\textbf{L40}} \\
405B & L40 & 7.99 & 7.67 & 7.69 & 7.78 & 7.79 & 7.67 & 4.04 & 2.05 \\
123B & L40 & 8.47 & 7.58 & 7.57 & 7.57 & 7.52 & 7.44 & 3.97 & 2.11 \\
70B & L40 & 7.91 & 7.39 & 7.46 & 7.49 & 7.64 & 6.90 & 3.80 & 2.02 \\
34B & L40 & 7.93 & 7.32 & 7.27 & 7.38 & 7.57 & 6.48 & 3.63 & 2.01 \\
13B & L40 & 9.98 & 6.83 & 6.83 & 6.83 & 6.74 & 5.46 & 3.22 & 1.86 \\
8B & L40 & 8.30 & 6.57 & 6.68 & 6.61 & 6.42 & 4.41 & 2.69 & 1.70 \\
3B & L40 & 5.98 & 6.97 & 6.41 & 6.83 & 6.67 & 3.05 & 1.97 & 1.49 \\
\midrule
\multicolumn{10}{c}{\textbf{A100 (SXM)}} \\
405B & A100 & 4.25 & 4.25 & 4.30 & 4.28 & 4.25 & 3.05 & 1.86 & 1.28 \\
123B & A100 & 4.25 & 4.11 & 4.08 & 4.14 & 4.04 & 3.06 & 1.95 & 1.12 \\
70B & A100 & 4.66 & 3.90 & 3.87 & 3.93 & 3.95 & 2.88 & 1.81 & 1.13 \\
34B & A100 & 3.66 & 3.63 & 3.66 & 3.64 & 3.62 & 2.70 & 1.73 & 1.13 \\
13B & A100 & 3.33 & 3.30 & 3.42 & 3.41 & 3.42 & 2.32 & 1.57 & 1.26 \\
8B & A100 & 2.79 & 2.68 & 2.82 & 2.90 & 2.86 & 2.08 & 1.49 & 1.17 \\
3B & A100 & 2.26 & 2.44 & 2.26 & 2.57 & 2.22 & 1.42 & 1.45 & 0.97 \\
\midrule
\multicolumn{10}{c}{\textbf{A100 (PCIe)}} \\
405B & A100 & 4.98 & 4.96 & 4.99 & 5.06 & 4.94 & 3.52 & 2.19 & 1.23 \\
123B & A100 & 4.89 & 4.88 & 4.74 & 4.86 & 4.87 & 3.51 & 2.24 & 1.27 \\
70B & A100 & 4.67 & 4.48 & 4.49 & 4.55 & 4.58 & 3.35 & 2.10 & 1.31 \\
34B & A100 & 4.21 & 4.24 & 4.23 & 4.24 & 4.28 & 3.10 & 1.98 & 1.29 \\
13B & A100 & 3.86 & 3.97 & 3.57 & 3.96 & 3.96 & 2.73 & 1.82 & 1.42 \\
8B & A100 & 3.27 & 3.37 & 3.33 & 3.09 & 3.37 & 2.30 & 1.64 & 1.32 \\
3B & A100 & 2.93 & 2.75 & 2.74 & 2.72 & 2.93 & 1.86 & 1.22 & 1.09 \\
\midrule
\multicolumn{10}{c}{\textbf{4090}} \\
405B & 4090 & 7.67 & 7.65 & 7.69 & 7.71 & 7.77 & 5.29 & 2.74 & 1.48 \\
123B & 4090 & 7.64 & 7.68 & 7.68 & 7.71 & 7.74 & 5.20 & 2.80 & 1.41 \\
70B & 4090 & 4.96 & 6.65 & 7.45 & 7.51 & 7.20 & 5.02 & 2.60 & 1.40 \\
34B & 4090 & 7.25 & 7.28 & 6.82 & 6.84 & 7.34 & 4.89 & 2.59 & 1.43 \\
13B & 4090 & 6.36 & 6.39 & 6.39 & 6.39 & 6.36 & 4.18 & 2.32 & 1.43 \\
8B & 4090 & 5.42 & 5.54 & 5.70 & 5.94 & 5.53 & 3.56 & 1.99 & 1.37 \\
3B & 4090 & 3.96 & 4.29 & 4.40 & 4.32 & 4.15 & 2.52 & 1.37 & 1.37 \\
\bottomrule
\end{tabular}
\caption{Speedup over FP16 PyTorch (using CUTLASS) across different batch sizes for all ternary linear layers in a transformer block, accounting for the matrix structures in models ranging from 3B to 405B on L40, A100(SXM), A100 (PCIe) and 4090 GPUs.}
\label{tab:transformer_block_trirun performance_2}
\end{table*}

\begin{figure*}[t!] % or [!htbp] for more flexible positioning
    \centering
    \includegraphics[width=\textwidth]{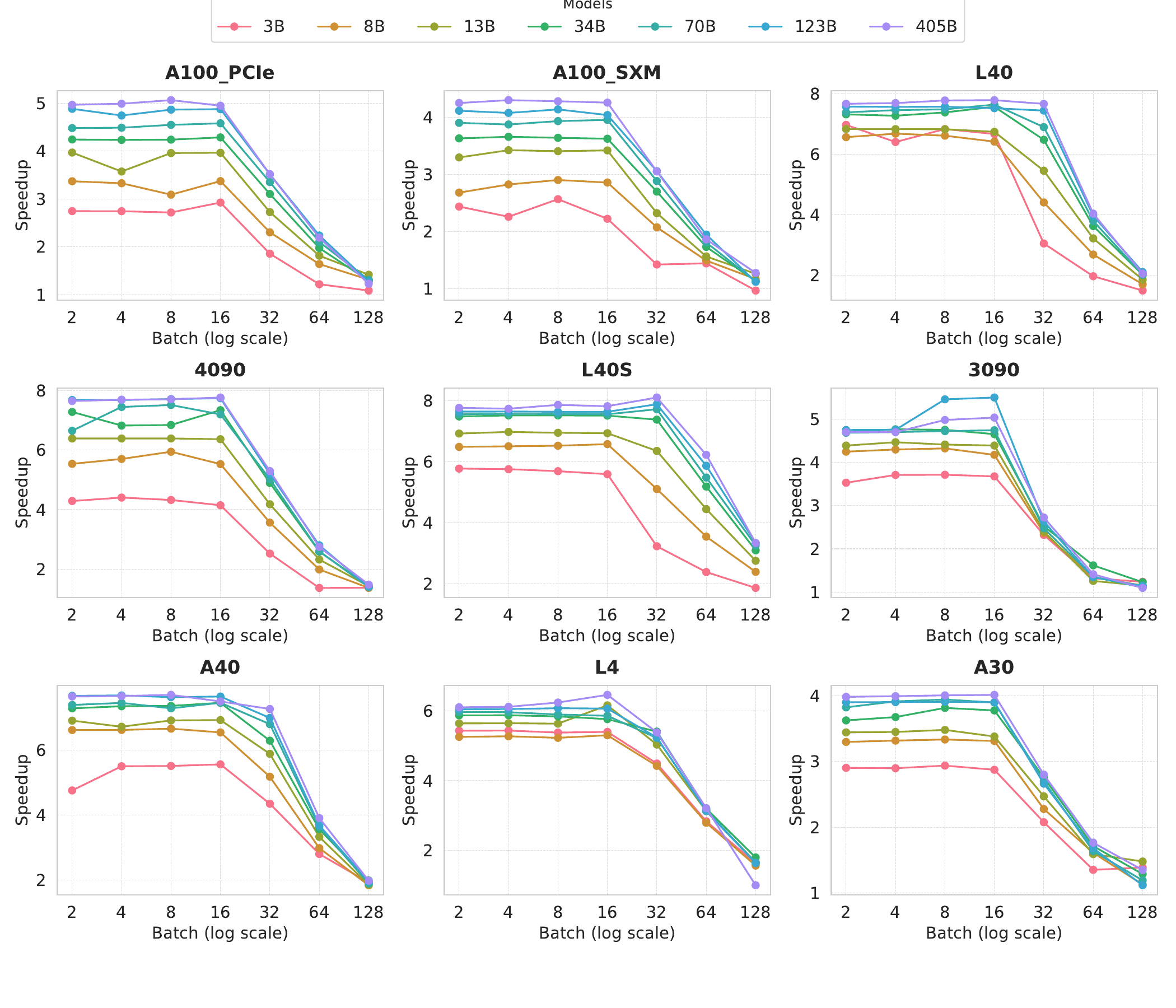} % Replace with your figure's path
    \caption{ We evaluate the performance of ternary layers in transformer blocks, showing near-optimal speedup over PyTorch FP16 on different NVIDIA GPUs using CUTLASS.}
    \label{fig:TriRun_benchmark_transformer_block}
\end{figure*}

\paragraph{Performance Acceleration of Ternary Linear Layers in Transformer Blocks.}   TriRun kernels in transformer blocks including ternary linear layers demonstrate significant performance improvements across various hardware configurations. As shown in \cref{fig:TriRun_benchmark_transformer_block}, we evaluated TriRun’s performance against a standard FP16 PyTorch implementation (CUTLASS) across multiple NVIDIA GPU platforms, including the L40, A100 SXM/PCIe, A40, 3090, A30, L40s, and RTX 4090. The performance analysis covered models ranging from 3B to 405B parameters under varying batch sizes, with speedup quantified as the ratio of FP16 baseline execution time to TriRun execution time. The results indicate sustained performance for batch sizes up to 16–32 across all tested GPUs. Moreover, larger models, which incorporate a higher proportion of ternary weights relative to their total parameters, achieve more substantial speedups compared to smaller models. This performance trend suggests a positive correlation between model size and the efficiency gains offered by TriRun’s implementation. Detailed per-layer results are provided in Tables \cref{tab:transformer_block_trirun performance-1} and \cref{tab:transformer_block_trirun performance_2}. 

\begin{figure*}[h]
    \centering
    \includegraphics[height=0.23\textheight]{images/TTFT_plots_L40S.pdf}
    \includegraphics[height=0.23\textheight]{images/TPOT_plots_L40S.pdf}
    \includegraphics[height=0.23\textheight]{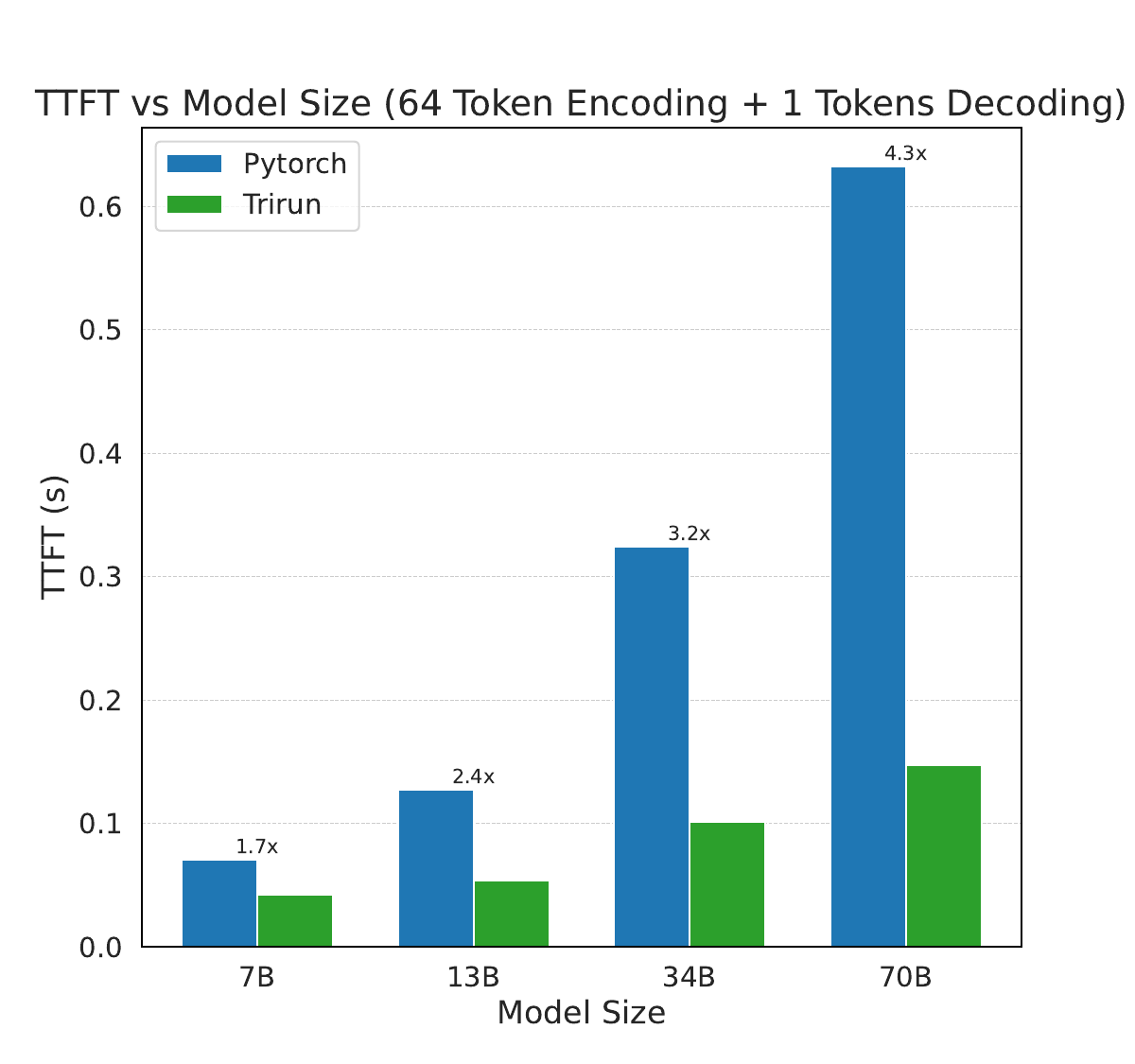}
    \includegraphics[height=0.23\textheight]{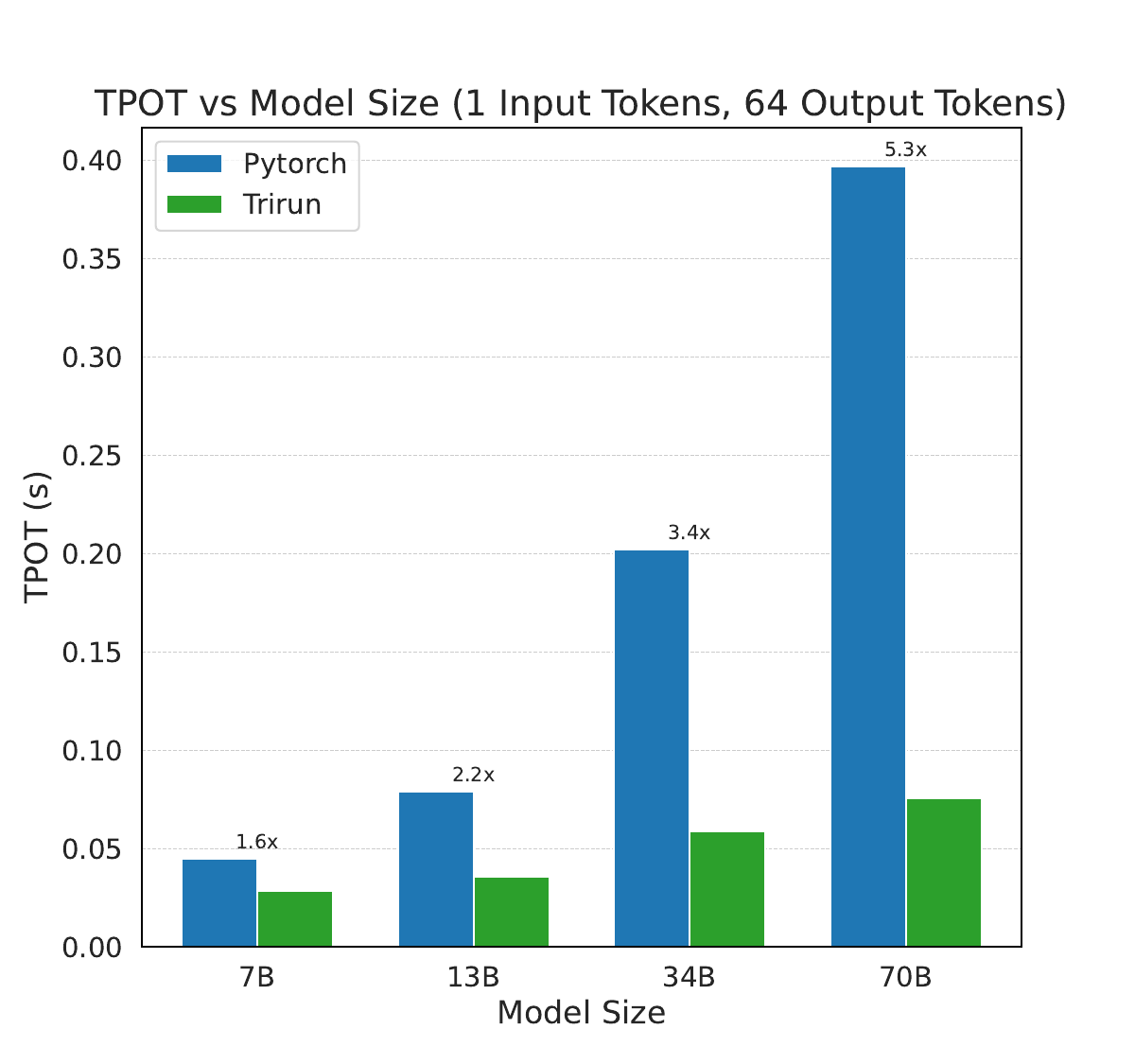}
    \includegraphics[height=0.23\textheight]{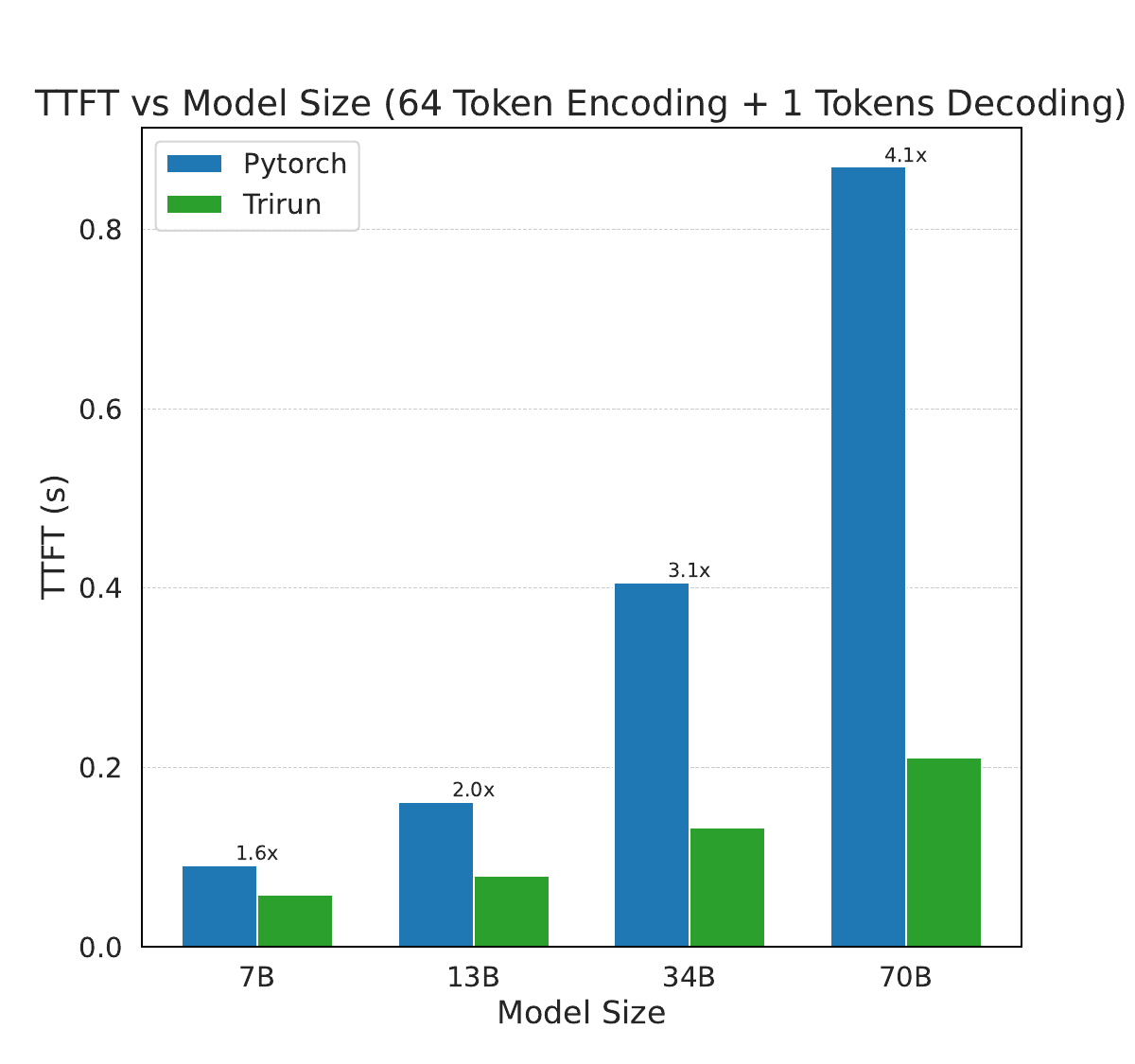}
    \includegraphics[height=0.23\textheight]{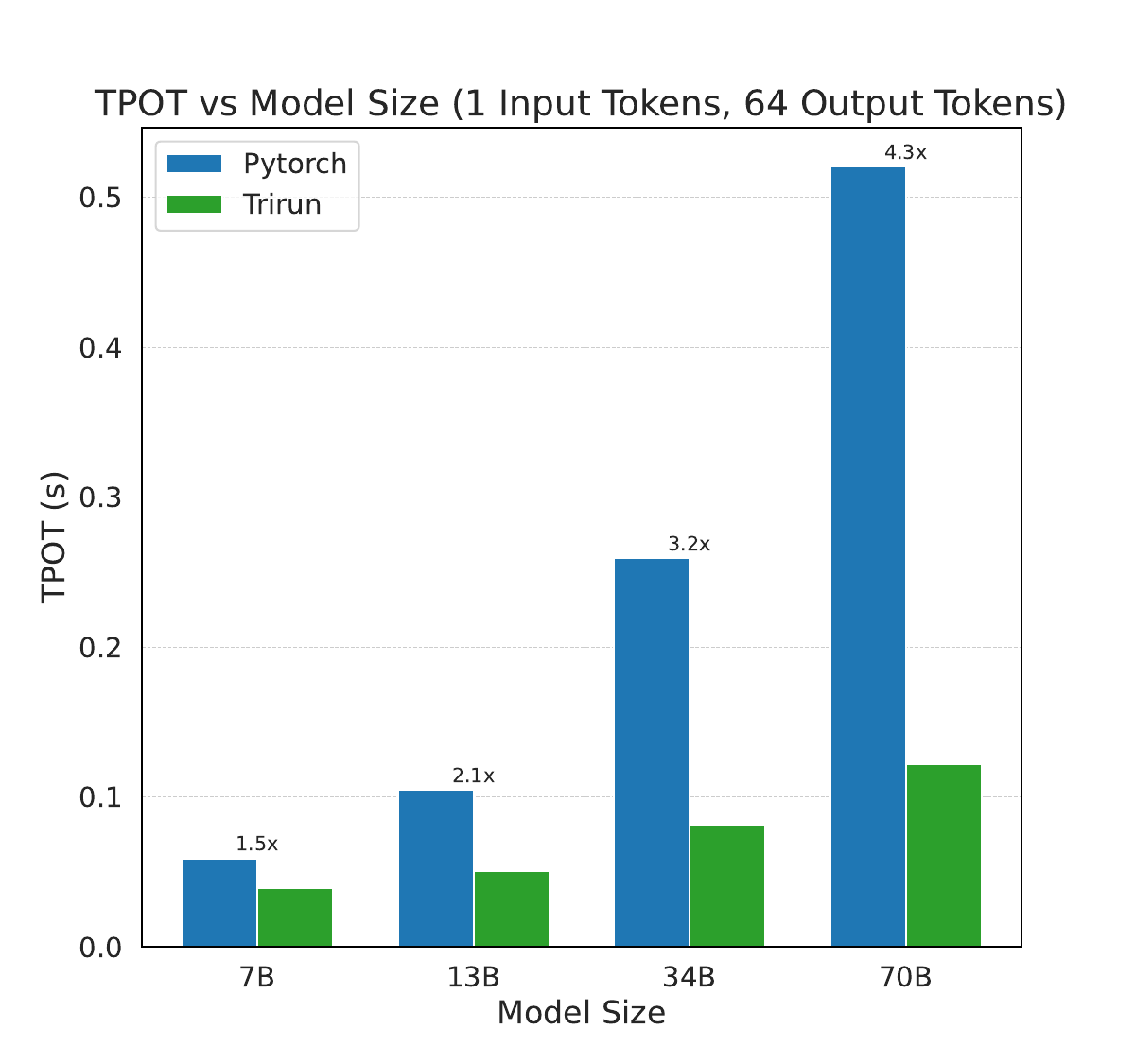}
    \includegraphics[height=0.23\textheight]{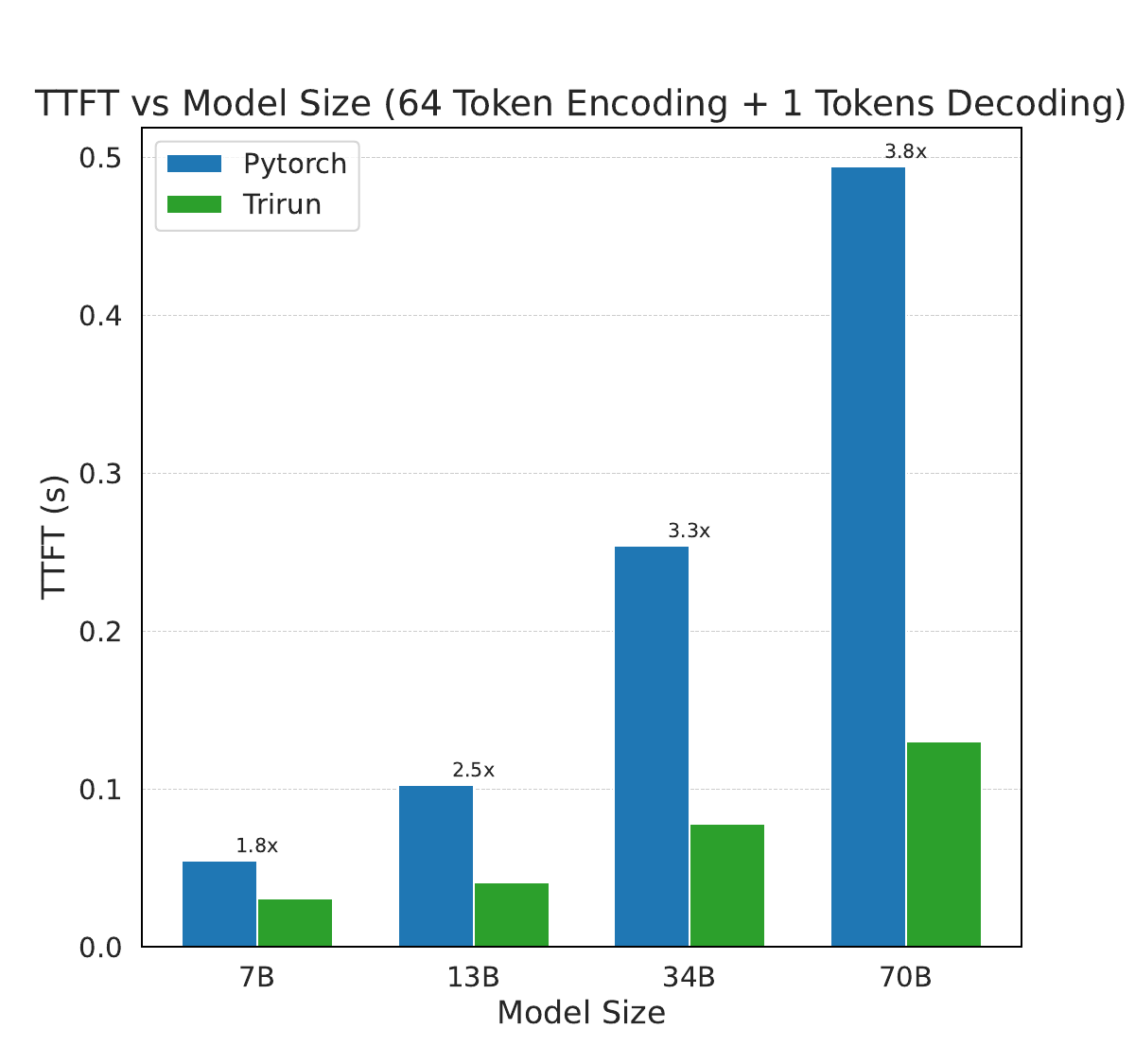}
    \includegraphics[height=0.23\textheight]{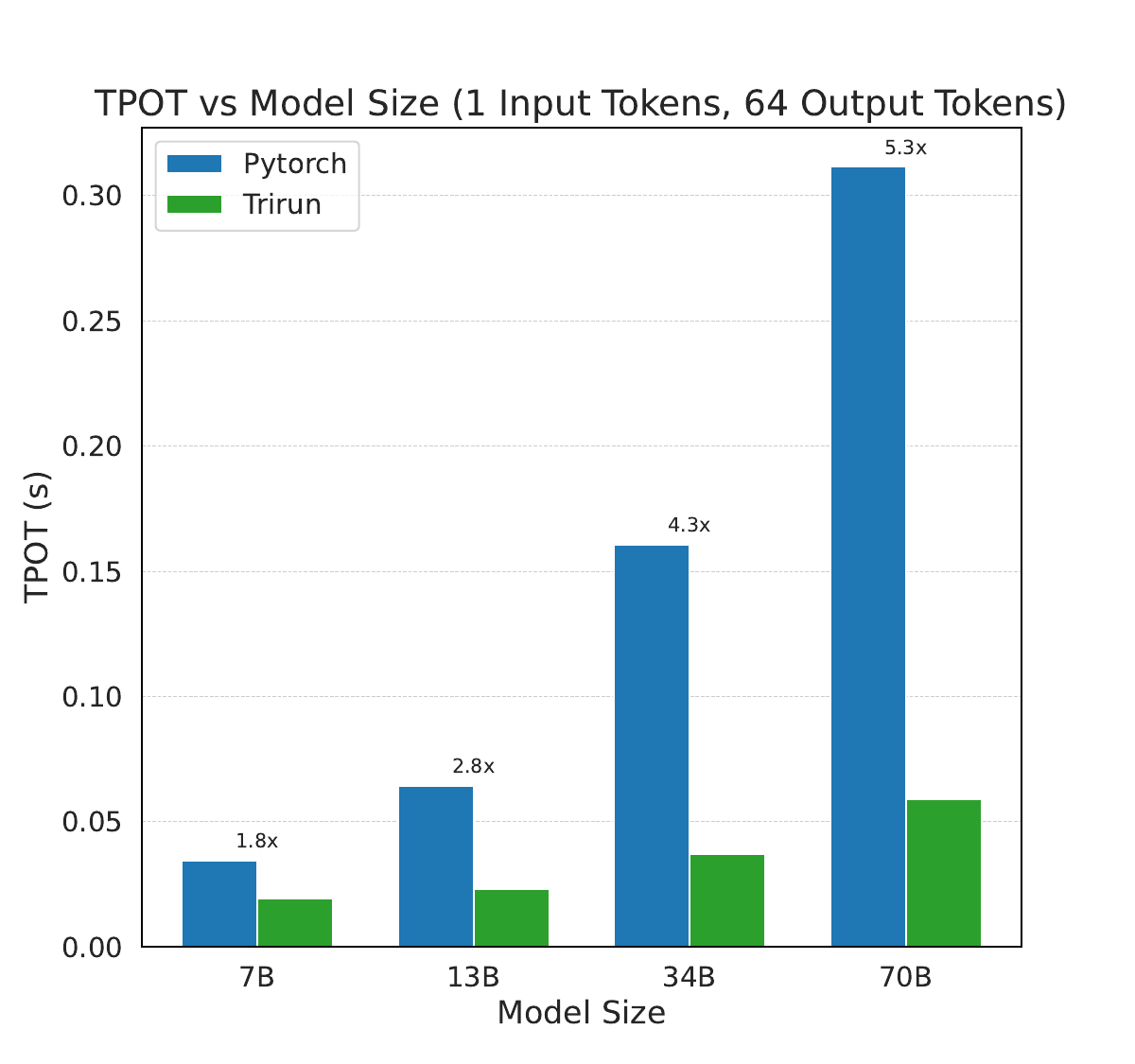}
    \caption{\small Comparison of TriRun kernels with the FP16 PyTorch baseline on NVIDIA L40S, L40, A40, and 4090 (top to bottom). 
    More details in \cref{tab:end2end_L40s}, \cref{tab:end2end_4090}, \cref{tab:end2end_L40} and \cref{tab:end2end_A40}  
    \textbf{(a)} Left: Time to first token,  
    \textbf{(b)} Right: Time per output token.}
    \label{fig:end2end_trirun_gpu}
\end{figure*}

\paragraph{End-to-End Generation Performance.} \cref{fig:end2end_trirun_gpu} presents the end-to-end token generation speedup. For a comprehensive view, \cref{fig:end2end_trirun_gpu} (with detailed results in \cref{tab:end2end_L40s}, \cref{tab:end2end_4090}, \cref{tab:end2end_L40}, and \cref{tab:end2end_A40}) shows the total time for end-to-end token generation using TriRun on Nvidia L40s, L40, A40, and 4090 GPUs for models ranging from 7B to 70B parameters. This reflects the output token throughput, with TriRun achieving up to approximately 5× speedup. The slightly lower end-to-end speedup compared to the per-layer results can be attributed to additional inference overheads beyond the linear layers, which are specifically accelerated by TriRun. In this case as well, larger models achieve higher speedups due to their increased proportion of ternary weights.

\section{Artifacts Released}
To foster open research, we will be making the artifacts from this paper publicly available. The following resources are provided:

\begin{itemize}
\item \textbf{Spectra-1.1 Suite.} We are releasing all models from the Spectra-1.1 suite (as described in \cref{sec:scaling_ternary_models_to_1T_tokens}), along with all intermediate checkpoints, under the  MIT license.

\item \textbf{TriRun Kernels:} We plan to open-source the TriRun Library (see \cref{sec:trirun_main_section}) under the Apache 2.0 license at \url{https://github.com/NolanoOrg/TriRun}.
\end{itemize}

\newpage

% ==========L40s End to End table ===========

\begin{table*}
\centering
\begin{tabular}{lllrrrrrrr}
\toprule

Size &  Kernel & \#GPU &      1 &      2 &      4 &      8 &     16 &     32 &     64 \\
\midrule
\multicolumn{9}{c}{\textbf{Encoding}} \\
\midrule
  7B & Pytorch &        1 & 0.0210 & 0.0218 & 0.0219 & 0.0221 & 0.0225 & 0.0237 & 0.0244 \\
  7B &  Trirun &        1 & 0.0135 & 0.0146 & 0.0146 & 0.0145 & 0.0146 & 0.0146 & 0.0146 \\
  7B & Speedup &        - & 1.5556 & 1.4932 & 1.5000 & 1.5241 & 1.5411 & 1.6233 & 1.6712 \\
\midrule
 13B & Pytorch &        1 & 0.0380 & 0.0401 & 0.0402 & 0.0405 & 0.0411 & 0.0431 & 0.0461 \\
 13B &  Trirun &        1 & 0.0184 & 0.0195 & 0.0193 & 0.0194 & 0.0207 & 0.0195 & 0.0195 \\
 13B & Speedup &        - & 2.0652 & 2.0564 & 2.0829 & 2.0876 & 1.9855 & 2.2103 & 2.3641 \\
\midrule
 34B & Pytorch &        2 & 0.0986 & 0.1025 & 0.1027 & 0.1036 & 0.1051 & 0.1126 & 0.1213 \\
 34B &  Trirun &        1 & 0.0277 & 0.0292 & 0.0288 & 0.0287 & 0.0288 & 0.0288 & 0.0339 \\
 34B & Speedup &        - & 3.5596 & 3.5103 & 3.5660 & 3.6098 & 3.6493 & 3.9097 & 3.5782 \\
\midrule
 70B & Pytorch &        4 & 0.1952 & 0.2062 & 0.2066 & 0.2076 & 0.2093 & 0.2300 & 0.2352 \\
 70B &  Trirun &        1 & 0.0381 & 0.0399 & 0.0397 & 0.0396 & 0.0397 & 0.0403 & 0.0544 \\
 70B & Speedup &        - & 5.1234 & 5.1679 & 5.2040 & 5.2424 & 5.2720 & 5.7072 & 4.3235 \\
\midrule
123B &  Trirun &        1 & 0.0544 & 0.0556 & 0.0557 & 0.0559 & 0.0566 & 0.0617 & 0.0850 \\

\midrule
\multicolumn{9}{c}{\textbf{Decoding - (Input Length: 1 Token)}} \\
\midrule

  7B & Pytorch &        1 & 0.0229 & 0.0449 & 0.0889 & 0.1768 & 0.3529 & 0.7047 &  1.4133 \\
  7B &  Trirun &        1 & 0.0152 & 0.0299 & 0.0596 & 0.1193 & 0.2374 & 0.4748 &  0.9463 \\
  7B & Speedup &        - & 1.5066 & 1.5017 & 1.4916 & 1.4820 & 1.4865 & 1.4842 &  1.4935 \\
\midrule
 13B & Pytorch &        1 & 0.0399 & 0.0791 & 0.1573 & 0.3132 & 0.6250 & 1.2494 &  2.5005 \\
 13B &  Trirun &        1 & 0.0196 & 0.0388 & 0.0777 & 0.1555 & 0.3114 & 0.6219 &  1.2411 \\
 13B & Speedup &        - & 2.0357 & 2.0387 & 2.0245 & 2.0141 & 2.0071 & 2.0090 &  2.0147 \\
\midrule
 34B & Pytorch &        2 & 0.1009 & 0.2009 & 0.4011 & 0.8014 & 1.6022 & 3.2050 &  6.4134 \\
 34B &  Trirun &        1 & 0.0300 & 0.0603 & 0.1203 & 0.2408 & 0.4815 & 0.9632 &  1.9272 \\
 34B & Speedup &        - & 3.3633 & 3.3317 & 3.3342 & 3.3281 & 3.3275 & 3.3275 &  3.3278 \\
\midrule
 70B & Pytorch &        4 & 0.1975 & 0.3941 & 0.7877 & 1.5746 & 3.1491 & 6.2989 & 12.6034 \\
 70B &  Trirun &        1 & 0.0400 & 0.0808 & 0.1615 & 0.3244 & 0.6501 & 1.3005 &  2.6025 \\
 70B & Speedup &        - & 4.9375 & 4.8775 & 4.8774 & 4.8539 & 4.8440 & 4.8434 &  4.8428 \\
\midrule
123B &  Trirun &        1 & 0.0566 & 0.1126 & 0.2246 & 0.4488 & 0.8976 & 1.7936 &  3.5924 \\
\bottomrule
\end{tabular}
\caption{End-to-end inference time (in seconds) on NVIDIA L40s GPUs, comparing Trirun kernels to PyTorch FP16 for varying sequence lengths, showing the speedup of Trirun relative to PyTorch FP16.}
\label{tab:end2end_L40s}
\end{table*}

% ==========4090 End to End table ===========

% Encoding Table
\begin{table*}
\centering
\begin{tabular}{lllrrrrrrr}
\toprule
Size &  Kernel & \#GPU &      1 &      2 &      4 &      8 &     16 &     32 &     64 \\
\midrule
\multicolumn{9}{c}{\textbf{Encoding}} \\
\midrule
7B & Pytorch & 1 & 0.0166 & 0.0174 & 0.0175 & 0.0175 & 0.0179 & 0.0185 & 0.0198 \\
7B & Trirun & 1 & 0.0085 & 0.0092 & 0.0093 & 0.0093 & 0.0093 & 0.0092 & 0.0117 \\
\midrule
7B & Speedup & - & 1.9484 & 1.8871 & 1.8869 & 1.8890 & 1.9288 & 2.0141 & 1.6963 \\
\midrule
13B & Pytorch & 1 & 0.0316 & 0.0325 & 0.0326 & 0.0330 & 0.0336 & 0.0364 & 0.0380 \\
13B & Trirun & 1 & 0.0106 & 0.0115 & 0.0113 & 0.0112 & 0.0112 & 0.0113 & 0.0177 \\
\midrule
13B & Speedup & - & 2.9774 & 2.8384 & 2.8953 & 2.9307 & 2.9911 & 3.2055 & 2.1415 \\
\midrule
34B & Pytorch & 2 & 0.0794 & 0.0809 & 0.0815 & 0.0823 & 0.0840 & 0.0899 & 0.0934 \\
34B & Trirun & 1 & 0.0171 & 0.0180 & 0.0178 & 0.0178 & 0.0187 & 0.0258 & 0.0408 \\
\midrule
34B & Speedup & - & 4.6470 & 4.5056 & 4.5781 & 4.6294 & 4.4869 & 3.4866 & 2.2901 \\
\midrule
70B & Pytorch & 4 & 0.1547 & 0.1598 & 0.1605 & 0.1615 & 0.1633 & 0.1667 & 0.1825 \\
70B & Trirun & 1 & 0.0285 & 0.0293 & 0.0294 & 0.0295 & 0.0300 & 0.0418 & 0.0712 \\
\midrule
70B & Speedup & - & 5.4276 & 5.4595 & 5.4535 & 5.4685 & 5.4451 & 3.9857 & 2.5645 \\

\midrule
\multicolumn{9}{c}{\textbf{Decoding - (Input Length: 1 Token)}} \\
\midrule

7B & Pytorch & 1 & 0.0175 & 0.0346 & 0.0688 & 0.1371 & 0.2744 & 0.5477 & 1.0979 \\
7B & Trirun & 1 & 0.0097 & 0.0193 & 0.0390 & 0.0773 & 0.1546 & 0.3095 & 0.6196 \\
\midrule
7B & Speedup & - & 1.7969 & 1.7952 & 1.7629 & 1.7734 & 1.7749 & 1.7694 & 1.7720 \\
\midrule
13B & Pytorch & 1 & 0.0325 & 0.0645 & 0.1286 & 0.2566 & 0.5122 & 1.0240 & 2.0510 \\
13B & Trirun & 1 & 0.0116 & 0.0232 & 0.0462 & 0.0925 & 0.1850 & 0.3702 & 0.7381 \\
\midrule
13B & Speedup & - & 2.8040 & 2.7841 & 2.7841 & 2.7751 & 2.7683 & 2.7658 & 2.7787 \\
\midrule
34B & Pytorch & 2 & 0.0805 & 0.1607 & 0.3211 & 0.6422 & 1.2843 & 2.5693 & 5.1450 \\
34B & Trirun & 1 & 0.0185 & 0.0373 & 0.0748 & 0.1499 & 0.3039 & 0.6014 & 1.2210 \\
\midrule
34B & Speedup & - & 4.3532 & 4.3132 & 4.2912 & 4.2828 & 4.2254 & 4.2721 & 4.2136 \\
\midrule
70B & Pytorch & 4 & 0.1559 & 0.3115 & 0.6230 & 1.2461 & 2.4920 & 4.9844 & 9.9766 \\
70B & Trirun & 1 & 0.0297 & 0.0590 & 0.1180 & 0.2356 & 0.4725 & 0.9471 & 1.8942 \\
\midrule
70B & Speedup & - & 5.2519 & 5.2760 & 5.2804 & 5.2893 & 5.2741 & 5.2626 & 5.2670 \\
\bottomrule
\end{tabular}
% \caption{End-2-end inference performance across varying number of tokens on 4090 with our Kernels over Pytorch's Fp16}
\caption{End-to-end inference time (in seconds) on NVIDIA 4090 GPUs, comparing Trirun kernels to PyTorch FP16 for varying sequence lengths, showing the speedup of Trirun relative to PyTorch FP16.}

\label{tab:end2end_4090}
\end{table*}

% =============== L40 ===============
% Encoding Table
\begin{table*}
\centering
\begin{tabular}{lllrrrrrrr}
\toprule
Size &  Kernel & \#GPU &      1 &      2 &      4 &      8 &     16 &     32 &     64 \\
\midrule
\multicolumn{9}{c}{\textbf{Encoding}} \\
\midrule
7B & Pytorch & 1 & 0.0210 & 0.0219 & 0.0220 & 0.0221 & 0.0225 & 0.0244 & 0.0258 \\
7B & Trirun & 1 & 0.0129 & 0.0139 & 0.0139 & 0.0142 & 0.0139 & 0.0138 & 0.0138 \\
\midrule
7B & Speedup & - & 1.6239 & 1.5749 & 1.5810 & 1.5552 & 1.6193 & 1.7617 & 1.8681 \\
\midrule
13B & Pytorch & 1 & 0.0381 & 0.0402 & 0.0403 & 0.0407 & 0.0413 & 0.0458 & 0.0481 \\
13B & Trirun & 1 & 0.0157 & 0.0170 & 0.0173 & 0.0174 & 0.0174 & 0.0170 & 0.0184 \\
\midrule
13B & Speedup & - & 2.4219 & 2.3633 & 2.3306 & 2.3368 & 2.3765 & 2.6850 & 2.6097 \\
\midrule
34B & Pytorch & 2 & 0.0990 & 0.1028 & 0.1031 & 0.1038 & 0.1054 & 0.1180 & 0.1226 \\
34B & Trirun & 1 & 0.0268 & 0.0281 & 0.0280 & 0.0278 & 0.0281 & 0.0279 & 0.0427 \\
\midrule
34B & Speedup & - & 3.6857 & 3.6630 & 3.6814 & 3.7370 & 3.7462 & 4.2331 & 2.8736 \\
\midrule
70B & Pytorch & 4 & 0.1958 & 0.2068 & 0.2071 & 0.2080 & 0.2097 & 0.2167 & 0.2358 \\
70B & Trirun & 1 & 0.0358 & 0.0371 & 0.0367 & 0.0368 & 0.0372 & 0.0439 & 0.0723 \\
\midrule
70B & Speedup & - & 5.4757 & 5.5771 & 5.6356 & 5.6579 & 5.6366 & 4.9367 & 3.2611 \\

\midrule
\multicolumn{9}{c}{\textbf{Decoding - (Input Length: 1 Token)}} \\
\midrule

7B & Pytorch & 1 & 0.0227 & 0.0448 & 0.0889 & 0.1769 & 0.3531 & 0.7043 & 1.4104 \\
7B & Trirun & 1 & 0.0144 & 0.0285 & 0.0571 & 0.1144 & 0.2284 & 0.4579 & 0.9093 \\
\midrule
7B & Speedup & - & 1.5759 & 1.5705 & 1.5579 & 1.5464 & 1.5461 & 1.5382 & 1.5511 \\
\midrule
13B & Pytorch & 1 & 0.0399 & 0.0792 & 0.1578 & 0.3145 & 0.6273 & 1.2536 & 2.5101 \\
13B & Trirun & 1 & 0.0175 & 0.0356 & 0.0711 & 0.1407 & 0.2779 & 0.5560 & 1.1106 \\
\midrule
13B & Speedup & - & 2.2862 & 2.2228 & 2.2203 & 2.2343 & 2.2569 & 2.2546 & 2.2600 \\
\midrule
34B & Pytorch & 2 & 0.1013 & 0.2020 & 0.4031 & 0.8054 & 1.6102 & 3.2202 & 6.4492 \\
34B & Trirun & 1 & 0.0291 & 0.0588 & 0.1173 & 0.2355 & 0.4711 & 0.9484 & 1.8970 \\
\midrule
34B & Speedup & - & 3.4876 & 3.4350 & 3.4359 & 3.4195 & 3.4183 & 3.3955 & 3.3997 \\
\midrule
70B & Pytorch & 4 & 0.1987 & 0.3966 & 0.7924 & 1.5842 & 3.1681 & 6.3388 & 12.6871 \\
70B & Trirun & 1 & 0.0374 & 0.0754 & 0.1514 & 0.3002 & 0.6015 & 1.1935 & 2.3860 \\
\midrule
70B & Speedup & - & 5.3117 & 5.2600 & 5.2330 & 5.2774 & 5.2666 & 5.3110 & 5.3172 \\
\bottomrule
\end{tabular}
\caption{End-to-end inference time (in seconds) on NVIDIA L40 GPUs, comparing Trirun kernels to PyTorch FP16 for varying sequence lengths, showing the speedup of Trirun relative to PyTorch FP16.}

\label{tab:end2end_L40}
\end{table*}
% =============Table for A40===============
% Encoding Table
\begin{table*}
\centering
\begin{tabular}{lllrrrrrrr}
\toprule
Size &  Kernel & \#GPU &      1 &      2 &      4 &      8 &     16 &     32 &     64 \\
\midrule
\multicolumn{9}{c}{\textbf{Encoding}} \\
\midrule
7B & Pytorch & 1 & 0.0280 & 0.0281 & 0.0283 & 0.0286 & 0.0300 & 0.0322 & 0.0325 \\
7B & Trirun & 1 & 0.0177 & 0.0188 & 0.0189 & 0.0190 & 0.0192 & 0.0189 & 0.0190 \\
\midrule
7B & Speedup & - & 1.5840 & 1.4939 & 1.4952 & 1.5058 & 1.5642 & 1.7012 & 1.7150 \\
\midrule
13B & Pytorch & 1 & 0.0509 & 0.0522 & 0.0524 & 0.0528 & 0.0535 & 0.0556 & 0.0573 \\
13B & Trirun & 1 & 0.0280 & 0.0301 & 0.0302 & 0.0302 & 0.0302 & 0.0301 & 0.0291 \\
\midrule
13B & Speedup & - & 1.8139 & 1.7323 & 1.7377 & 1.7470 & 1.7745 & 1.8431 & 1.9732 \\
\midrule
34B & Pytorch & 2 & 0.1277 & 0.1331 & 0.1339 & 0.1335 & 0.1357 & 0.1433 & 0.1472 \\
34B & Trirun & 1 & 0.0438 & 0.0452 & 0.0462 & 0.0462 & 0.0460 & 0.0453 & 0.0514 \\
\midrule
34B & Speedup & - & 2.9170 & 2.9425 & 2.8964 & 2.8877 & 2.9513 & 3.1624 & 2.8633 \\
\midrule
70B & Pytorch & 4 & 0.2580 & 0.2673 & 0.2685 & 0.2702 & 0.2752 & 0.2845 & 0.3494 \\
70B & Trirun & 1 & 0.0765 & 0.0792 & 0.0784 & 0.0786 & 0.0790 & 0.0789 & 0.0898 \\
\midrule
70B & Speedup & - & 3.3734 & 3.3747 & 3.4234 & 3.4393 & 3.4826 & 3.6055 & 3.8922 \\

\midrule
\multicolumn{9}{c}{\textbf{Decoding - (Input Length: 1 Token)}} \\
\midrule

7B & Pytorch & 1 & 0.0299 & 0.0590 & 0.1171 & 0.2334 & 0.4662 & 0.9318 & 1.8670 \\
7B & Trirun & 1 & 0.0195 & 0.0391 & 0.0773 & 0.1587 & 0.3101 & 0.6184 & 1.2324 \\
\midrule
7B & Speedup & - & 1.5341 & 1.5073 & 1.5145 & 1.4709 & 1.5033 & 1.5067 & 1.5150 \\
\midrule
13B & Pytorch & 1 & 0.0527 & 0.1046 & 0.2085 & 0.4159 & 0.8314 & 1.6628 & 3.3312 \\
13B & Trirun & 1 & 0.0253 & 0.0506 & 0.1013 & 0.2029 & 0.4051 & 0.8122 & 1.6293 \\
\midrule
13B & Speedup & - & 2.0871 & 2.0666 & 2.0581 & 2.0503 & 2.0526 & 2.0474 & 2.0446 \\
\midrule
34B & Pytorch & 2 & 0.1302 & 0.2594 & 0.5178 & 1.0353 & 2.0716 & 4.1386 & 8.2878 \\
34B & Trirun & 1 & 0.0400 & 0.0818 & 0.1631 & 0.3278 & 0.6513 & 1.3201 & 2.6081 \\
\midrule
34B & Speedup & - & 3.2521 & 3.1735 & 3.1737 & 3.1582 & 3.1804 & 3.1351 & 3.1777 \\
\midrule
70B & Pytorch & 4 & 0.2608 & 0.5204 & 1.0404 & 2.0823 & 4.1608 & 8.3258 & 16.6702 \\
70B & Trirun & 1 & 0.0790 & 0.1219 & 0.2229 & 0.4486 & 0.9942 & 1.8002 & 3.4719 \\
\midrule
70B & Speedup & - & 3.2997 & 4.2700 & 4.6675 & 4.6413 & 4.1851 & 4.6250 & 4.8015 \\
\bottomrule
\end{tabular}
\caption{End-to-end inference time (in seconds) on NVIDIA A40 GPUs, comparing Trirun kernels to PyTorch FP16 for varying sequence lengths, showing the speedup of Trirun relative to PyTorch FP16.}
\label{tab:end2end_A40}
\end{table*}
\end{document}